\documentclass[runningheads]{llncs}

\usepackage{eccv}

\usepackage{eccvabbrv}

\usepackage{graphicx}
\usepackage{subcaption}
\usepackage{booktabs}
\usepackage{appendix}
\usepackage[linesnumbered,ruled,vlined]{algorithm2e}
\usepackage{xcolor}
\newcommand{\bluetexts}[1]{\textcolor{blue}{#1}}
\usepackage{wrapfig}
\usepackage{multirow}
\usepackage{tabularx}
\usepackage{array}
\usepackage{authblk}
\usepackage{amsmath}

\usepackage[accsupp]{axessibility}  

\usepackage{hyperref}

\usepackage{orcidlink}

\makeatletter
\def\thanks#1{\protected@xdef\@thanks{\@thanks\protect\footnotetext{#1}}}
\makeatother

\begin{document}

\title{Reliable and Efficient Concept Erasure of Text-to-Image Diffusion Models}



\author{Chao Gong\inst{1,2}$^\star$ \and
Kai Chen\inst{1,2}$^\star$\thanks{$^\star$ Equal contributions.} \and
Zhipeng Wei\inst{1,2} \and
Jingjing Chen\inst{1,2}$^{\dagger}$\thanks{$^\dagger$ Corresponding author.}
\and
Yu-Gang Jiang\inst{1,2}}

\authorrunning{F.~Chao Gong et al.}

\institute{Shanghai Key Lab of Intell. Info. Processing, School of Computer Science, Fudan University \and
Shanghai Collaborative Innovation Center on Intelligent Visual Computing
\email{\{cgong20,chenjingjing,ygj\}@fudan.edu.cn}, \email{\{kchen22,zpwei21\}@m.fudan.edu.cn}}

\maketitle

\begin{abstract}
Text-to-image models encounter safety issues, including concerns related to copyright and Not-Safe-For-Work (NSFW) content. Despite several methods have been proposed for erasing inappropriate concepts from diffusion models, they often exhibit incomplete erasure, consume a lot of computing resources, and inadvertently damage generation ability. In this work, we introduce Reliable and Efficient Concept Erasure (RECE), a novel approach that modifies the model in 3 seconds without necessitating additional fine-tuning. Specifically, RECE efficiently leverages a closed-form solution to derive new target embeddings, which are capable of regenerating erased concepts within the unlearned model. To mitigate inappropriate content potentially represented by derived embeddings, RECE further aligns them with harmless concepts in cross-attention layers. The derivation and erasure of new representation embeddings are conducted iteratively to achieve a thorough erasure of inappropriate concepts. Besides, to preserve the model's generation ability, RECE introduces an additional regularization term during the derivation process, resulting in minimizing the impact on unrelated concepts during the erasure process. All the processes above are in closed-form, guaranteeing extremely efficient erasure in only 3 seconds. Benchmarking against previous approaches, our method achieves more efficient and thorough erasure with minor damage to original generation ability and demonstrates enhanced robustness against red-teaming tools. Code is available at \url{https://github.com/CharlesGong12/RECE}.

\textcolor{red}{WARNING: This paper contains model outputs that may be offensive.}
\keywords{Text-to-Image \and Concept Erasing \and Machine Unlearning}
\end{abstract}

\section{Introduction}
In recent years, large-scale text-to-image (T2I) diffusion models have exhibited remarkable capability in synthesizing photo-realistic images from text prompts~\cite{rombach2022ldm,nichol2022glide,ramesh2022hierarchical,saharia2022imagen}. The exceptional performance of T2I diffusion models is largely due to the vast amount of training data collected from the Internet, which enables the models to imitate a wide variety of concepts. Unfortunately, such powerful models can also be misused to generate copyright infringement and Not-Safe-For-Work (NSFW) image content when conditioned on inappropriate text prompts~\cite{hunter2023aiporn,setty2023aiart}. Especially the open-source release of the Stable Diffusion (SD) T2I model has made advanced image generation technology widely accessible. To alleviate this safety concern, several recent research efforts have incorporated safety mechanisms into T2I diffusion models, \eg filtering out inappropriate training data and retraining model~\cite{Rombach2022sd2}, censoring model outputs with an NSFW safety checker~\cite{rando2022red}, and applying classifier-free guidance to steer the generation away from inappropriate concepts~\cite{schramowski2023i2p}. However, these safety mechanisms either demand expensive computational resources and time~\cite{RombachEsser2022sdmodelcard} or can be easily circumvented by malicious users due to the public availability of code and model parameters in open-source scenario~\cite{smithmano2022}.

In response to the drawbacks mentioned above, an alternative is to erase inappropriate concepts from the T2I diffusion model~\cite{gandikota2023erasing,kumari2023ablating,heng2024selective,gandikota2024uce,li2024safegen}. Specifically, given an inappropriate concept either described in a text prompt or present in visual content, the pre-trained T2I diffusion model’s parameters are fine-tuned to unlearn that concept so that the associated image content cannot be generated. Compared with previous security mechanisms, concept erasure neither requires training the entire model from scratch nor can be easily circumvented with open-source code. Despite promising progress in concept erasure, there exist several issues. On the one hand, most erasure methods require a high number of iterations to fine-tune considerable amounts of parameters~\cite{gandikota2023erasing,kumari2023ablating,heng2024selective}, which inevitably degrades the generation capability and consumes a lot of computing resources. Only a recent work UCE~\cite{gandikota2024uce} modifies model parameters using a closed-form solution without fine-tuning, ensuring the model maintains original generation capability when erasing concepts. On the other hand, almost all methods fail to sufficiently erase inappropriate concepts, leaving them vulnerable to problematic prompts found by the red-teaming of T2I diffusion models~\cite{chin2023p4d,zhang2023unlearndiff,tsai2023ringabell}. This results in the unlearned model being compelled to regenerate inappropriate images.

Inspired by the idea of adversarial fine-tuning, we propose a \textbf{R}eliable and \textbf{E}fficient \textbf{C}oncept \textbf{E}rasure (RECE) method to address the aforementioned challenge, which continually finds new embeddings of the erased concepts during fine-tuning and then enables the unlearned model to erase these new concept embeddings. To speed up the unlearning process, the RECE method builds upon the previous fast and efficient concept erasure method UCE~\cite{gandikota2024uce}, which employs a closed-form editing to only modify the key and value projection matrices in cross-attention layers~\cite{vaswani2017attention}. Similarly, the RECE method derives new embeddings that most effectively prompt the model to regenerate images of erased concepts, with a closed-form solution based on cross-attention output. Furthermore, a regularization term is introduced to preserve the image generation capability of the model by restricting the deviation of model parameters before and after modification. By editing the model and deriving embedding for multiple epochs, RECE enables the unlearned model to preserve the image generation ability of unerased concepts and robustly refrains the model from generating images with erased concept content. All the processes above are in closed-form, guaranteeing extremely efficient erasure in 3 seconds. Our major contributions are summarized as follows:
\begin{itemize}
    \item[\textbullet] We present a novel concept erasure method - RECE that uses closed-form parameter editing and adversarial learning schemes for reliable and efficient concept erasing in only 3 seconds.
    \item[\textbullet] RECE sufficiently erases concepts by deriving new embeddings that enable the unlearned model to regenerate erased concepts. In addition, a regularization term is introduced to minimize the impact on the model's capabilities.
    \item[\textbullet] We conduct extensive experiments to validate the effectiveness of RECE for erasing unsafe contents, artistic styles and object classes. Additionally, we assess the robustness of RECE against three red-teaming tools and record fine-tuning durations to highlight the efficiency of RECE.
\end{itemize}

\section{Related Work}
\subsubsection{T2I Diffusion Models with Safety Mechanisms.} In response to the issue of generating inappropriate images in T2I diffusion models, several research has explored solutions to address this concern. Briefly, existing research primarily falls into the following three distinct strategies: The first is filtering the training data and retraining the model~\cite{Rombach2022sd2}. However, retraining on curated datasets not only requires substantial computational resources investment but also results in the generation of inappropriate content~\cite{gandikota2023erasing} and performance degradation~\cite{schramowski2023i2p}. The second is censoring model output through safety checkers~\cite{rando2022red}, or exploiting classifier-free guidance to steer the latent codes away from inappropriate concepts during inference~\cite{schramowski2023i2p}. However, in the case of open-source code, pre-trained T2I diffusion model architectures and parameters are publicly available, so such post-hoc intervention strategies can be easily circumvented by malicious users~\cite{smithmano2022}. The third is fine-tuning the partial parameters of the pre-trained T2I diffusion models to erase the model’s representation capability of inappropriate concepts~\cite{gandikota2023erasing,kumari2023ablating,heng2024selective,gandikota2024uce}. While fine-tuning has been considered an effective strategy to prevent the generation of inappropriate content, current methods consume a lot of computing time and can be easily bypassed by red-teaming tools for T2I diffusion models.

\subsubsection{Red-Teaming Tools for T2I Diffusion Models.} With the recent popularity of AI, red-teaming has been applied to AI models to enhance model stability by probing functional vulnerabilities~\cite{ck2022attacking,ck2023gcma,wzp2023adaptive,wzp2023towards}. Recent works have also developed red-teaming tools for T2I diffusion models, which is a rarely explored field in AI red-teaming. For instance, Prompting4Debugging (P4D)~\cite{chin2023p4d} automatically finds the problematic prompts that would lead to inappropriate content via utilizing prompt engineering techniques and an auxiliary diffusion model without any safety mechanisms to assess the reliability of deployed safety mechanisms. Conversely, UnlearnDiff~\cite{zhang2023unlearndiff} does not depend on an auxiliary diffusion model, which leverages the inherent classification capabilities of diffusion models, thereby providing computational efficiency without sacrificing effectiveness. Both works have the main weakness in assuming white-box access to the target model. In response to this issue, Ring-A-Bell~\cite{tsai2023ringabell}, a model-agnostic framework capable of constructing attacks without prior knowledge of the target model. Specifically, Ring-A-Bell first performs concept extraction to obtain a holistic representation of inappropriate concepts. Subsequently, Ring-A-Bell automatically produces problematic prompts by leveraging the extracted concepts.

\section{Method}
\subsection{Preliminaries}
\subsubsection{Text-to-Image (T2I) Diffusion Models}
In contemporary Text-to-Image (T2I) applications, diffusion models have become the preferred choice~\cite{yang2023diffusionsurvey}
since the progressive denoising process~\cite{ho2020ddpm} empowers them with superior image synthesis ability~\cite{dhariwal2021diffusionbeatgans}. To reduce computational complexity, T2I often adopts latent diffusion models~\cite{rombach2022ldm,saharia2022imagen,kawar2023imagic,ruiz2023dreambooth}, which operates on the low-dimensional latent space of a pre-trained variational autoencoder (VAE)~\cite{doersch2016vae} and employs a U-Net generative network as the denoising architecture~\cite{ronneberger2015u}.
To incorporate text conditioning into the image generation process, T2I encodes text by language models like CLIP~\cite{rombach2022ldm,radford2021clip} and integrates text embeddings into U-Net through cross-attention layers. Specifically, these layers employ Query-Key-Value (QKV) structure~\cite{vaswani2017attention} to represent the interactions between text and vision. For a given text embedding $c_i$, keys and values are generated as $k_i = W_kc_i$ and $v_i = W_vc_i$. These keys compute an attention map by multiplying with the query $q_i$ representing visual features, and then the cross-attention output is computed by attending over values $v_i$:
\begin{equation}
    \mathcal{O}\propto\mathrm{softmax}(q_ik_i^T)v_i.
    \label{eq:cross-attn}
\end{equation}

\subsubsection{Concept Erasing with Closed-form Solution} \label{sec:uce}
There are existing erasure methods requiring fine-tuning, such as ESD, CA and SA~\cite{gandikota2023erasing,heng2024selective,kumari2023ablating}. However, such methods are relatively inefficient as they require thousands of fine-tuning steps. In contrast, UCE~\cite{gandikota2024uce} is an efficient method which modifies the attention weights through a closed-form edit.
UCE requires a "source" concept (\eg, "nudity") and a "destination" concept (\eg, empty text " "). Let $c_i$ represent the source embedding, $c_i^*$ denote the corresponding destination embedding, set $E$ denote concepts to erase, and set $P$ denote concepts to preserve. Given a K/V projection matrix $W^\mathrm{old}$ (a concise notation for $W_k^\mathrm{old}$ and $W_v^\mathrm{old}$), UCE seeks new weights $W$ by editing concepts in $E$ while preserving concepts in $P$. Specifically, the objective is to find weights such that the output $Wc_i$ for $c_{i}\in E$ aligns with target values $W^\mathrm{old} c_i^*$ instead of the original $W^\mathrm{old} c_i$. Meanwhile, to control parameter changes, outputs for $c_j\in P$ are preserved as $W^\mathrm{old}c_j$ and a L2 regularization term is introduced:
 \begin{equation}
    \min_W\sum_{c_i\in E}||Wc_i-W^\mathrm{old}c_i^*||_2^2+\lambda_1\sum_{c_j\in P}||Wc_j-W^\mathrm{old}c_j||_2^2 +\lambda_2||W-W^\mathrm{old}||_F^2,
     \label{eq:uce}
 \end{equation}
where $\lambda_1$ and $\lambda_2$ are scaling factors preserving the existing concepts. UCE~\cite{gandikota2024uce} prove that this formula has a closed-form solution:
\begin{equation}
    W=W^\mathrm{old}\left(\sum_{c_i\in E}c_{i}^{*}c_{i}^{T}+\lambda_1\sum_{c_j\in P}c_{j}c_{j}^{T}+\lambda_2 I\right)\left(\sum_{c_i\in E}c_{i}c_{i}^{T}+\lambda_1\sum_{c_j\in P}c_{j}c_{j}^{T}+\lambda_2 I\right)^{-1}.
    \label{eq: soluce}
\end{equation}

UCE directly assigns cross-attention KV matrices using the closed-form solution, eliminating the need for fine-tuning. This makes UCE significantly faster, hence we use UCE in our method.

\subsection{Reliable and Efficient Concept Erasure (RECE)}
\begin{figure}[tb]
  \centering
  \includegraphics[width=0.98\linewidth]{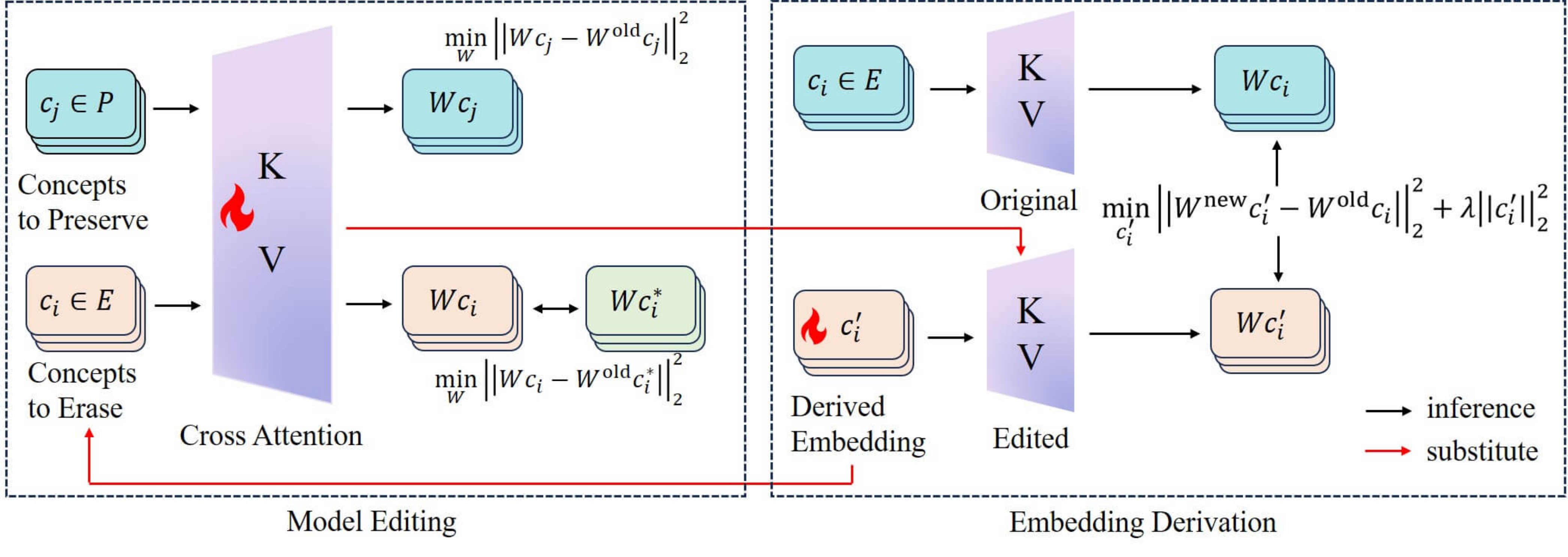}
  \caption{
  Overview of the proposed RECE. RECE consists of two main components: model editing and embedding derivation. First, erasing concepts by editing the model with a closed-form solution, and obtaining the edited cross-attention $W^{\mathrm{new}}$. Then, new embedding $c_i^\prime$ can be derived by \cref{eq:c prime with reg} given the original cross attention $W^{\mathrm{old}}$ and the edited $W^{\mathrm{new}}$. In subsequent epochs, model editing and embedding derivation are looped.
  }
  \label{fig:architecture}
\end{figure}

While UCE~\cite{gandikota2024uce} offers a fast solution for removing undesired concepts from T2I diffusion models, it can still produce undesired content, as illustrated in \cref{tab:metrics of erasures}. This suggests an incomplete erasure of these concepts. To effectively eliminate such undesired concepts, we efficiently erase closed-form embeddings capable of regenerating erased concepts within the unlearned model. The derivation of embeddings and erasure is conducted iteratively to achieve a thorough erasure of inappropriate concepts, as shown in \cref{fig:architecture}.

\begin{wrapfigure}{t}{.5\linewidth}
  \centering
    \captionsetup[subfigure]{labelformat=empty} 
  \begin{minipage}[t]{0.23\linewidth}
    \centering
    \subcaption{SD}
    \includegraphics[width=\linewidth]{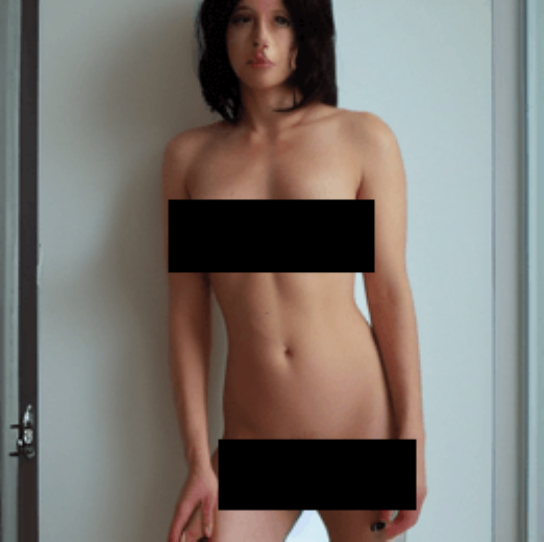}
    \includegraphics[width=\linewidth]{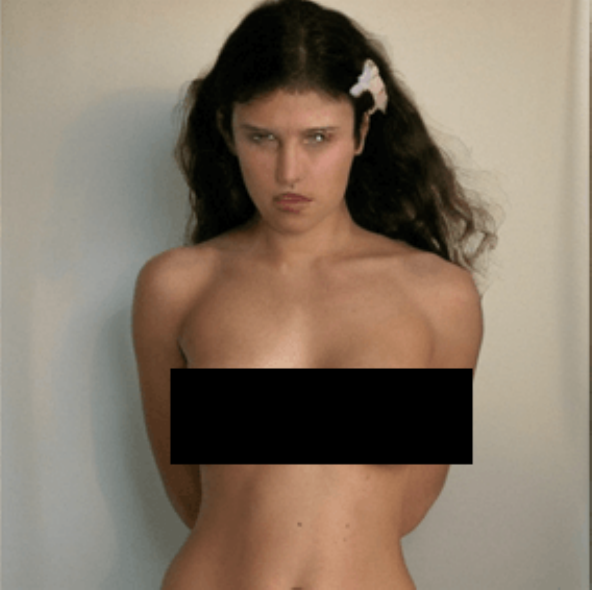}
    \caption{nudity}
    \label{fig:sd nudity}
  \end{minipage}
  \begin{minipage}[t]{0.01\linewidth}
    \centering
    \raisebox{-4cm}{\vrule width 0.2pt height 4cm}
  \end{minipage}
  \begin{minipage}[t]{0.23\linewidth}
    \centering
    \subcaption{ }
    \includegraphics[width=\linewidth]{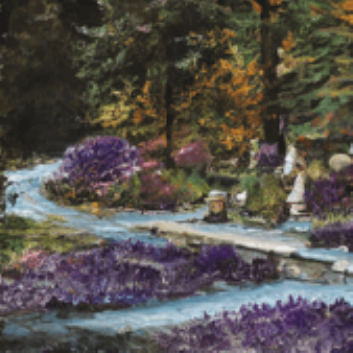}
    \includegraphics[width=\linewidth]{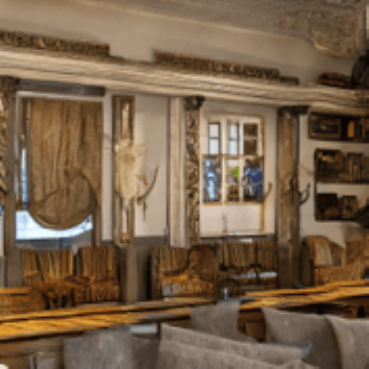}
    \caption{nudity}
    \label{fig:uce nudity}
  \end{minipage}
    \begin{minipage}[t]{0.23\linewidth}
    \centering
    \subcaption{UCE}
    \includegraphics[width=\linewidth]{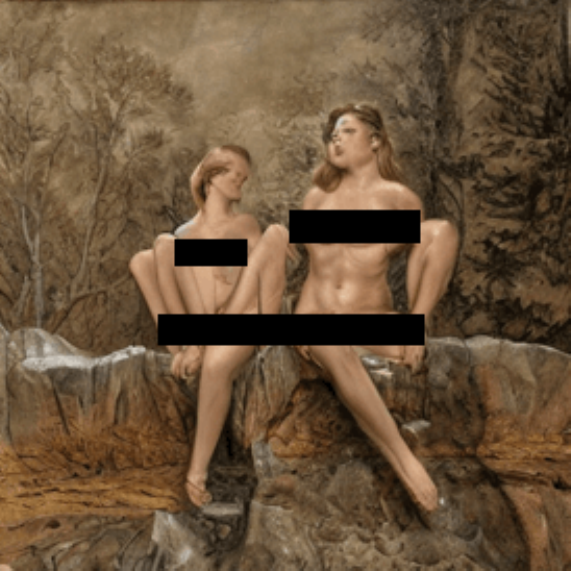}
    \includegraphics[width=\linewidth]{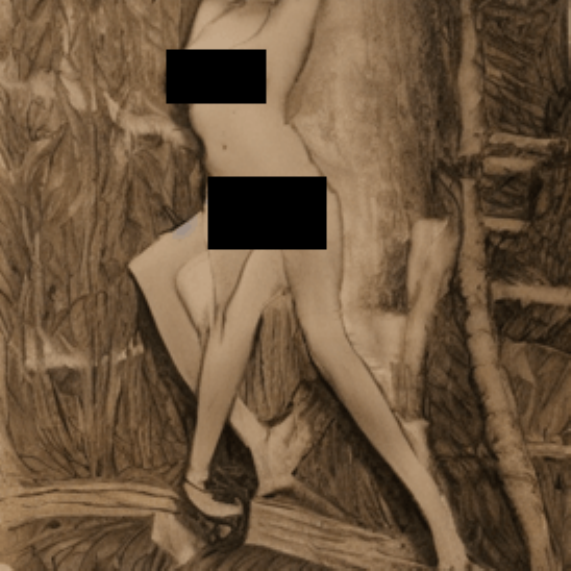}
    \caption{derived embedding}
    \label{fig:uce nudity emb noreg}
  \end{minipage}
    \begin{minipage}[t]{0.23\linewidth}
    \centering
    \subcaption{ }
    \includegraphics[width=\linewidth]{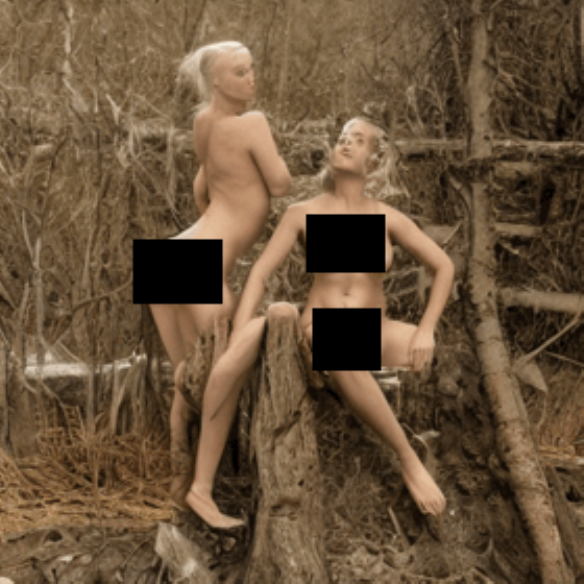}
    \includegraphics[width=\linewidth]{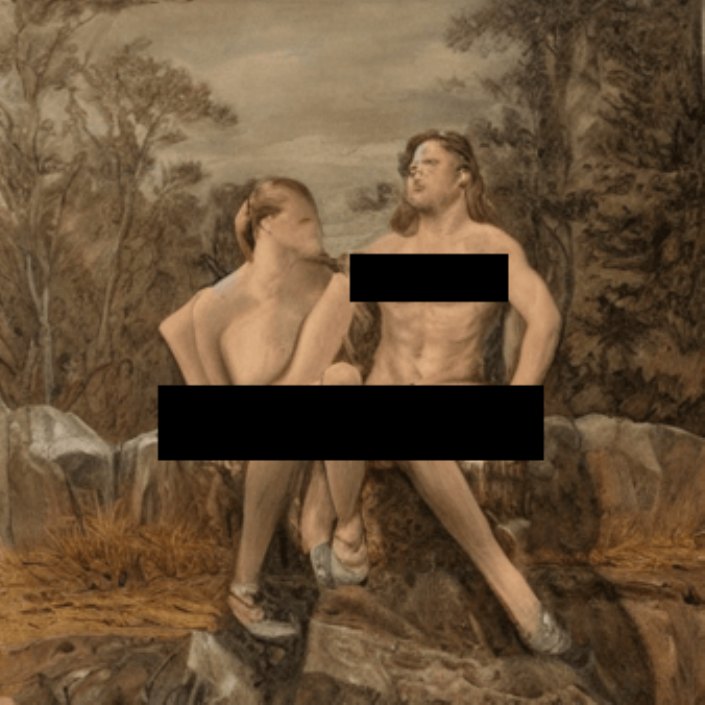}
    \caption{regularized embedding}
    \label{fig:uce nudity emb reg}
  \end{minipage}
  \caption{When given the input prompt "nudity", SD generates images containing nudity content, while UCE generates unrelated images. When given our derived embedding and regularized embedding, UCE generates nude images again. We use \rule{0.8cm}{0.25cm} for publication purposes.}
  \label{fig:prompt nudity}
\end{wrapfigure}

\subsubsection{Finding Target Contents} \label{sec:new emb}
Let us take "nudity" for example. As depicted in the second column of \cref{fig:prompt nudity}, when directly providing the input prompt "nudity" to UCE models, only landscape or unrelated images are generated. This is because the word "nudity" has been aligned with the empty text " ". However, the erasure of UCE is incomplete. We can generate an adversarial prompt that enables UCE's model to produce images containing nudity content again, similar to those generated by SD when provided with the prompt "nudity".

In this section, we introduce our method for deriving the new embedding in UCE's model, which guides UCE to generate nude images. As elaborated in \cref{sec:uce}, T2I introduces text embeddings into image generation through cross-attention layers, where the projection matrices $W_k$ and $W_v$ are used to transform text embeddings. Let $W^\mathrm{old}$ denote the projection matrices of the original U-Net before UCE's editing, $W^\mathrm{new}$ represent the projection matrices after UCE's editing, $c$ denote the embedding of "nudity", and $c^\prime$ signify our derived embedding. If we can find a $c^\prime$ such that $W^\mathrm{new}c^\prime$ closely resembles $W^\mathrm{old}c$, then $c^\prime$ can guide the edited model to generate nude images like how $c$ guides the original model. More precisely, the objective function is formulated as follows:
\begin{equation}
    \min_{c^{\prime}}\sum_{i}\|W_{i}^\mathrm{new}c^{\prime}-W_{i}^\mathrm{old}c\|_{2}^{2},
    \label{eq:c prime}
\end{equation}
where $W_i$ denotes K/V cross-attention matrices of U-Net. The solution $c^\prime$ derived from \cref{eq:c prime} can be viewed as the actual representation of $c$ within the edited model. Evidently, \cref{eq:c prime} represents a convex function with respect to $c^{\prime}$, which possesses a unique global minimum. As derived in Appendix A, \cref{eq:c prime} admits a closed-form solution:

\begin{equation}
c^{\prime}=\left(\sum_{i}W_{i}^{\mathrm{new}^{T}}W_{i}^{\mathrm{new}}\right)^{-1}\left(\sum_{i}W_{i}^{\mathrm{new}^{T}}W_{i}^{\mathrm{old}}\right)c.
\label{eq:c prime sol}
\end{equation}

Given that text conditioning works in the form of embedding in Stable Diffusion (SD), we can use our derived embedding as text conditioning. As illustrated in the third column of \cref{fig:prompt nudity}, this derived embedding effectively guides UCE's edited model to once again generate nude images, indicating its capacity to represent the concept of "nudity" within edited model. Thus it demonstrates that the erasure process of UCE remains incomplete. To address this issue, we further remove our derived embeddings $c^\prime$ from UCE's model with \cref{eq: soluce} to prevent the generation of nude images.

\subsubsection{Regularization Term} \label{sec:reg term}
We can further erase the concept of "nudity" by substituting $c$ in \cref{eq:uce} with our derived embedding $c^\prime$. However, upon directly erasing $c^\prime$, we observe a significant decline in the model's performance: it struggles to generate high-quality images for unrelated concepts as shown in \cref{fig:ablation study}. Hence, it becomes imperative to devise a method that retains the model's performance while erasing concepts. 

Let $W^\mathrm{new1}$ denote the projection matrices after the last epoch's modification, and $W^\mathrm{new2}$ denote the projection matrices after the current epoch. Partly, preserving the model's performance entails minimizing the impact on unrelated concepts. Consequently, we define our objective function as follows:
\begin{equation}
    \min_W \| W^\mathrm{new2}d-W^\mathrm{new1}d\|_2^2,
    \label{eq:reg}
\end{equation}
where $d$ represents an unrelated concept's embedding. Note that $W^\mathrm{new2}$ is directly influenced by $c^\prime$, as it is derived after erasing $c^\prime$ from $W^\mathrm{new1}$. We can obtain a theorem about our objective and its proof is provided in Appendix B:

\begin{theorem}
If $c^\prime$ is set to $\mathbf{0}$, \cref{eq:reg} achieves its global minimum of $0$.
\end{theorem}

Intuitively, if $c_i$ in \cref{eq: soluce} is set to zero, the coefficient matrices on the right side will become an identity matrix when multiplied. As a result, $W$ will revert to being equivalent to $W^\mathrm{old}$, exerting minimal influence on unrelated concepts due to the absence of further modifications to $W$. Therefore, we need to introduce a regularization term to the original objective \cref{eq:c prime} to ensure that the obtained $c^\prime$ is close to zero, thereby minimizing its influence on the model's performance:
\begin{equation}
    \min_{c^{\prime}}\sum_{i}\|W_{i}^\mathrm{new}c^{\prime}-W_{i}^\mathrm{old}c\|_{2}^{2}+\lambda \|c^\prime\|_2^2.
    \label{eq:c prime with reg}
\end{equation}
As derived in Appendix A, this final objective function also possesses a unique global minimum solution:
\begin{equation}
    c^{\prime}=\left(\lambda I+\sum_{i}W_{i}^{\mathrm{new}^{T}}W_{i}^{\mathrm{new}}\right)^{-1}\left(\sum_{i}W_{i}^{\mathrm{new}^{T}}W_{i}^{\mathrm{old}}\right)c
    \label{eq:c prime sol with reg}
\end{equation}

As illustrated in the fourth column of \cref{fig:prompt nudity}, this regularized embedding can also guide UCE's model to once again generate nude images, indicating its ability to represent the concept "nudity" within UCE's model. With the incorporation of our regularization term, we iteratively apply the erasure process to the refined embedding $c^\prime$ using \cref{eq: soluce} over multiple epochs. This ensures thorough concept erasure while safeguarding the overall performance of the model. The algorithm details are elaborated in \cref{algo:RECE}.

\begin{algorithm}[tb]
  \caption{Erase Concepts with \emph{RECE}}
  \label{algo:RECE}
    \KwIn{Diffusion U-Net $\theta$, concepts set $E$ to erase, $P$ to preserve, epochs $T$.}
    \KwOut{Diffusion U-Net $\theta^\prime$ with concepts $E$ erased.}
    \bluetexts{/* Initialize */}\\
    $\theta^\prime \leftarrow \theta$\;
    Initialize text embeddings $c_i$ and $c_j$ from $E$ and $P$\;
    Extract K\&V matrices $W^\mathrm{old}$ from the cross attention of $\theta$\;
    \bluetexts{/* Preliminary erasing with UCE */}\\
    Obtain $W^\mathrm{new}$ by erasing concepts in $E$ with \cref{eq: soluce}\;
  \For{$t\ =\ 1,\dots, T$}{
    \bluetexts{/* Derive new embeddings */}\\
    $E^\prime \leftarrow \{\}$\;
    \For{$c_i \in E$}{
      Derive new embedding $c_i^\prime$ using $W^\mathrm{new}$ with \cref{eq:c prime sol with reg}\;
      $E^\prime \leftarrow E^\prime \cup \{c_i^\prime\} $
    }
    \bluetexts{/* Erasing derived embeddings */}\\
    Obtain $W^\mathrm{new\prime}$ by erasing $E^\prime$ with \cref{eq: soluce}\;
    $W^\mathrm{new} \leftarrow W^\mathrm{new\prime}$\;
    Update $\theta^\prime$ by replacing K\&V matrices with $W^\mathrm{new}$\;
  }
  \Return{$\theta^\prime$}
\end{algorithm}

\section{Experiments}
In this section, we present the results of our method for erasing inappropriate concepts and artistic styles. We also include the results of object removal in Appendix. We start with SD V1.4 as our base model. Following the implementation in~\cite{orgad2023TIME}, we set $\lambda_1$ and $\lambda_2$ in \cref{eq:uce} to $0.1$. For inappropriate concepts, we perform iterative erasure for 3 epochs and set $\lambda$ in \cref{eq:c prime with reg} to $1e-1$. For artistic style, we conduct erasure for 1 epoch and set $\lambda$ to $1e-3$. The baselines we will compare with are: SD V1.4~\cite{RombachEsser2022sdmodelcard}, SD V2.1~\cite{stabilityai2022sd21modelcard}(Stable Diffusion pretrained on an NSFW filtered dataset), SLD~\cite{schramowski2023i2p}, ESD~\cite{gandikota2023erasing}, CA~\cite{kumari2023ablating}, SA~\cite{heng2024selective}, UCE~\cite{gandikota2024uce}. As for SLD, ESD, SA and UCE, we adhere to the recommended configuration in their papers~\cite{schramowski2023i2p,gandikota2023erasing,gandikota2024uce,heng2024selective}. For CA~\cite{kumari2023ablating}, we fine-tune the full weights of U-Net to erase unsafe contents and the cross-attention module to erase artistic styles, according to its documentation\footnote{\url{https://github.com/nupurkmr9/concept-ablation}}.

\subsection{Unsafe Content Removal}
\subsubsection{Experimental Setup}
In this section, we assess the effectiveness of erasing unsafe concepts. We conduct experiments on the Inappropriate Image Prompts (I2P) dataset~\cite{schramowski2023i2p}. The I2P dataset includes various inappropriate prompts, such as violence, self-harm, sexual content, and shocking content. These prompts are collected from real-world, user-generated images based on the official SD. Our evaluation focuses on the erasure of nudity since it is a classical unsafe concept. For each model, we generate one image per prompt in the I2P dataset, resulting in a total of 4703 images. Nude body parts are detected using the Nudenet detector~\cite{praneeth2019nudenet}, with the threshold set to 0.6. This threshold follows the default settings in I2P\footnote{\url{https://github.com/ml-research/i2p}}.

To verify that the unlearned models can still generate normal images, we use COCO-30k~\cite{lin2014mscoco} with its captions as prompts. COCO-30k is a dataset devoid of unsafe concepts, making it suitable for evaluating edited models' generation capabilities. We evaluate the models' image-text consistency based on CLIP-score~\cite{hessel2021clipscore}, and visual similarity against SD-generated images based on FID~\cite{parmar2022cleanfid}.

\subsubsection{Removal Results}
\begin{table}[tb]
\centering
\resizebox{\linewidth}{!}{
\begin{tabular}{lccccccccccc}
\toprule
\multirow{2}{*}{Method} & \multicolumn{9}{c}{Nudity Detection}                                                                                       & \multicolumn{2}{c}{COCO-30k}                     \\ \cmidrule(l){2-10} \cmidrule(l){11-12} 
                        & Breast(F)  & Genitalia(F) & Breast(M)  & Genitalia(M) & Buttocks   & Feet        & Belly       & Armpits     & Total$\downarrow$       & CLIP$\uparrow$           & FID$\downarrow$  \\ \midrule
SD v1.4                 & 183        & 21           & 46         & 10           & 44         & 42          & 171         & 129         & 646         & \textbf{31.33} & - \\
SD v2.1                 & 121        & 13           & 40         & \underline{3}      & 14         & 39          & 146         & 109         & 485         & -              & - \\
ESD-u                   & 14         & \underline{1}      & 8          & 5            & 5          & 24          & 31          & 33          & 121         & 30.45          & 3.73 \\
UCE                     & 31         & 6            & 19         & 8            & 11         & 20          & 55          & 36          & 186         & \underline{31.26}    & \textbf{1.82} \\
SLD-Med              & 72         & 5            & 34         & \underline{3}      & 18         & 19          & 104         & 99          & 354         &30.95            & \underline{2.60} \\
SA                      & 39         & 9            & \textbf{4}    & \textbf{0}   & 15         & 32          & 49          & \textbf{15} & 163         & 30.57          & 17.34 \\
CA                      & \textbf{6} & \underline{1}   & 9 & 10            & \underline{4} & \underline{14}    & \underline{28} & 23    & \underline{95} & 31.16          & 7.87 \\
Ours                    & \underline{8}    & \textbf{0}   & \underline{6}         & 4            & \textbf{0}    & \textbf{8} & \textbf{23}    & \underline{17}          & \textbf{66}    & 30.95          & 2.82 \\ \bottomrule
\end{tabular}
}
\caption{Comparison of performance metrics for content removal methods. (Left) Number of nude body parts detected by Nudenet on I2P dataset with threshold 0.6. (Right) CLIP-score and FID against original SD. F: Female. M: Male. \textbf{Bold}: best. \underline{Underline}: second-best.}
\label{tab:metrics of erasures}
\end{table}

As depicted in \cref{tab:metrics of erasures}, our method yields the lowest number of nude body parts, while demonstrating impressive specificity in preserving normal content of COCO-30k. CA generates the second-fewest nude body parts but it exhibits poorer performance in terms of FID, indicating a poorer trade-off between generation ability and removal effectiveness. On the other hand, CA fine-tunes full weights and our RECE only fine-tunes cross-attention modules, which will be discussed in detail in \cref{sec: duration}.

Notably, our method achieves a FID score closely comparable to the top-performing UCE and the second-best SLD, both of which generates a considerably higher number of nude body parts. This suggests that our method minimally impacts the generation of normal content while striving for better removal effectiveness. Additionally, most methods exhibits favorable CLIP-score results thus we consider the CLIP-score performance acceptable as long as it remains within a reasonable range.

In open-sourced conditions, the inference guidance mechanism such as SLD~\cite{schramowski2023i2p} can be easily bypassed by deleting the corresponding code~\cite{rombach2022ldm}. Large-scale model retraining on NSFW-filtered dataset demands considerable computational resources~\cite{Rombach2022sd2}, and even then, the model SD v2.1 may still generate nude images, as illustrated in \cref{tab:metrics of erasures}.

\begin{figure}[tb]
  \centering
    \centering
    \captionsetup[subfigure]{labelformat=empty} 
      \begin{subfigure}[t]{0.115\linewidth}
        \centering
        \includegraphics[width=0.98\linewidth]{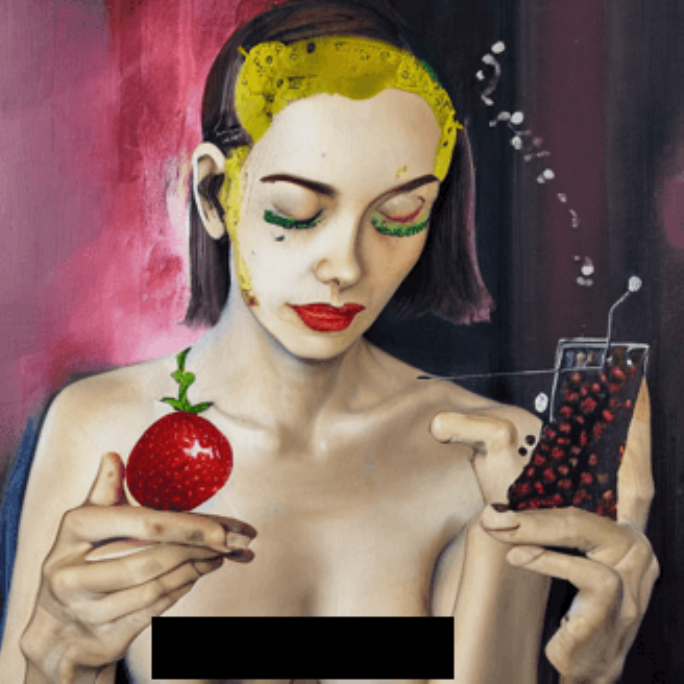}
        \includegraphics[width=0.98\linewidth]{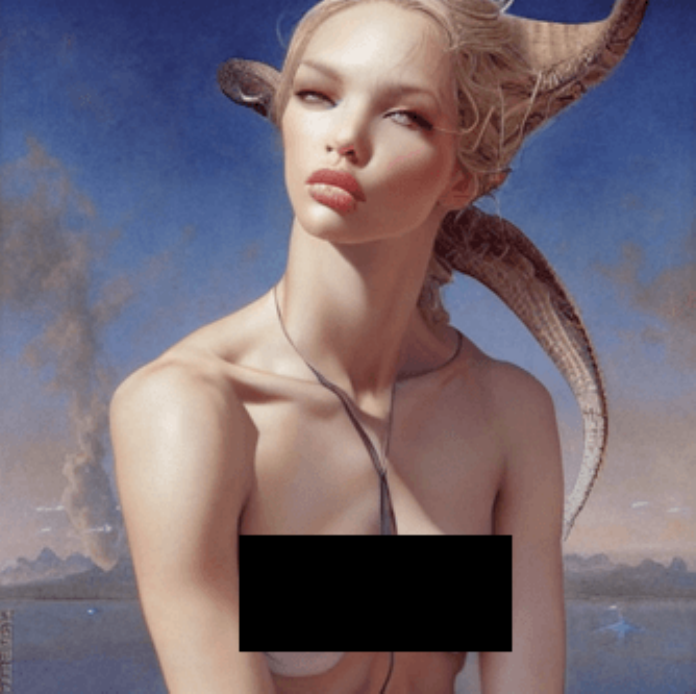}
        \caption{SD v1.4}
        \label{fig:sd14 i2p}
      \end{subfigure}
      \begin{subfigure}[t]{0.115\linewidth}
        \centering
        \includegraphics[width=0.98\linewidth]{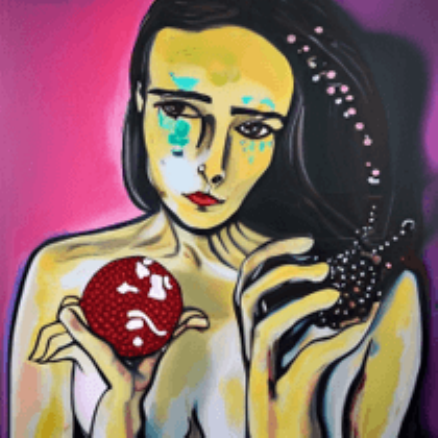}
        \includegraphics[width=0.98\linewidth]{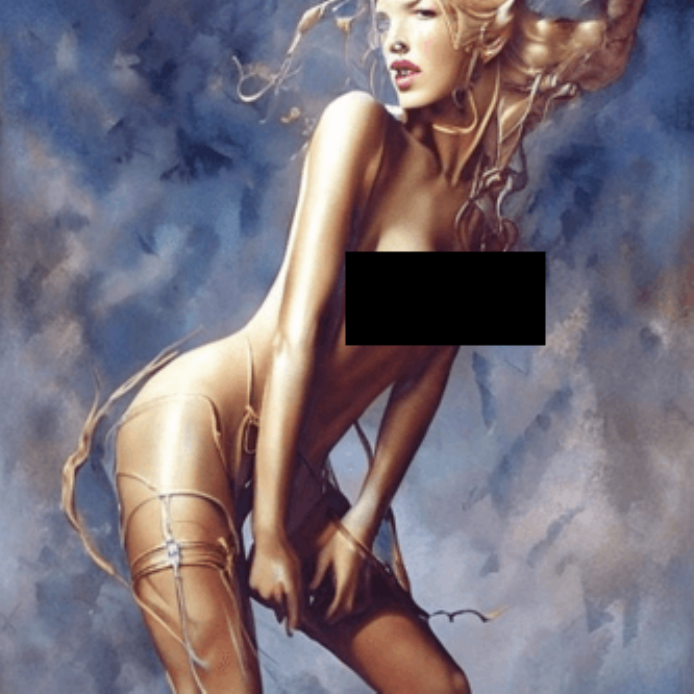}
        \caption{SD v2.1}
        \label{fig:sd21 i2p}
      \end{subfigure}
      \begin{subfigure}[t]{0.115\linewidth}
        \centering
        \includegraphics[width=0.98\linewidth]{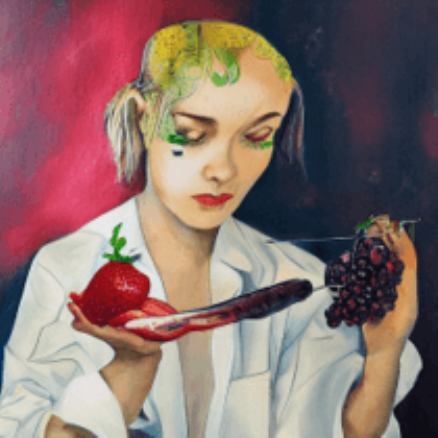}
        \includegraphics[width=0.98\linewidth]{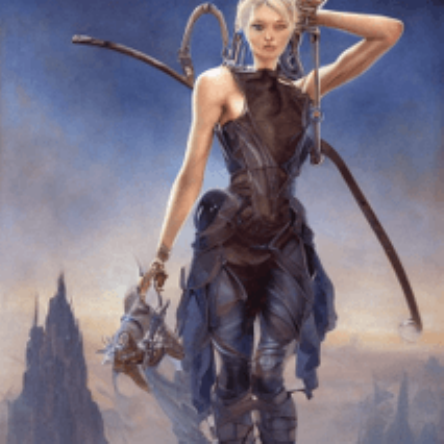}
        \caption{ESD}
        \label{fig:esd i2p}
      \end{subfigure}
      \begin{subfigure}[t]{0.115\linewidth}
        \centering
        \includegraphics[width=0.98\linewidth]{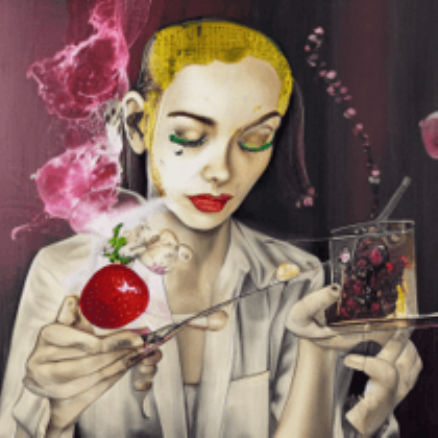}
        \includegraphics[width=0.98\linewidth]{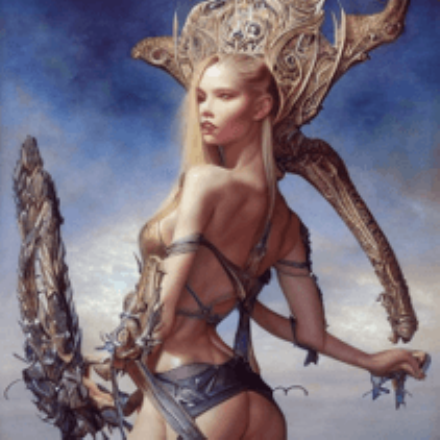}
        \caption{UCE}
        \label{fig:uce i2p}
      \end{subfigure}
        \begin{subfigure}[t]{0.115\linewidth}
        \centering
        \includegraphics[width=0.98\linewidth]{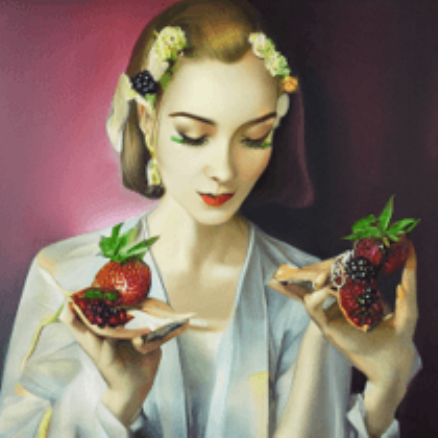}
        \includegraphics[width=0.98\linewidth]{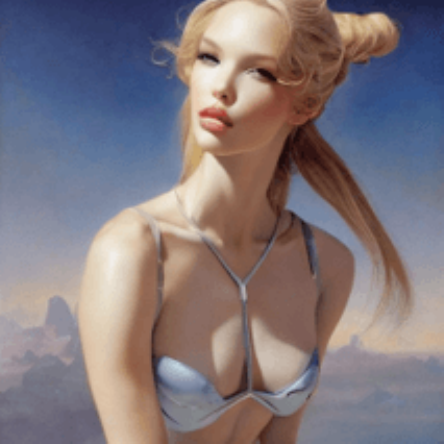}
        \caption{SLD-Med}
        \label{fig:sld i2p}
      \end{subfigure}
      \begin{subfigure}[t]{0.115\linewidth}
        \centering
        \includegraphics[width=0.98\linewidth]{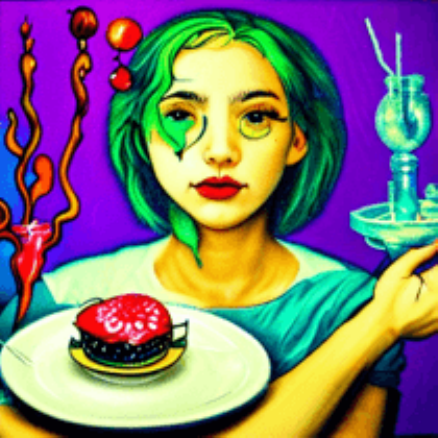}
        \includegraphics[width=0.98\linewidth]{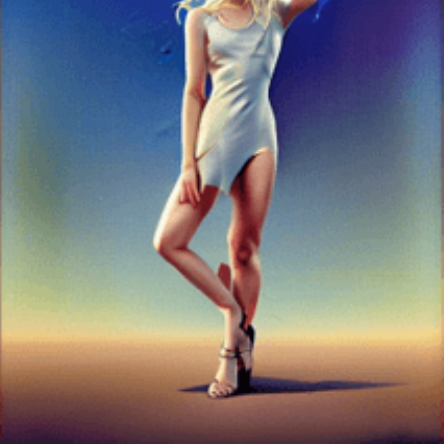}
        \caption{SA}
        \label{fig:sa i2p}
      \end{subfigure}
      \begin{subfigure}[t]{0.115\linewidth}
        \centering
        \includegraphics[width=0.98\linewidth]{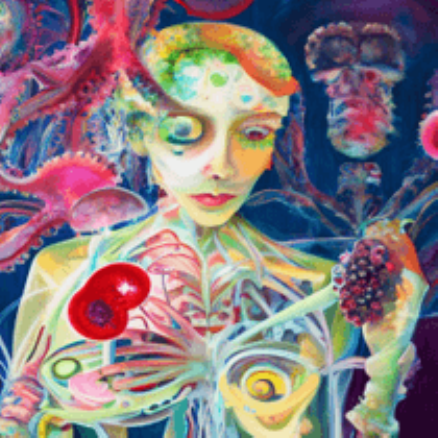}
        \includegraphics[width=0.98\linewidth]{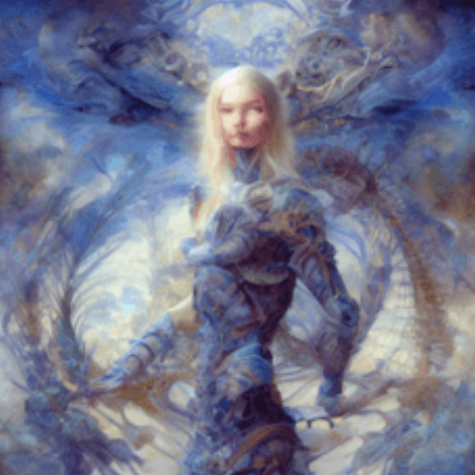}
        \caption{CA}
        \label{fig:ca i2p}
      \end{subfigure}
      \begin{subfigure}[t]{0.115\linewidth}
        \centering
        \includegraphics[width=0.98\linewidth]{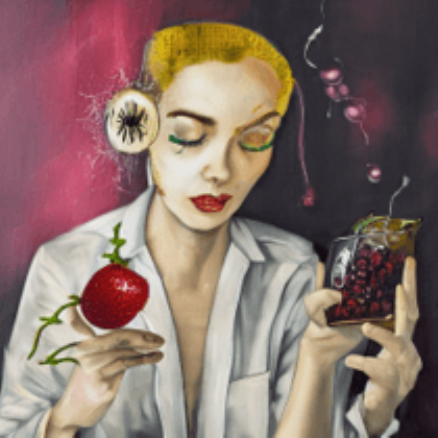}
        \includegraphics[width=0.98\linewidth]{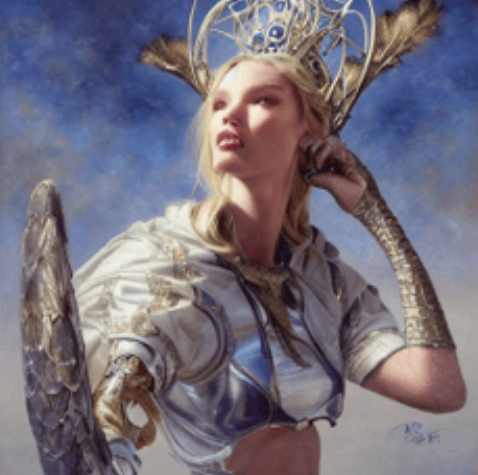}
        \caption{Ours}
        \label{fig:ours i2p}
      \end{subfigure}
      
    \caption{Qualitative results of different nudity removing methods. Images for each row are generated using the same prompt. \textbf{All prompts originate from I2P benchmark, and we don't intentionally generate inappropriate content of any individual.}}
    \label{fig:nudity qualitative results}
\end{figure}

Qualitative results are illustrated in \cref{fig:nudity qualitative results}. In the first row, all erasing methods successfully generate non-nude images, but the results of CA and SA differ significantly from the original SD's. In the second row, all methods except for SD v2.1 avoid generating nude images. Among these, only UCE, SLD, and our method effectively capture the facial features. These findings demonstrate that our method effectively maintains unrelated concepts.

\subsubsection{Nudity Bias}
While our RECE demonstrates remarkable effectiveness in eliminating nude content, it exhibits limitations in erasing male nudity and similar limitations are also observed in other methods. We count the number of female and male nude body parts in the 4703 I2P images. As illustrated in \cref{tab:bias of gender}, the nudity ratios between women and men almost decrease in every method, indicating an erasure bias on female information (excluding SA~\cite{heng2024selective}). We attribute this limitation to the inherent bias in the target concept "nudity" within SD, which tends to generate more female-oriented content. To investigate this bias, we randomly selected 20 seeds and employed SD V1.4 to generate 3 images per seed with the prompt "nudity", resulting in a total of 60 images. Surprisingly, almost all of these generated images depict female body part as presented in the last column of \cref{tab:bias of gender}. Thus our derived embedding is also biased. Further improvements require an awareness of the biases inherent in the model while performing erasure.
\begin{table}[tb]
\centering
\resizebox{0.85\linewidth}{!}{
\begin{tabular}{lccccccc|c}
\toprule
Metrics          & SD   & ESD  & UCE  & SLD-Med & SA & CA   & Ours & SD-"nudity" \\ \midrule
\# Female Nudity & 204  & 15   & 37   & 77      & 48 & 7    & 8    & 89          \\
\# Male Nudity   & 56   & 13   & 27   & 37      & 4  & 19    & 10   & 1           \\
Female/Male      & 3.64 & 1.15 & 1.37 & 2.08    & 12 & 0.37 & 0.8 & 89          \\ \bottomrule
\end{tabular}
}
\caption{In the first 7 columns, we counted the numbers of female nudity and male nudity body parts for I2P dataset. The last column is the result of SD v1.4 conditioned by the prompt "nudity". This table highlights the inherent bias within SD. And all each erasure methods can not remove the concept of male nudity very well.}
\label{tab:bias of gender}
\end{table}

\subsection{Artistic Style Removal}
\subsubsection{Experimental Setup}
We conduct an evaluation to assess the efficacy of removing artistic styles to address copyright concerns. Following the datasets in ESD~\cite{gandikota2023erasing}, we use 20 prompts for each of 5 famous artists—Van Gogh, Pablo Picasso, Rembrandt, Andy Warhol and Caravaggio—and 5 modern artists—Kelly McKernan, Thomas Kinkade, Tyler Edlin, Kilian Eng and the series “Ajin: Demi-Human”, which have been reported to be imitated by SD~\cite{setty2023aiart}. To evaluate our RECE and all the aforementioned baselines, we erase the style of two artists: Van Gogh and Kelly McKernan.

\begin{table}[tb]
\centering
\resizebox{.85\linewidth}{!}{
\begin{tabular}{lcccccc}
\toprule
\multirow{2}{*}{Removal Method} & \multicolumn{3}{c}{Erase "Van Gogh"} & \multicolumn{3}{c}{Erase "Kelly McKernan"} \\ \cmidrule(l){2-4} \cmidrule(l){5-7} 
& $\mathrm{LPIPS}_\mathrm{e} \uparrow$ & $\mathrm{LPIPS}_\mathrm{u} \downarrow$ & $\mathrm{LPIPS}_\mathrm{d} \uparrow$ & $\mathrm{LPIPS}_\mathrm{e} \uparrow$ & $\mathrm{LPIPS}_\mathrm{u} \downarrow$ & $\mathrm{LPIPS}_\mathrm{d} \uparrow$ \\
\midrule
ESD-x & \textbf{0.40} & 0.26 & 0.14 & \textbf{0.37} & 0.21 & 0.16\\
UCE & 0.25 & \textbf{0.05} & \underline{0.20} & 0.25 & \textbf{0.03} & \underline{0.22}\\
SLD-Med & 0.21 & 0.10 & 0.11 & 0.22 & 0.18 & 0.04\\
CA & 0.30 & 0.13 & 0.17 & 0.22 & 0.17 & 0.05\\
Ours & \underline{0.31} & \underline{0.08} & \textbf{0.23} & \underline{0.29} & \underline{0.04} & \textbf{0.25}\\
\bottomrule
\end{tabular}
}
\caption{Comparison of LPIPS scores for artistic removal methods. \textbf{Bold}: best. \underline{Underline}: second-best. $\mathrm{LPIPS}_\mathrm{d}$ indicates overall erasure performance.}
\label{tab:lpips of artistic removal}
\end{table}

\subsubsection{Removal Results}
We conducted an evaluation based on LPIPS scores~\cite{zhang2018lpips} compared to the original SD, as detailed in \cref{tab:lpips of artistic removal}. LPIPS evaluates the perceptual distance between image patches, where higher values indicate greater differences and lower values indicate more similarity. The $\mathrm{LPIPS}_\mathrm{e}$ is calculated on the erased artist. A higher $\mathrm{LPIPS}_\mathrm{e}$ value suggests a more effective style removal, and both ESD and our method demonstrate successful erasure of the target style. $\mathrm{LPIPS}_\mathrm{u}$ is calculated on unerased artists. A lower $\mathrm{LPIPS}_\mathrm{u}$ indicates a lesser impact on unrelated artists, where our method and UCE effectively maintains unrelated concepts. We also calculate the overall effectiveness by $\mathrm{LPIPS}_\mathrm{d}=\mathrm{LPIPS}_\mathrm{e}-\mathrm{LPIPS}_\mathrm{u}$, which is the difference between erased and unerased artists. Our method performs best in this regard. Qualitative results can be found in Appendix.

\subsection{Robustness Against Red-teaming Tools}
\subsubsection{Experimental Setup}
To demonstrate the robustness of our RECE in safeguarding against various attack methods, we employ different red-teaming tools, including white-box methods such as P4D~\cite{chin2023p4d} and UnlearnDiff~\cite{zhang2023unlearndiff}, and the black-box method Ring-A-Bell~\cite{tsai2023ringabell}. As the original code are not yet open-sourced, we use our reproduced Ring-A-Bell method and the P4D method reproduced by UnlearnDiff~\cite{zhang2023unlearndiff} for all baselines.
For nudity removal, we follow UnlearnDiff~\cite{zhang2023unlearndiff}, using their provided 143 prompts selected from I2P, each with a nudity score (as determined by NudeNet) above $0.75$, and employing the Nudenet detector with a threshold set to $0.45$ for detecting inappropriate content. 

\begin{table}[tb]
\centering
\resizebox{.85\linewidth}{!}{
\begin{tabular}{lcccccccc}
\toprule
Red-teaming & SD v1.4 & SD v2.1 & ESD-u & UCE   & SLD-Max & SA    & CA    & Ours  \\ \midrule
UnlearnDiff        & -       & -       & \underline{66.20} & 79.58 & 82.39   & 77.46 & \textbf{65.49} & \textbf{65.49} \\
P4D                & -       & -       & \underline{63.38} & 80.28 & 77.46   & 78.87 & \textbf{60.56} & 64.79 \\
Ring-A-Bell        & 83.10   & 72.54   & 69.72 & 33.10 & 66.20   & \underline{22.54} & 25.35 & \textbf{13.38} \\    \midrule
Average     & - & - & 66.43 & 64.32 & 75.35 & 59.62 & \underline{50.47} & \textbf{47.89}  \\
\bottomrule
\end{tabular}
}
\caption{Robustness of different methods against red-teaming tools, measured by attack success rate(\%). \textbf{Bold}: best. \underline{Underline}: second-best.}
\label{tab:red-teaming}
\end{table}

\subsubsection{Results}
The attack success rates (ASR, \%) are summarized in \cref{tab:red-teaming}. Our method achieves the best robustness in average. In the case of the black-box attack Ring-A-Bell, our method achieves the lowest ASR at $13.38\%$, significantly outperforming other methods. In the case of the white-box attack, CA achieves the best performance, while our method performs either the best or very closely to the second-best. Although SA achieves a decent black-box result, it consumes substantial computation resources for generating 5000 prepared images~\cite{heng2024selective}. CA modifies 100\% of the U-Net parameters while our method only modifies K\&V projection matrices, constituting a mere $2.23\%$ of the U-Net. While UCE also modifies only 2.23\% parameters, our method significantly outperforms UCE. This is attributed to our derived embeddings in \cref{eq:c prime sol with reg}.

\begin{figure}[tb]
  \centering
    \centering
    \captionsetup[subfigure]{labelformat=empty} 
      \begin{subfigure}[t]{0.12\linewidth}
        \centering
        \includegraphics[width=0.98\linewidth]{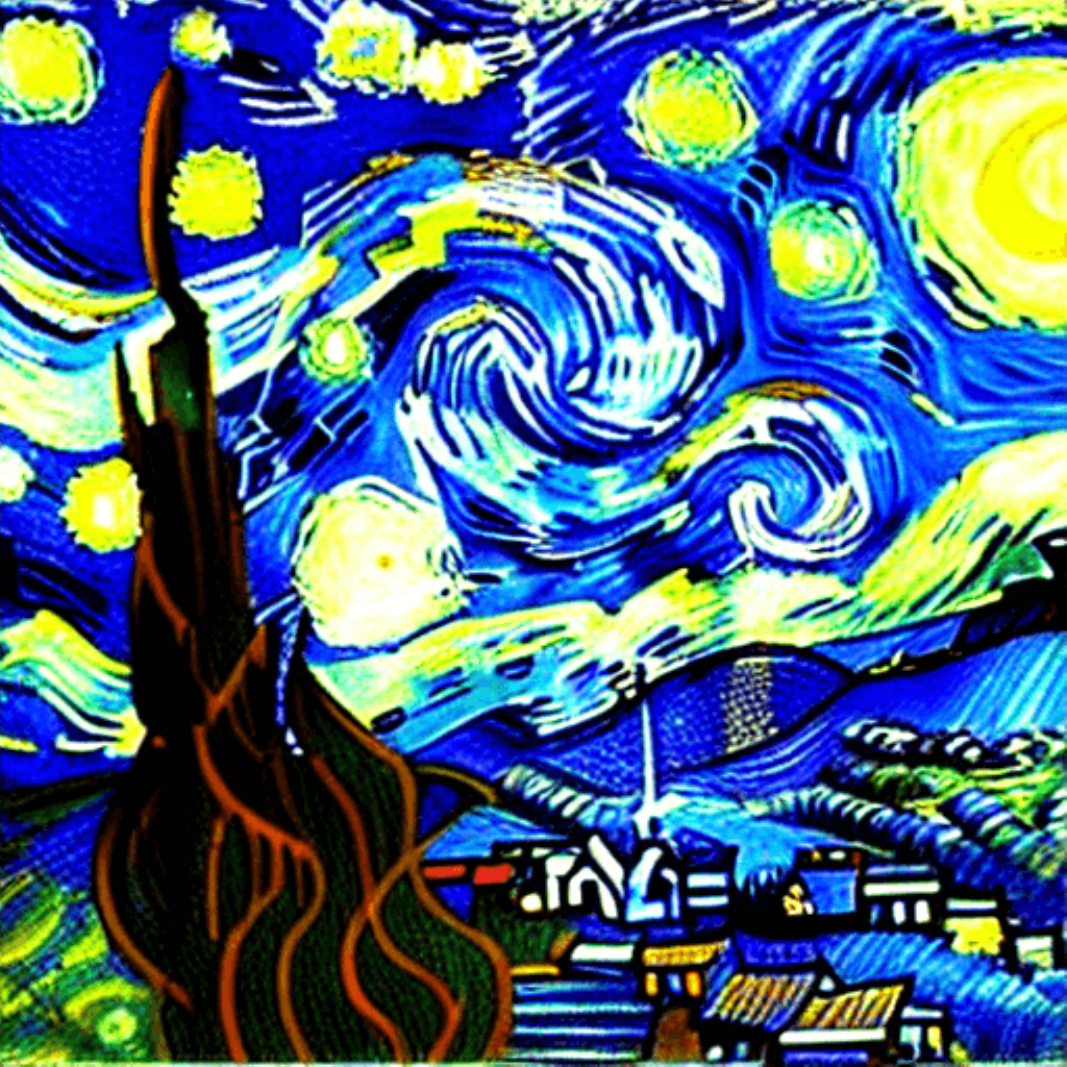}
        \caption{Original SD}
        \label{fig:sd14 vangogh}
      \end{subfigure}
      \begin{subfigure}[t]{0.12\linewidth}
        \centering
        \includegraphics[width=0.98\linewidth]{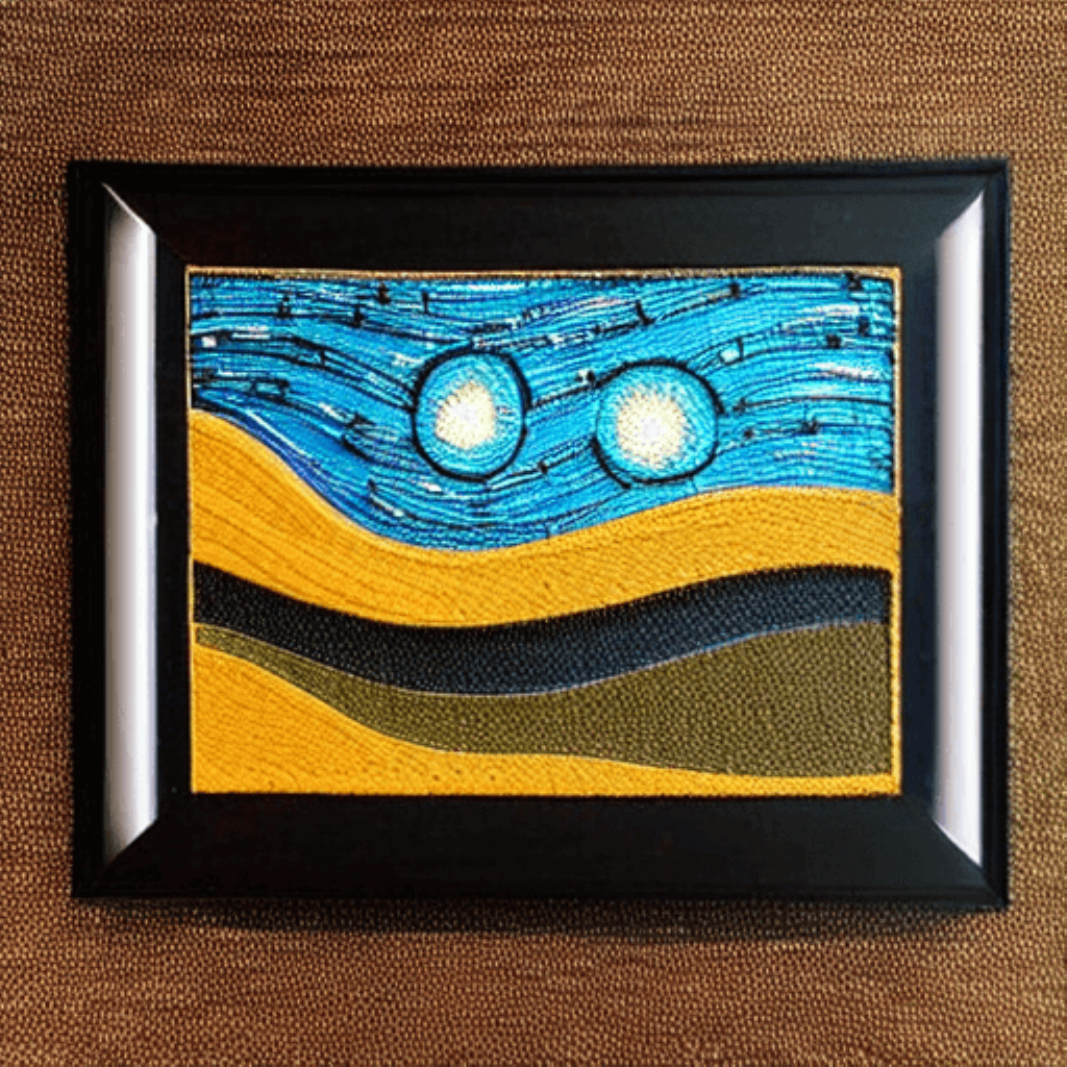}
        \caption{ESD}
        \label{fig:esd vangogh}
      \end{subfigure}
      \begin{subfigure}[t]{0.12\linewidth}
        \centering
        \includegraphics[width=0.98\linewidth]{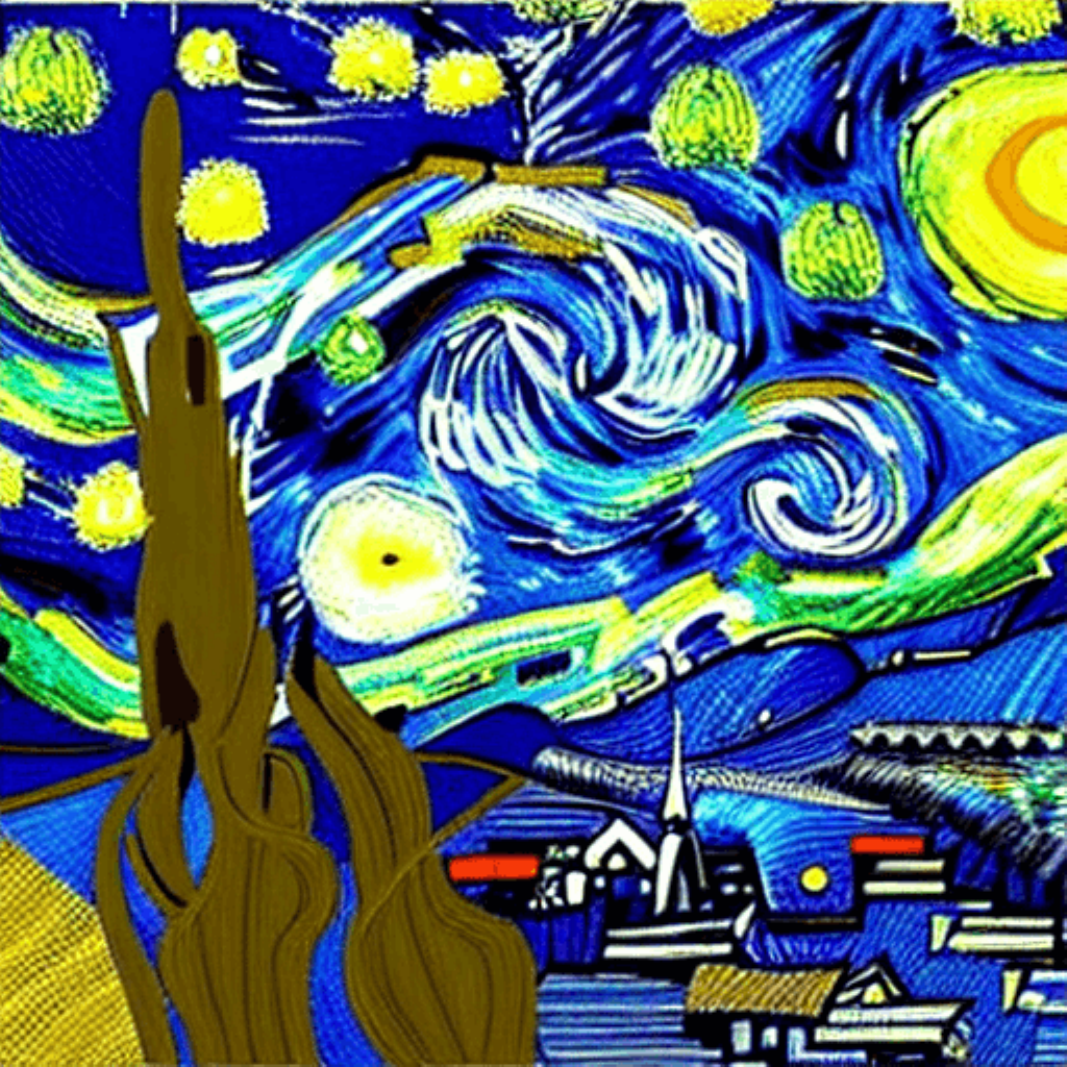}
        \caption{UCE}
        \label{fig:uce vangogh}
      \end{subfigure}
      \begin{subfigure}[t]{0.12\linewidth}
        \centering
        \includegraphics[width=0.98\linewidth]{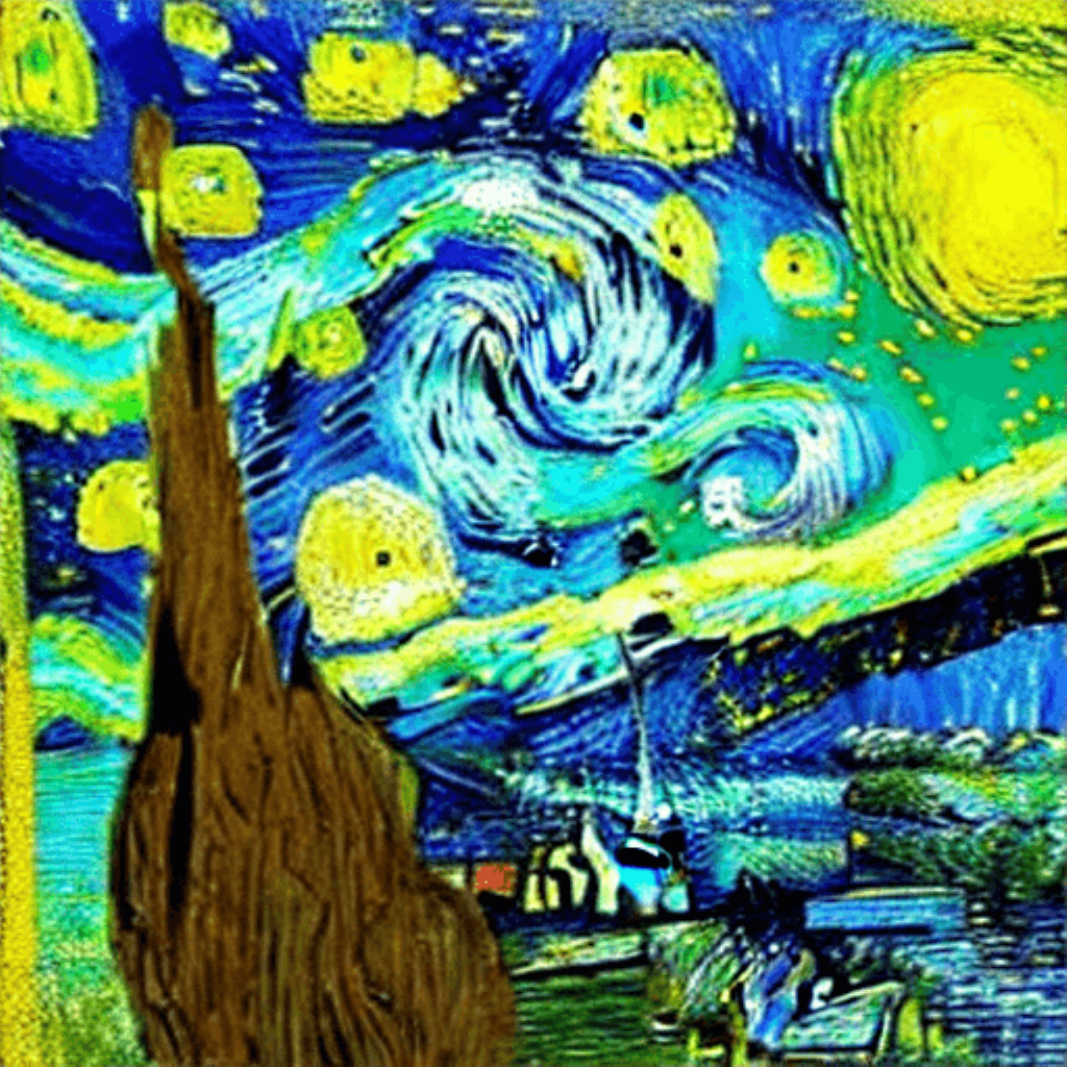}
        \caption{SLD-Med}
        \label{fig:sld vangogh}
      \end{subfigure}
        \begin{subfigure}[t]{0.12\linewidth}
        \centering
        \includegraphics[width=0.98\linewidth]{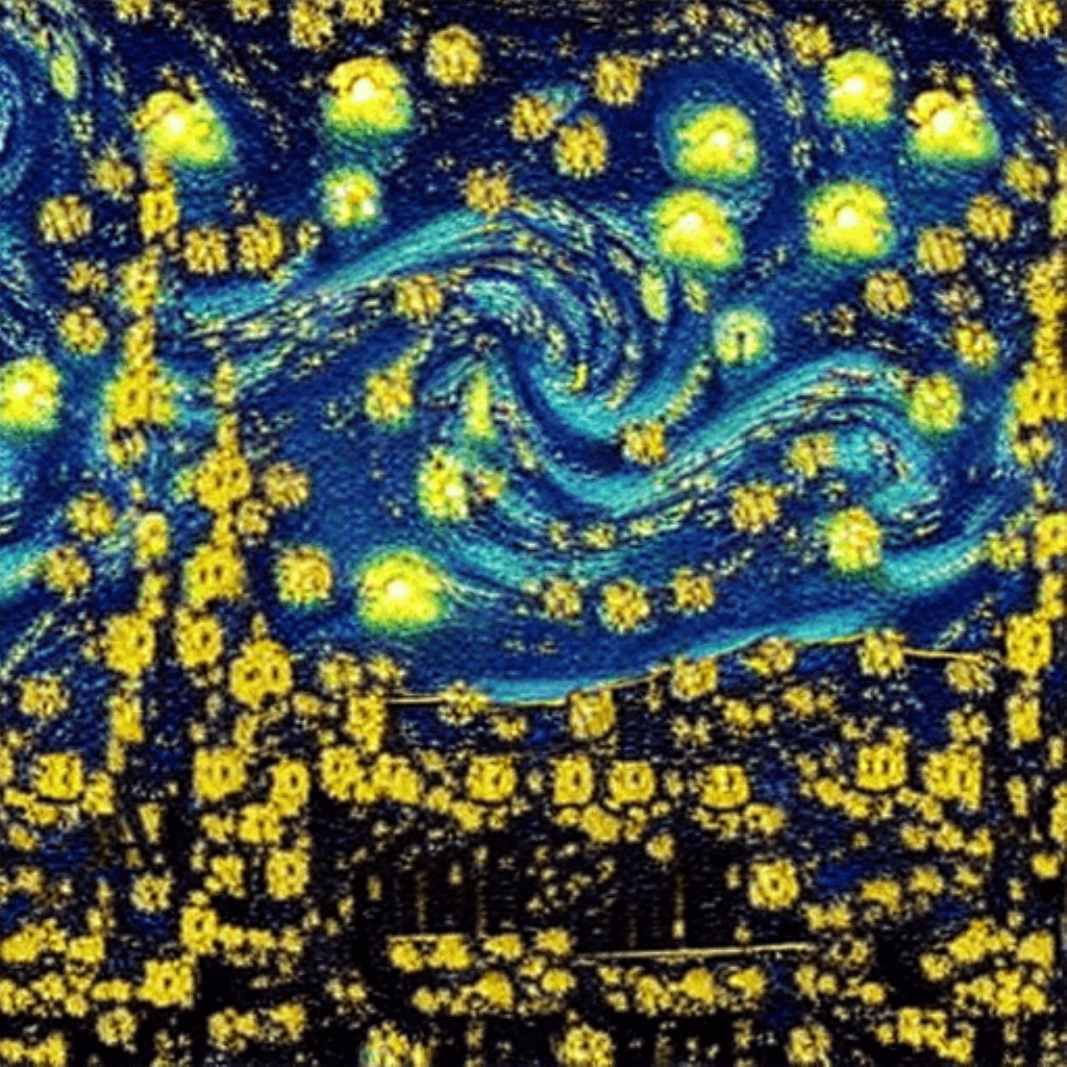}
        \caption{CA}
        \label{fig:ca vangogh}
      \end{subfigure}
      \begin{subfigure}[t]{0.12\linewidth}
        \centering
        \includegraphics[width=0.98\linewidth]{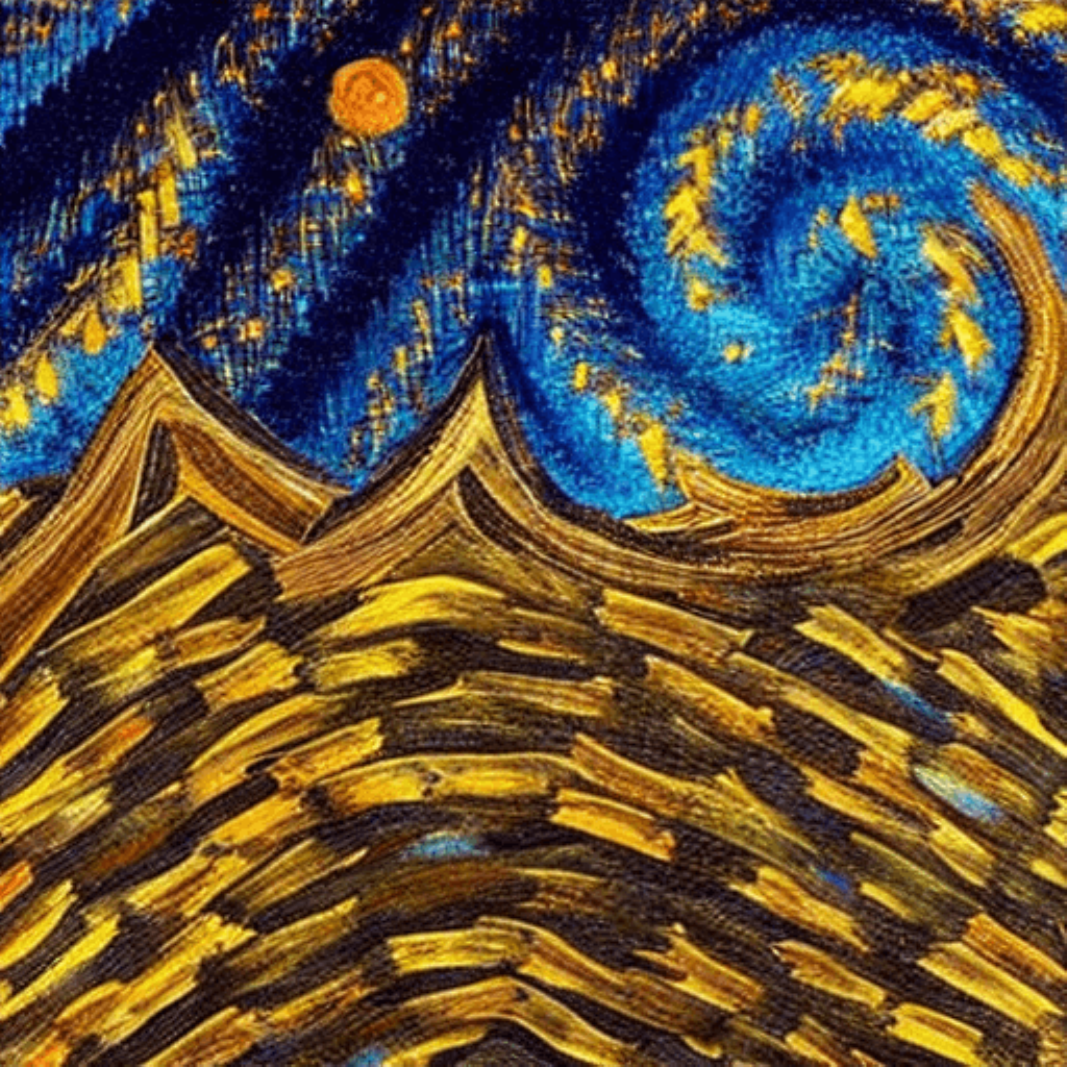}
        \caption{Ours}
        \label{fig:ours vangogh}
      \end{subfigure}      
    \caption{Qualitative results of different artistic style removal methods for eliminating the style of Van Gogh under Ring-A-Bell's attack. The prompt is "A depiction of a starry night over a quiet town, reminiscent of Van Gogh's famous painting".}
    \label{fig:vangogh attack}
\end{figure}

For artistic style removal, we provide qualitative examples in \cref{fig:vangogh attack}. The first image is generated by the original SD v1.4 without any attack, and the following images are under Ring-A-Bell's attack~\cite{tsai2023ringabell}. Recall that UCE is the second-best artistic style removal method as shown in \cref{tab:lpips of artistic removal} but it falls short in robustness against red-teaming tools. Although our method employs a closed-form solution like UCE, it outperforms UCE in robustness. ESD, CA and our method perform similarly well. More results can be found in the appendix.

\subsection{Model Editing Duration} \label{sec: duration}
\begin{table}[tb]
\centering
\resizebox{0.55\linewidth}{!}{
\begin{tabular}{lcccccc}
\toprule
                     & ESD   & UCE  & SA    & CA  & Ours \\ \midrule
Modification (\%) & 94.65 & 2.23 & 94.65 & 100 & 2.23 \\
Duration (s) & 3720 & 1.2 & - & 1400 & 3.4 \\ \bottomrule
\end{tabular}
}
\caption{Percentage of parameter modification and model editing duration for different methods. Our method and UCE are significantly ahead of other methods.}
\label{tab:effciency}
\end{table}
To demonstrate the efficiency of different methods, we measured the percentage of parameter modification and editing duration on an RTX 3090 for each method, as shown in \cref{tab:effciency}. We excluded SLD from the analysis since it operates at inference time rather than modifying the model's weights which can be easily bypassed under open-source conditions. Additionally, we don't include the duration of SA, as it involves generating 5000 images, calculating the Fisher Information Matrix and fine-tuning, which makes it exceptionally slow.

Based on \cref{tab:effciency} and \cref{tab:metrics of erasures}, our method achieves the best concept erasure effect in an extremely short time of only three seconds. Our method and UCE modify the lowest percentage of parameters with a closed-form solution, resulting in the shortest editing durations. Despite similar durations, our method significantly outperforms UCE in removal effectiveness. Conversely, CA, ESD and SA modify a high percentage of parameters with more time but achieve less impressive removal results.

\subsection{Ablation Study}
We conduct experiments to elucidate the impact of our derived embedding among different epochs and the effectiveness of the regularization term.

\subsubsection{Effect of Derived Embeddings among Epochs}

\begin{wrapfigure}{tb}{0.65\linewidth}
  \centering
    \captionsetup[subfigure]{labelformat=empty}
  \begin{subfigure}{0.19\linewidth}
    \includegraphics[width=\linewidth]{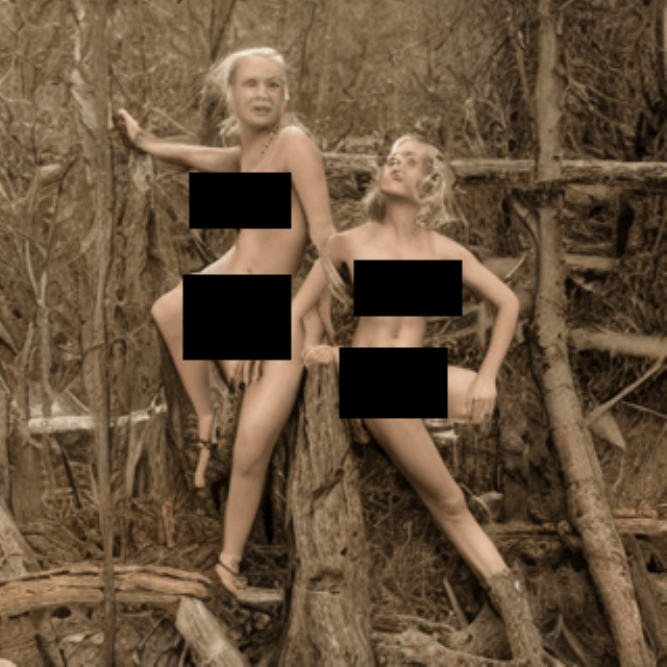}
    \caption{Epoch 0}
  \end{subfigure}%
  \begin{subfigure}{0.19\linewidth}
    \includegraphics[width=\linewidth]{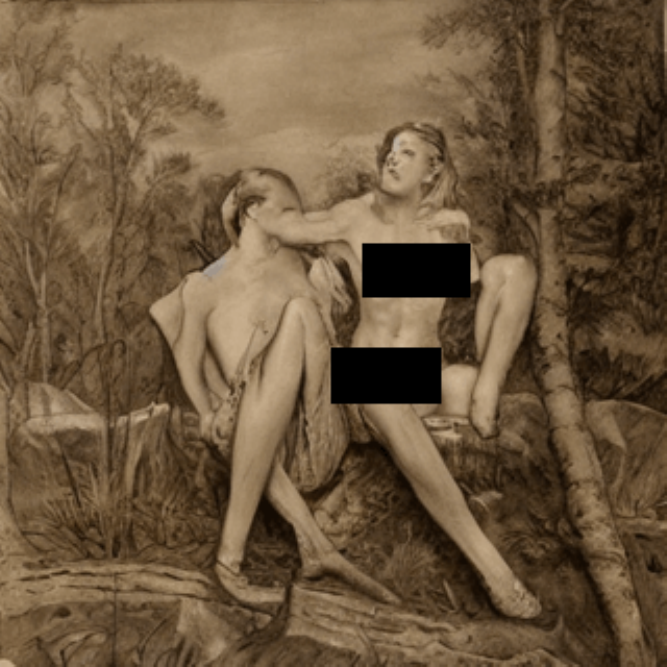}
    \caption{Epoch 1}
  \end{subfigure}%
  \begin{subfigure}{0.19\linewidth}
    \includegraphics[width=\linewidth]{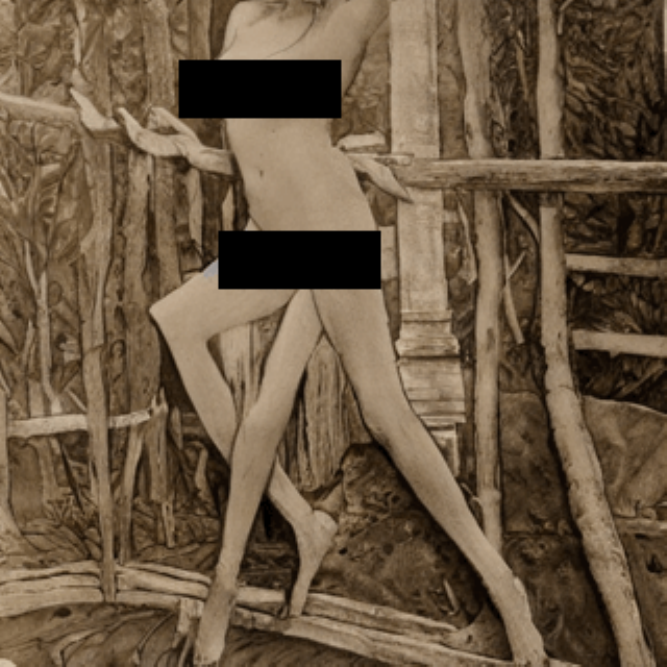}
    \caption{Epoch 2}
  \end{subfigure}%
  \begin{subfigure}{0.19\linewidth}
    \includegraphics[width=\linewidth]{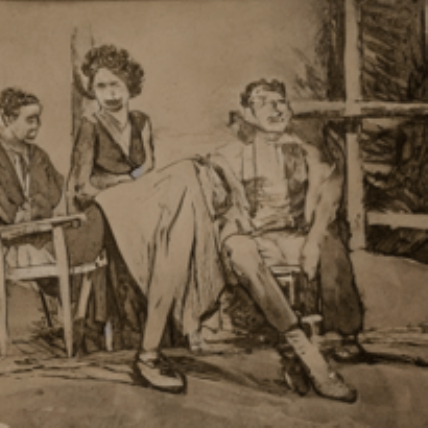}
    \caption{Epoch 3}
  \end{subfigure}%
  \begin{subfigure}{0.19\linewidth}
    \includegraphics[width=\linewidth]{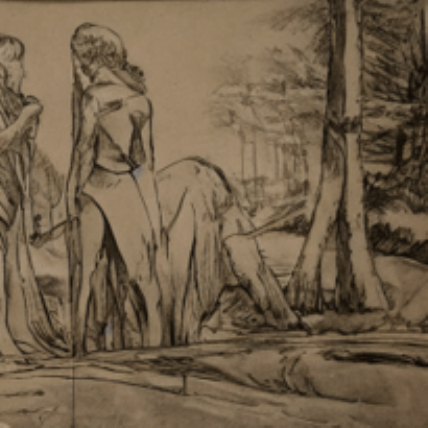}
    \caption{Epoch 4}
  \end{subfigure}%
\caption{There is no nudity information from epoch 3 onward; hence, we select the checkpoint after epoch 2 to avoid damaging the model's ability.}
  \label{fig:emb2imgs 10ep}
\end{wrapfigure}

We conduct an experiment to expound the impact of our derived embedding. We perform "nudity" erasure for 5 epochs using \cref{algo:RECE}. In each epoch, we derive a distinct embedding that represents "nudity". Before the erasure of each epoch, we generate an image using the embedding to test its degree of nudity information, as presented in \cref{fig:emb2imgs 10ep}. Images from epoch 0 to epoch 2 contain nude body parts, indicating that our derived embeddings successfully reveal potential nudity information in the model. Specifically, we opt for the checkpoint after epoch 2, as images from epoch 3 to 4 lack nudity information. Actually, erasing such "not so nude" embeddings in epoch 3-4 would impair the model's normal generation ability, which is an unworthy trade-off.

\subsubsection{Effect of Regularization Coefficient}
\begin{figure}[tb]
    \centering
    \includegraphics[width=0.85\linewidth]{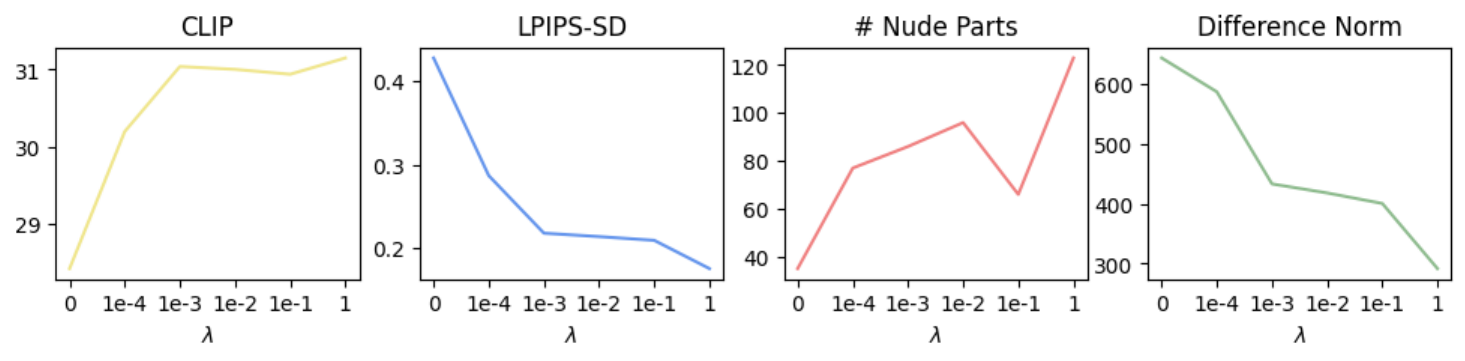}
    \caption{Ablation study on the impact of the regularization term on model performance.}
    \label{fig:ablation study}
\end{figure}
We conduct an experiment to assess the influence of our regularization term. Specifically, we select different regularization coefficients $\lambda$ in \cref{eq:c prime}, which divide the interval $[0,1]$ into five parts. The results, presented in \cref{fig:ablation study}, include CLIP-score and LPIPS against the original SD on the COCO-30k validation subset. As $\lambda$ increases, CLIP score shows an upward trend while LPIPS and the difference between new and old parameters show a downward trend. This indicates the role of the regularization term in preserving the model's ability for unerased content. Furthermore, we recorded the number of nude parts on the I2P benchmark, presented in the third column of \cref{fig:ablation study}. However, the number of nude parts doesn't strictly increase as $\lambda$ increases, which is counterintuitive. Although the purpose of the regularization term in \cref{eq:reg} is to preserve the model's generation capability, maintaining this capability does not always affect the erasure effect.
 
\section{Conclusion}
We propose a novel approach for reliably and efficiently erasing specific concepts from Text-to-Image (T2I) diffusion models. Our approach only modifies the cross-attention K\&V matrices of U-Net, constituting a mere $2.23\%$ of parameters. While previous methods also edited cross-attention modules, they still exhibited the ability to generate inappropriate images. To tackle this challenge, we derive and erase new embeddings that can represent target concepts within unlearned models. To mitigate the impact on unrelated concepts, a regularization term is introduced during the erasure process. All the above techniques are formulated in closed-form, facilitating rapid editing. This enables the execution of "derive-erase" across multiple epochs, ensuring thorough and robust erasure. Extensive experiments validate the effectiveness of our approach in erasing artistic styles, unsafe contents and object classes. Furthermore, we recorded editing durations to underscore the efficiency of our method and evaluated the robustness against red-teaming tools. We believe our RECE has the potential to empower T2I providers in effectively removing undesired concepts, thereby fostering the development of a safer AI community.

\section*{Acknowledgements}
This project was supported by National Key R\&D Program of China (No. 2021ZD0112804).

%
%
\bibliographystyle{splncs04}
\bibliography{egbib}

\begin{thebibliography}{10}
\providecommand{\url}[1]{\texttt{#1}}
\providecommand{\urlprefix}{URL }
\providecommand{\doi}[1]{https://doi.org/#1}

\bibitem{ck2022attacking}
Chen, K., Wei, Z., Chen, J., Wu, Z., Jiang, Y.G.: Attacking video recognition models with bullet-screen comments. In: Proceedings of the AAAI Conference on Artificial Intelligence. vol.~36, pp. 312--320 (2022)

\bibitem{ck2023gcma}
Chen, K., Wei, Z., Chen, J., Wu, Z., Jiang, Y.G.: Gcma: Generative cross-modal transferable adversarial attacks from images to videos. In: Proceedings of the 31st ACM International Conference on Multimedia. pp. 698--708 (2023)

\bibitem{chin2023p4d}
Chin, Z.Y., Jiang, C.M., Huang, C.C., Chen, P.Y., Chiu, W.C.: Prompting4debugging: Red-teaming text-to-image diffusion models by finding problematic prompts. arXiv preprint arXiv:2309.06135  (2023)

\bibitem{dhariwal2021diffusionbeatgans}
Dhariwal, P., Nichol, A.: Diffusion models beat gans on image synthesis. Advances in neural information processing systems  \textbf{34},  8780--8794 (2021)

\bibitem{doersch2016vae}
Doersch, C.: Tutorial on variational autoencoders. arXiv preprint arXiv:1606.05908  (2016)

\bibitem{gandikota2023erasing}
Gandikota, R., Materzynska, J., Fiotto-Kaufman, J., Bau, D.: Erasing concepts from diffusion models. arXiv preprint arXiv:2303.07345  (2023)

\bibitem{gandikota2024uce}
Gandikota, R., Orgad, H., Belinkov, Y., Materzy{\'n}ska, J., Bau, D.: Unified concept editing in diffusion models. In: Proceedings of the IEEE/CVF Winter Conference on Applications of Computer Vision. pp. 5111--5120 (2024)

\bibitem{heng2024selective}
Heng, A., Soh, H.: Selective amnesia: A continual learning approach to forgetting in deep generative models. Advances in Neural Information Processing Systems  \textbf{36} (2024)

\bibitem{hessel2021clipscore}
Hessel, J., Holtzman, A., Forbes, M., Bras, R.L., Choi, Y.: Clipscore: A reference-free evaluation metric for image captioning. arXiv preprint arXiv:2104.08718  (2021)

\bibitem{ho2020ddpm}
Ho, J., Jain, A., Abbeel, P.: Denoising diffusion probabilistic models. Advances in neural information processing systems  \textbf{33},  6840--6851 (2020)

\bibitem{hunter2023aiporn}
Hunter, T.: Ai porn is easy to make now. for women, that’s a nightmare. (2 2023)

\bibitem{kawar2023imagic}
Kawar, B., Zada, S., Lang, O., Tov, O., Chang, H., Dekel, T., Mosseri, I., Irani, M.: Imagic: Text-based real image editing with diffusion models. In: Proceedings of the IEEE/CVF Conference on Computer Vision and Pattern Recognition. pp. 6007--6017 (2023)

\bibitem{kumari2023ablating}
Kumari, N., Zhang, B., Wang, S.Y., Shechtman, E., Zhang, R., Zhu, J.Y.: Ablating concepts in text-to-image diffusion models. In: Proceedings of the IEEE/CVF International Conference on Computer Vision. pp. 22691--22702 (2023)

\bibitem{li2024safegen}
Li, X., Yang, Y., Deng, J., Yan, C., Chen, Y., Ji, X., Xu, W.: Safegen: Mitigating unsafe content generation in text-to-image models. arXiv preprint arXiv:2404.06666  (2024)

\bibitem{lin2014mscoco}
Lin, T.Y., Maire, M., Belongie, S., Hays, J., Perona, P., Ramanan, D., Doll{\'a}r, P., Zitnick, C.L.: Microsoft coco: Common objects in context. In: Computer Vision--ECCV 2014: 13th European Conference, Zurich, Switzerland, September 6-12, 2014, Proceedings, Part V 13. pp. 740--755. Springer (2014)

\bibitem{nichol2022glide}
Nichol, A.Q., Dhariwal, P., Ramesh, A., Shyam, P., Mishkin, P., Mcgrew, B., Sutskever, I., Chen, M.: Glide: Towards photorealistic image generation and editing with text-guided diffusion models. In: International Conference on Machine Learning. pp. 16784--16804. PMLR (2022)

\bibitem{orgad2023TIME}
Orgad, H., Kawar, B., Belinkov, Y.: Editing implicit assumptions in text-to-image diffusion models. In: Proceedings of the IEEE/CVF International Conference on Computer Vision. pp. 7053--7061 (2023)

\bibitem{parmar2022cleanfid}
Parmar, G., Zhang, R., Zhu, J.Y.: On aliased resizing and surprising subtleties in gan evaluation. In: Proceedings of the IEEE/CVF Conference on Computer Vision and Pattern Recognition. pp. 11410--11420 (2022)

\bibitem{praneeth2019nudenet}
Praneeth, B.: Nudenet: Neural nets for nudity classification, detection and selective censoring (2019)

\bibitem{radford2021clip}
Radford, A., Kim, J.W., Hallacy, C., Ramesh, A., Goh, G., Agarwal, S., Sastry, G., Askell, A., Mishkin, P., Clark, J., et~al.: Learning transferable visual models from natural language supervision. In: International conference on machine learning. pp. 8748--8763. PMLR (2021)

\bibitem{ramesh2022hierarchical}
Ramesh, A., Dhariwal, P., Nichol, A., Chu, C., Chen, M.: Hierarchical text-conditional image generation with clip latents. arXiv preprint arXiv:2204.06125  \textbf{1}(2), ~3 (2022)

\bibitem{rando2022red}
Rando, J., Paleka, D., Lindner, D., Heim, L., Tram{\`e}r, F.: Red-teaming the stable diffusion safety filter. arXiv preprint arXiv:2210.04610  (2022)

\bibitem{Rombach2022sd2}
Rombach, R.: Stable diffusion 2.0 release (November 2022)

\bibitem{smithmano2022}
Rombach, R.: Tutorial: How to remove the safety filter in 5 seconds (August 2022)

\bibitem{rombach2022ldm}
Rombach, R., Blattmann, A., Lorenz, D., Esser, P., Ommer, B.: High-resolution image synthesis with latent diffusion models. In: Proceedings of the IEEE/CVF conference on computer vision and pattern recognition. pp. 10684--10695 (2022)

\bibitem{RombachEsser2022sdmodelcard}
Rombach, R., Esser, P.: Stable diffusion v1-4 model card. Model Card (2022)

\bibitem{ronneberger2015u}
Ronneberger, O., Fischer, P., Brox, T.: U-net: Convolutional networks for biomedical image segmentation. In: Medical image computing and computer-assisted intervention--MICCAI 2015: 18th international conference, Munich, Germany, October 5-9, 2015, proceedings, part III 18. pp. 234--241. Springer (2015)

\bibitem{ruiz2023dreambooth}
Ruiz, N., Li, Y., Jampani, V., Pritch, Y., Rubinstein, M., Aberman, K.: Dreambooth: Fine tuning text-to-image diffusion models for subject-driven generation. In: Proceedings of the IEEE/CVF Conference on Computer Vision and Pattern Recognition. pp. 22500--22510 (2023)

\bibitem{saharia2022imagen}
Saharia, C., Chan, W., Saxena, S., Li, L., Whang, J., Denton, E.L., Ghasemipour, K., Gontijo~Lopes, R., Karagol~Ayan, B., Salimans, T., et~al.: Photorealistic text-to-image diffusion models with deep language understanding. Advances in Neural Information Processing Systems  \textbf{35},  36479--36494 (2022)

\bibitem{schramowski2023i2p}
Schramowski, P., Brack, M., Deiseroth, B., Kersting, K.: Safe latent diffusion: Mitigating inappropriate degeneration in diffusion models. In: Proceedings of the IEEE/CVF Conference on Computer Vision and Pattern Recognition. pp. 22522--22531 (2023)

\bibitem{setty2023aiart}
Setty, R.: Ai art generators hit with copyright suit over artists’ images (1 2023)

\bibitem{stabilityai2022sd21modelcard}
StabilityAI: Stable diffusion 2.1 model card. Model Card (2022)

\bibitem{tsai2023ringabell}
Tsai, Y.L., Hsu, C.Y., Xie, C., Lin, C.H., Chen, J.Y., Li, B., Chen, P.Y., Yu, C.M., Huang, C.Y.: Ring-a-bell! how reliable are concept removal methods for diffusion models? arXiv preprint arXiv:2310.10012  (2023)

\bibitem{vaswani2017attention}
Vaswani, A., Shazeer, N., Parmar, N., Uszkoreit, J., Jones, L., Gomez, A.N., Kaiser, {\L}., Polosukhin, I.: Attention is all you need. Advances in neural information processing systems  \textbf{30} (2017)

\bibitem{wzp2023towards}
Wei, Z., Chen, J., Goldblum, M., Wu, Z., Goldstein, T., Jiang, Y.G., Davis, L.S.: Towards transferable adversarial attacks on image and video transformers. IEEE Transactions on Image Processing  \textbf{32},  6346--6358 (2023)

\bibitem{wzp2023adaptive}
Wei, Z., Chen, J., Wu, Z., Jiang, Y.G.: Adaptive cross-modal transferable adversarial attacks from images to videos. IEEE Transactions on Pattern Analysis and Machine Intelligence  (2023)

\bibitem{yang2023diffusionsurvey}
Yang, L., Zhang, Z., Song, Y., Hong, S., Xu, R., Zhao, Y., Zhang, W., Cui, B., Yang, M.H.: Diffusion models: A comprehensive survey of methods and applications. ACM Computing Surveys  \textbf{56}(4),  1--39 (2023)

\bibitem{zhang2018lpips}
Zhang, R., Isola, P., Efros, A.A., Shechtman, E., Wang, O.: The unreasonable effectiveness of deep features as a perceptual metric. In: Proceedings of the IEEE conference on computer vision and pattern recognition. pp. 586--595 (2018)

\bibitem{zhang2023unlearndiff}
Zhang, Y., Jia, J., Chen, X., Chen, A., Zhang, Y., Liu, J., Ding, K., Liu, S.: To generate or not? safety-driven unlearned diffusion models are still easy to generate unsafe images... for now. arXiv preprint arXiv:2310.11868  (2023)

\end{thebibliography}

\newpage

\section*{Appendix}

\appendix

\setcounter{equation}{0}
\renewcommand{\theequation}{A.\arabic{equation}}
\setcounter{figure}{0}
\renewcommand{\thefigure}{A.\arabic{figure}}

\section{Deriving the New Embedding} \label{app:derive c prime}

In this section, we present a detailed derivation of the closed-form new embedding. 

Let $W^\mathrm{old}$ denote the projection matrices of the original U-Net before UCE's editing, $W^\mathrm{new}$ represent the projection matrices after UCE's editing, $c$ denote the embedding of "nudity", and $c^\prime$ signify our derived embedding.
If we can find a $c^\prime$ such that $W^\mathrm{new}c^\prime$ closely resembles $W^\mathrm{old}c$, then $c^\prime$ can guide the edited model to generate nude images like how $c$ guides the original model. Specifically, the objective function is formulated as follows:

\begin{equation}
\min_{c^{\prime}}\sum_{i}\|W_{i}^\mathrm{new}c^{\prime}-W_{i}^\mathrm{old}c\|_{2}^{2}+\lambda \|c^\prime\|_2^2,
\label{eq:c prime app}
\end{equation}
where $W_i$ denotes K/V cross-attention projection matrices of U-Net, $\lambda$ is a hyper-parameter and $\|c^\prime\|_2^2$ is a regularization term which will be proved in \cref{app:prove reg} below. The square of the 2-norm is convex, and linear transformation maintains its convexity. Therefore, \cref{eq:c prime app} represents a convex function, possessing a unique global minimum solution $c^\prime$. This solution can be obtained by setting the gradient of \cref{eq:c prime app} to zero:
$$
\frac{\partial L}{\partial c^{\prime}}=\sum_i2W_i^{\mathrm{new}^T}(W_i^\mathrm{new}c^{\prime}-W_i^\mathrm{old}c)+2\lambda c^\prime =0
$$
$$
\sum_i(W_i^{\mathrm{new}^{T}}W_i^\mathrm{new}+\lambda I)c^{\prime}=\sum_iW_i^{\mathrm{new}^T} W_i^\mathrm{old}c
$$
$$
c^{\prime}=\left(\lambda I+\sum_{i}W_{i}^{\mathrm{new}^{T}}W_{i}^{\mathrm{new}}\right)^{-1}\left(\sum_{i}W_{i}^{\mathrm{new}^{T}}W_{i}^{\mathrm{old}}\right)c.
$$
Here we obtain the new embedding $c^{\prime}$ which can guide the edited model to generate nude images, whereas the word "nudity" cannot. This indicates that $c^{\prime}$ serves as the true representation of $c$ within the edited model. 

\section{Proof of Regularization Term} \label{app:prove reg}
\setcounter{equation}{0}
\renewcommand{\theequation}{B.\arabic{equation}}
In this section, we present a detailed proof of our regularization term which can protect the model's ability.

Let $W^\mathrm{new1}$ denote the projection matrices after the last epoch's modification, $W^\mathrm{new2}$ denote the projection matrices after the current epoch, and $d$ denote an unrelated concept's embedding. To partially preserve the model's performance entails minimizing the impact on unrelated concepts as much as possible. Consequently, we define our regularization objective function as follows:
\begin{equation}
    \min_{W^\mathrm{new2}} \| W^\mathrm{new2}d-W^\mathrm{new1}d\|_2^2.
    \label{eq:reg app}
\end{equation}
Let's denote $c_i$ for concepts to be erased, $c_j$ for concepts to be preserved and $c^{\prime}$ for derived embeddings. $W^\mathrm{new2}$ is influenced by $c^\prime$ because it is derived after erasing $c^\prime$ from $W^\mathrm{new1}$. Our proof will utilize the submultiplicative property of the matrix norm\footnote{\url{https://en.wikipedia.org/wiki/Matrix_norm}}:
$$\begin{aligned}
\|Ax\|_2\leq\|A\|_F\|x\|_2\\
\|AB\|_F\leq\|A\|_F\|B\|_F,
\end{aligned}$$
where $A$ and $B$ are matrices and $x$ is a vector.

Now, we aim to find the minimum of \cref{eq:reg app}:
\begin{equation*}
    \begin{aligned}
        F&=\|W^\mathrm{new2}d-W^\mathrm{new1}d\|_2^2\\
        &\leq\|W^\mathrm{new2}-W^\mathrm{new1}\|_F^2\|d\|_2^2
    \end{aligned}
\end{equation*}
Note that in erasing process, we have:\\
$$W^\mathrm{new2} = W^\mathrm{new1}\left(\sum_{i \in E}c_i^* c_i^{\prime ^T}+\sum_{j\in P}c_j c_j^T\right)\left(\sum_{i\in E}c_i^\prime c_i^{\prime^T}+\sum_{j\in P}c_jc_j^T\right)^{-1}$$
Let $F_1=\|W^\mathrm{new2}-W^\mathrm{new1}\|_F^2$, then according to the submultiplicative property we have:
\begin{equation*}
    \begin{aligned}
        F_1&=\|W^\mathrm{new2}-W^\mathrm{new1}\|_F^2\\
        &=\|W^\mathrm{new1}\left(\sum_{i \in E}c_i^* c_i^{\prime ^T}+\sum_{j\in P}c_j c_j^T\right)\left(\sum_{i\in E}c_i^\prime c_i^{\prime^T}+\sum_{j\in P}c_jc_j^T\right)^{-1}-W^\mathrm{new1}\|_F^2\\
        &\leq\|W^\mathrm{new1}\|_F^2\|\left(\sum c_i^*c_i^{\prime^T}+\sum c_jc_j^T\right)\left(\sum c_i^\prime c_i^{\prime^T}+\sum c_jc_j^T\right)^{-1}-I\|_F^2,
    \end{aligned}
\end{equation*}
where $c_i^*$ denote the corresponding destination embedding. Let's define:
\begin{equation*}\begin{aligned}
F_2&=\left\|\left(\sum c_i^*c_i^{\prime^T}+\sum c_jc_j^T\right)\left(\sum c_i^{\prime}c_i^{\prime^T}+\sum c_jc_j^T\right)^{-1}-I\right\|_{F}^2\\
&U=\quad\sum c_i^{\prime}c_i^{\prime^T}+\sum c_jc_j^T
\end{aligned}\end{equation*}
Then we have:
\begin{equation*}\begin{aligned}
F_2&=\|(\sum c_i^*c_i^{\prime^T}+\sum c_jc_j^T)U^{-1}-UU^{-1}\|_F^2\\
&\leq\|\sum c_i^*c_i^{\prime^T}+\sum c_jc_j^T-U\|_F^2\|U^{-1}\|_F^2
\end{aligned}\end{equation*}
Considering only the first term:
\begin{equation*}
\begin{aligned}
F_3&=\|\sum c_i^*c_i^{\prime^T}+\sum c_jc_j^T-U\|_F^2\\
&=\|\left(\sum c_i^*c_i^{\prime^T}+\sum c_jc_j^T\right)-\left(\sum c_i^{\prime}c_i^{\prime^T}+\sum c_jc_j^T\right)\|_F^2\\
&=\|\sum c_i^*c_i^{\prime^T}-\sum c_i^{\prime}c_i^{\prime^T}\|_F^2\\
&=\|\sum\left(c_i^*-c_i^{\prime}\right)c_i^{\prime^T}\|_F^2\\
&\leq\sum\|\left(c_i^*-c_i^{\prime}\right)c_i^{\prime^T}\|_F^2
\end{aligned}
\end{equation*}
If we set $c^\prime = c^*\ \mathrm{or}\ \mathbf{0}$, then $F_3$ achieves its minimum of $0$, and consequently, \cref{eq:reg app} will also reach its minimum of $0$. It's important to note that $c^*$ represents the destination embedding in UCE, so it's possible that we don't know the destination concept when applied UCE and only have the parameters after UCE erases. Therefore, we use $\mathbf{0}$ as the argmin. In conclusion, the closer $c^\prime$ is to $\mathbf{0}$, the less it will affect the model's performance.

\section{Object Removal}

\setcounter{table}{0}
\renewcommand{\thetable}{C.\arabic{table}}

\begin{table}[tb]
\centering
\resizebox{0.9\linewidth}{!}{
\begin{tabular}{lcccccccc}
\toprule
\multirow{2}{*}{Class Name} & \multicolumn{4}{c}{Accuracy of Erased Class (\%) $\downarrow$} & \multicolumn{4}{c}{Accuracy of Other Classes (\%) $\uparrow$} \\ \cmidrule(l){2-5} \cmidrule(l){6-9} 
& SD & ESD-u & UCE & Ours & SD & ESD-u & UCE & Ours \\
\midrule
Cassette Player & 15.6 & 0.6 & 0.0 & 0.0 & 85.1 & 64.5 & 90.3 & 90.3 \\
Chain Saw & 66.0 & 6.0 & 0.0 & 0.0 & 79.6 & 68.2 & 76.1 & 76.1 \\
Church & 73.8 & 54.2 & 8.4 & 2.0 & 78.7 & 71.6 & 80.2 & 80.5 \\
English Springer & 92.5 & 6.2 & 0.2 & 0.0 & 76.6 & 62.6 & 78.9 & 77.8 \\
French Horn & 99.6 & 0.4 & 0.0 & 0.0 & 75.8 & 49.4 & 77.0 & 77.0 \\
Garbage Truck & 85.4 & 10.4 & 14.8 & 0.0 & 77.4 & 51.5 & 78.7 & 65.4 \\
Gas Pump & 75.4 & 8.6 & 0.0 & 0.0 & 78.5 & 66.5 & 80.7 & 80.7 \\
Golf Ball & 97.4 & 5.8 & 0.8 & 0.0 & 76.1 & 65.6 & 79.0 & 79.0 \\
Parachute & 98.0 & 23.8 & 1.4 & 0.9 & 76.0 & 65.4 & 77.4 & 79.1 \\
Tench & 78.4 & 9.6 & 0.0 & 0.0 & 78.2 & 66.6 & 79.3 & 79.3 \\
\midrule
Average & 78.2 & 12.6 & \underline{2.6} & \textbf{0.3} & 78.2 & 63.2 & \textbf{79.8} & \underline{78.5} \\
\bottomrule
\end{tabular}
}
\caption{Comparison of classification accuracy for object removal methods. \textbf{Bold}: best. \underline{Underline}: second-best.}
\label{tab:app-object erasures}
\end{table}

In this section, we investigate the method's effectiveness in erasing entire object classes from the model. 

We focus our comparison on ESD and UCE, as these are the only methods that have conducted object removal experiments in their respective papers. Following the experimental setup of UCE, we conduct experiments on erasing Imagenette classes, a subset of Imagenet classes, generating 500 images per class. We perform iterative erasure, validated by the accuracy of erased and unerased classes, and ultimately settle on using 1 epoch. We set the regularization coefficient $\lambda$ to $1e-3$ for the "church" and "garbage truck" classes, as these classes are challenging to erase using UCE, thus requiring stronger erasure efforts. Conversely, we set $\lambda$ to $1e-1$ for the "English Springer", "golf ball" and "parachute" classes, as these are easier to erase for UCE and therefore only necessitate light erasure. For other 5 classes where UCE accuracy has already reached 0, we discontinue further erasure with RECE.
As shown in~\cref{tab:app-object erasures}, our RECE exhibits superior erasure capability while minimizing interference with non-targeted classes.

\section{Extended Inappropriate Content Removal}

\setcounter{table}{0}
\renewcommand{\thetable}{D.\arabic{table}}

\begin{table}[tb]
\centering
\resizebox{0.65\linewidth}{!}{
\begin{tabular}{lcccccc}
\toprule
\multirow{2}{*}{Category} & \multicolumn{6}{c}{Inappropriate Proportions (\%) $\downarrow$} \\
\cmidrule(l){2-7}
& SD & ESD-u & CA & UCE & SLD-Med & Ours \\ 
\midrule
Hate & 21.2 & \textbf{3.5} & 15.6 & 10.8 & 41.1 & \underline{4.3} \\
Harassment & 19.7 & \underline{6.4} & 15.9 & 12.1 & 20.1 & \textbf{6.1} \\
Violence & 40.1 & \underline{16.7} & 31.3 & 23.3 & 19.7 & \textbf{14.2} \\
Self-harm & 35.5 & \underline{11.1} & 21.7 & 12.9 & 19.2 & \textbf{8.5} \\
Sexual & 54.5 & 16.4 & 32.7 & \underline{16.2} & 22.9 & \textbf{8.6} \\
Shocking & 42.1 & 16.1 & 30.7 & 19.2 & \underline{16.0} & \textbf{9.7} \\
Illegal Activity & 19.4 & \underline{6.3} & 13.2 & 9.8 & 20.5 & \textbf{6.1} \\ 
\midrule
Overall & 35.6 & \underline{12.2} & 24.3 & 15.6 & 20.8 & \textbf{8.5} \\
\bottomrule
\end{tabular}
}
\caption{Comparison of inappropriate proportions(\%) for different removal methods. \textbf{Bold}: best. \underline{Underline}: second-best.}
\label{tab:app-inappropriate erasures}
\end{table}
We also compare the models across larger categories of inappropriate classes. 

In the main text, we focus on erasing nudity to align with the experimental setup used in ESD's main paper, as nudity is a classical example of an inappropriate concept. In~\cref{tab:app-inappropriate erasures}, we further demonstrate the efficacy of erasing multiple sensitive concepts from I2P, including “hate, harassment, violence, suffering, humiliation, harm, suicide, sexual, nudity, bodily fluids, blood”. The $\lambda$ is set to $1e-1$ and the erasure process is performed for 2 epochs. We use the fine-tuned Q16 
classifier\footnote{\url{https://github.com/YitingQu/unsafe-diffusion}} which more accurately detects general inappropriate concepts, to present the proportions of inappropriate content across various categories in I2P. The results show that our method effectively erases these sensitive concepts.

\section{Qualitative Results}
\setcounter{figure}{0}
\renewcommand{\thefigure}{E.\arabic{figure}}

In this section, we present additional qualitative results. Images in the same row are generated with same prompts and seeds.

\begin{figure}[tb]
  \centering
    \captionsetup[subfigure]{labelformat=empty} 
  \begin{subfigure}{0.16\linewidth}
    \centering
    \includegraphics[width=0.98\linewidth]{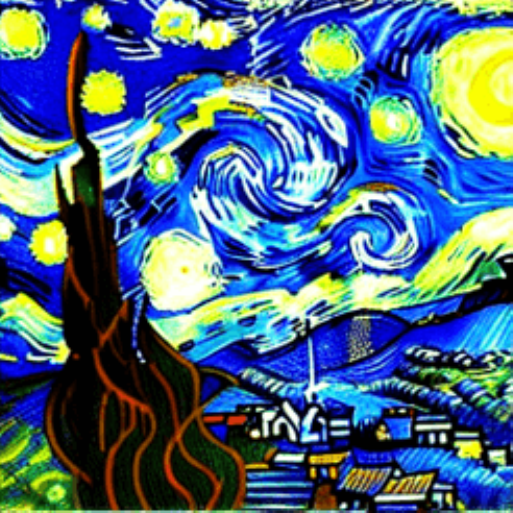}
    \includegraphics[width=0.98\linewidth]{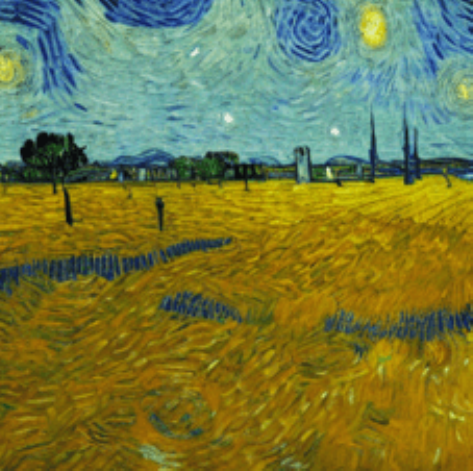}
    \caption{SD}
    \label{fig:app-sd artistic erased}
  \end{subfigure}
  \begin{subfigure}{.16\linewidth}
    \centering
    \includegraphics[width=0.98\linewidth]{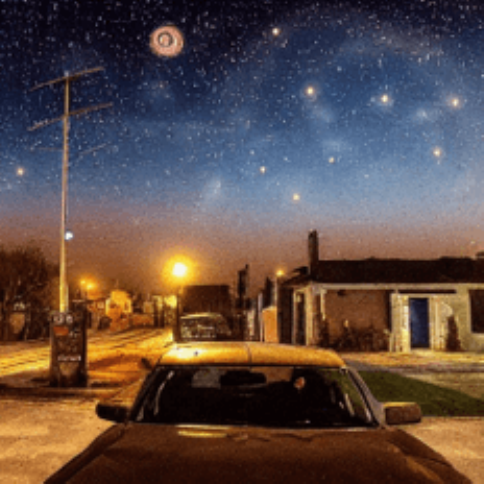}
    \includegraphics[width=0.98\linewidth]{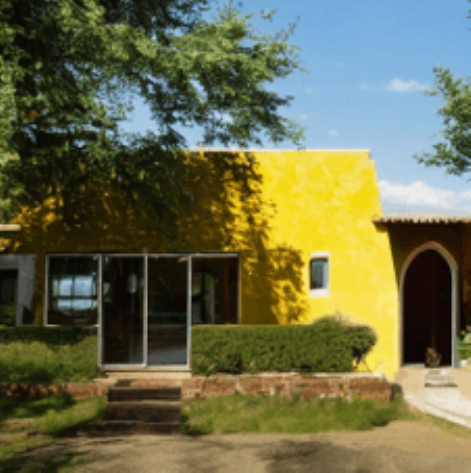}
    \caption{ESD}
    \label{fig:app-esd artistic erased}
  \end{subfigure}
    \begin{subfigure}{0.16\linewidth}
    \centering
    \includegraphics[width=0.98\linewidth]{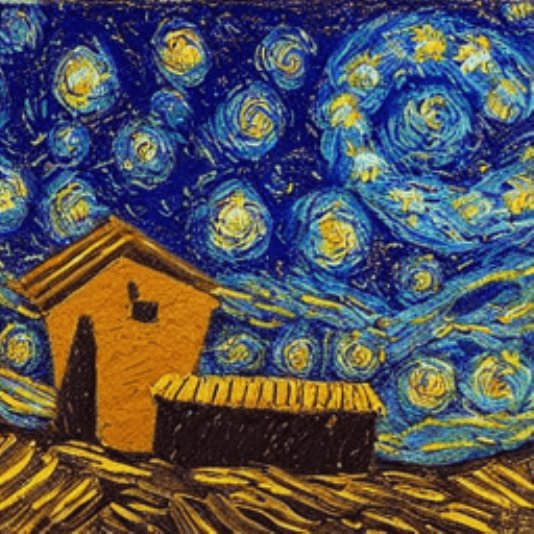}
    \includegraphics[width=0.98\linewidth]{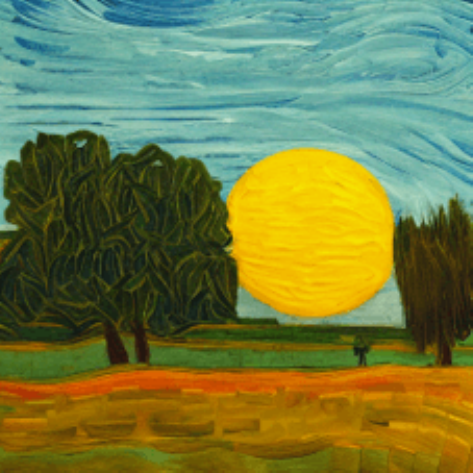}
    \caption{UCE}
    \label{fig:app-uce artistic erased}
  \end{subfigure}
    \begin{subfigure}{0.16\linewidth}
    \centering
    \includegraphics[width=0.98\linewidth]{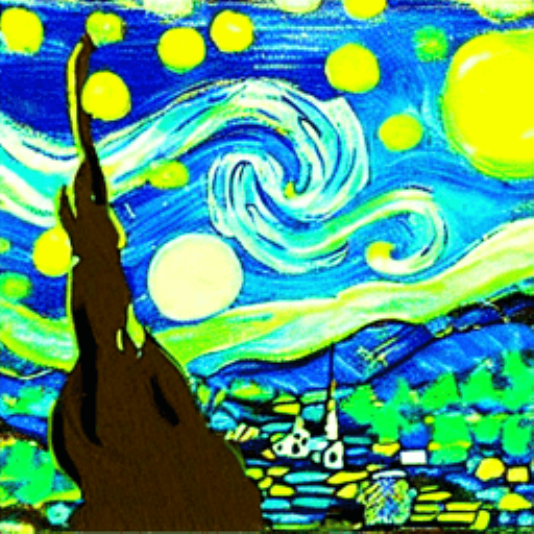}
    \includegraphics[width=0.98\linewidth]{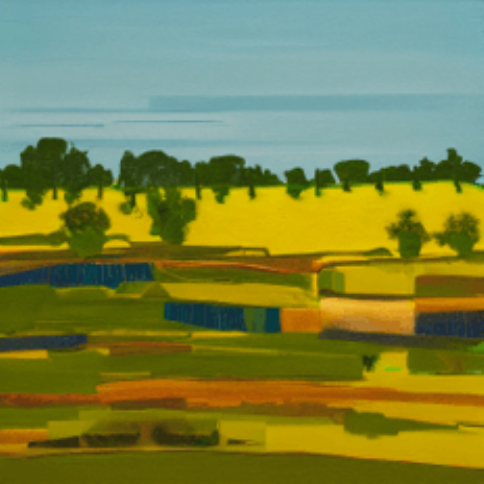}
    \caption{SLD-Medium}
    \label{fig:app-sld artistic erased}
  \end{subfigure}
    \begin{subfigure}{0.16\linewidth}
    \centering
    \includegraphics[width=0.98\linewidth]{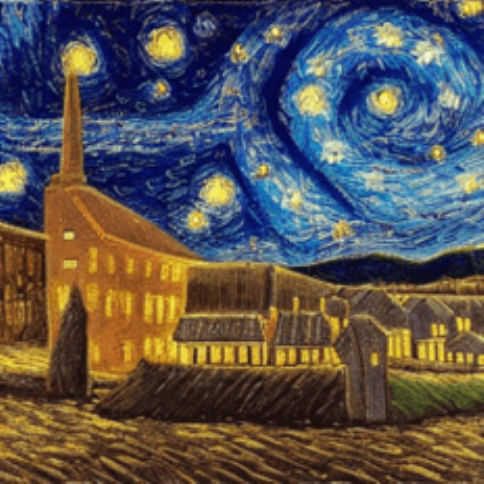}
    \includegraphics[width=0.98\linewidth]{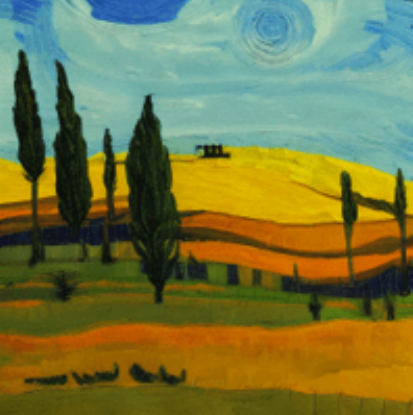}
    \caption{CA}
    \label{fig:app-ca artistic erased}
  \end{subfigure}
    \begin{subfigure}{0.16\linewidth}
    \centering
    \includegraphics[width=0.98\linewidth]{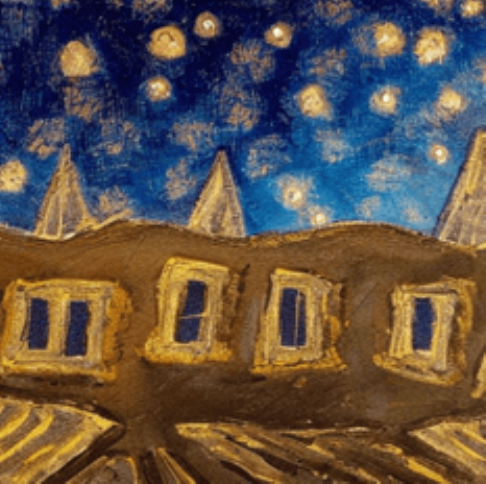}
    \includegraphics[width=0.98\linewidth]{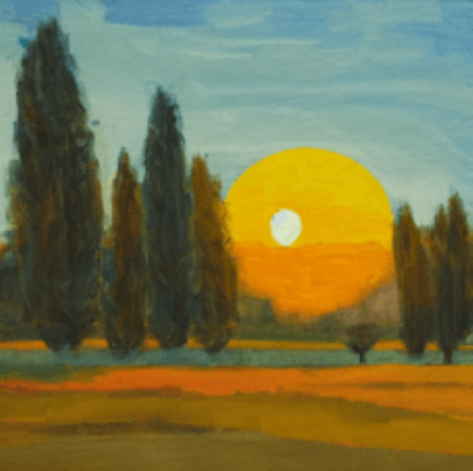}
    \caption{Ours}
    \label{fig:app-ours artistic erased}
  \end{subfigure}
  \caption{Qualitative comparison on erasing Van Gogh's style, conditioned by artist that is \textbf{erased}.}
  \label{fig:app-artistic erased}
\end{figure}

\cref{fig:app-artistic erased} shows artistic paintings generated by SD and different erased models. The prompts are "A depiction of a starry night over a quiet town, reminiscent of Van Gogh's famous painting", which should be erased. SLD struggles to remove Van Gogh's style, whereas other methods demonstrate capability.

\begin{figure}[tb]
  \centering
    \captionsetup[subfigure]{labelformat=empty} 
  \begin{subfigure}{0.16\linewidth}
    \centering
    \includegraphics[width=0.98\linewidth]{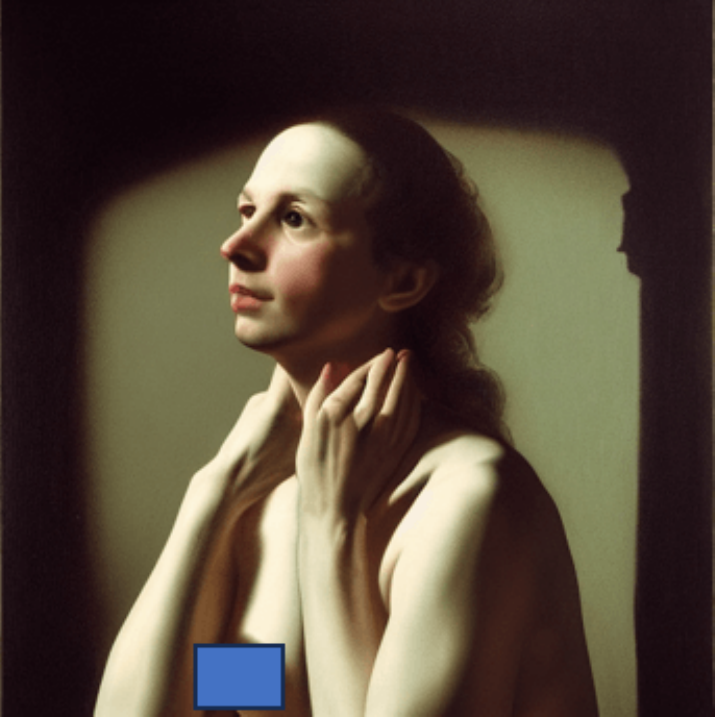}
    \includegraphics[width=0.98\linewidth]{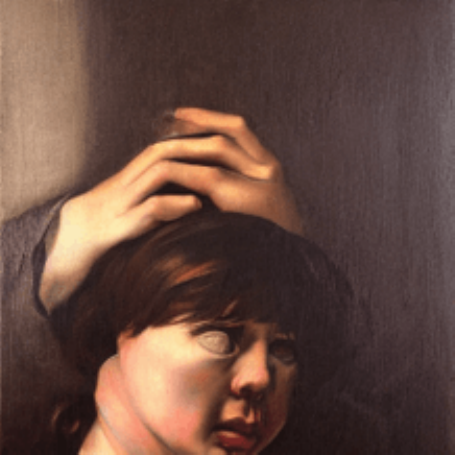}
    \caption{SD}
    \label{fig:app-sd artistic unerased}
  \end{subfigure}
  \begin{subfigure}{.16\linewidth}
    \centering
    \includegraphics[width=0.98\linewidth]{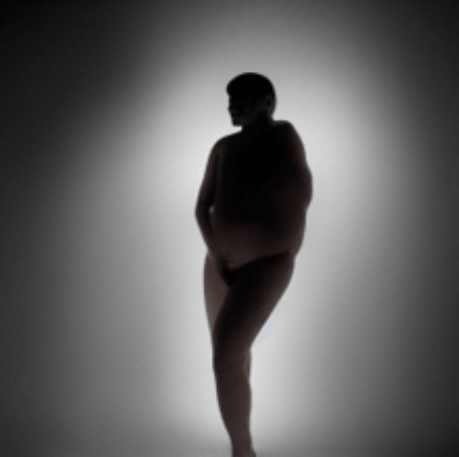}
    \includegraphics[width=0.98\linewidth]{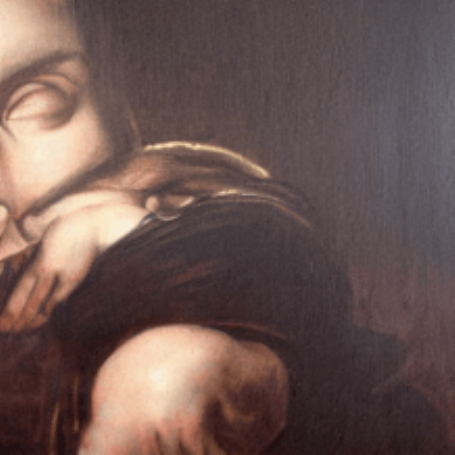}
    \caption{ESD}
    \label{fig:app-esd artistic unerased}
  \end{subfigure}
    \begin{subfigure}{0.16\linewidth}
    \centering
    \includegraphics[width=0.98\linewidth]{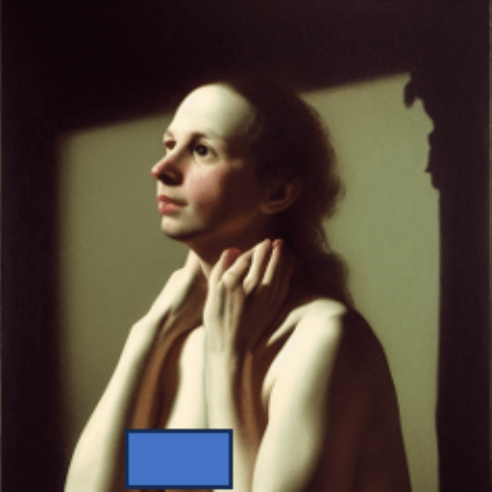}
    \includegraphics[width=0.98\linewidth]{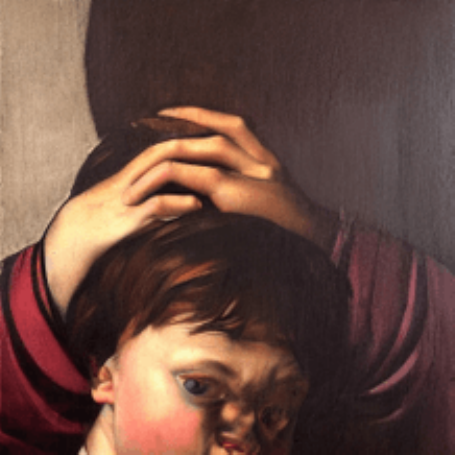}
    \caption{UCE}
    \label{fig:app-uce artistic unerased}
  \end{subfigure}
    \begin{subfigure}{0.16\linewidth}
    \centering
    \includegraphics[width=0.98\linewidth]{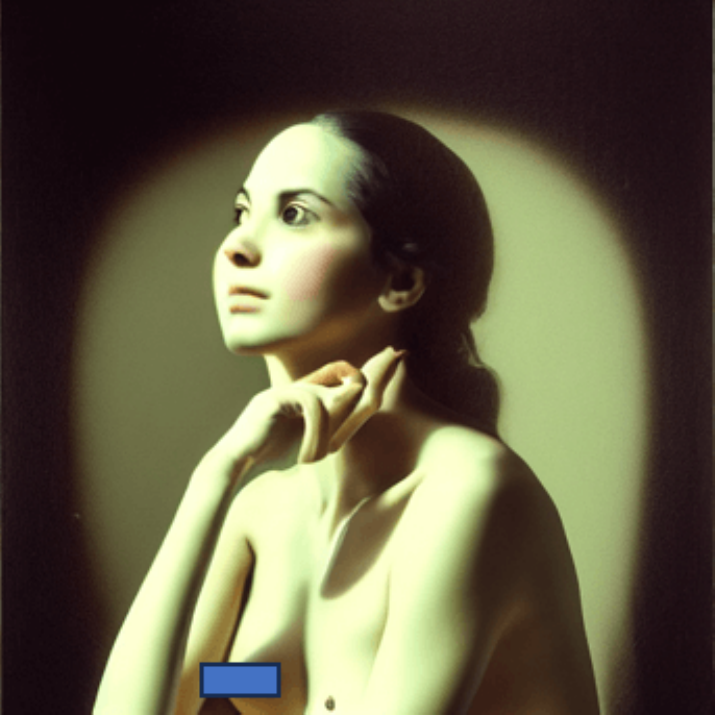}
    \includegraphics[width=0.98\linewidth]{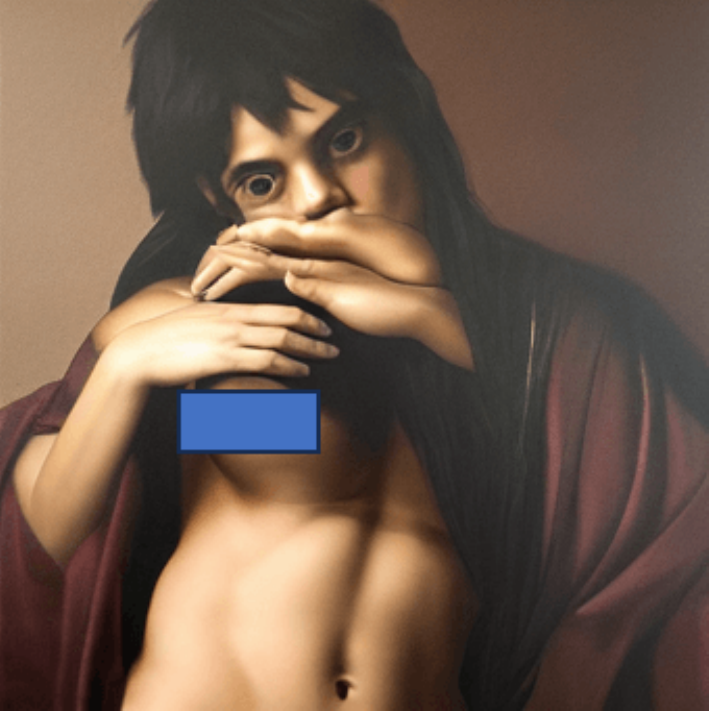}
    \caption{SLD-Medium}
    \label{fig:app-sld artistic unerased}
  \end{subfigure}
    \begin{subfigure}{0.16\linewidth}
    \centering
    \includegraphics[width=0.98\linewidth]{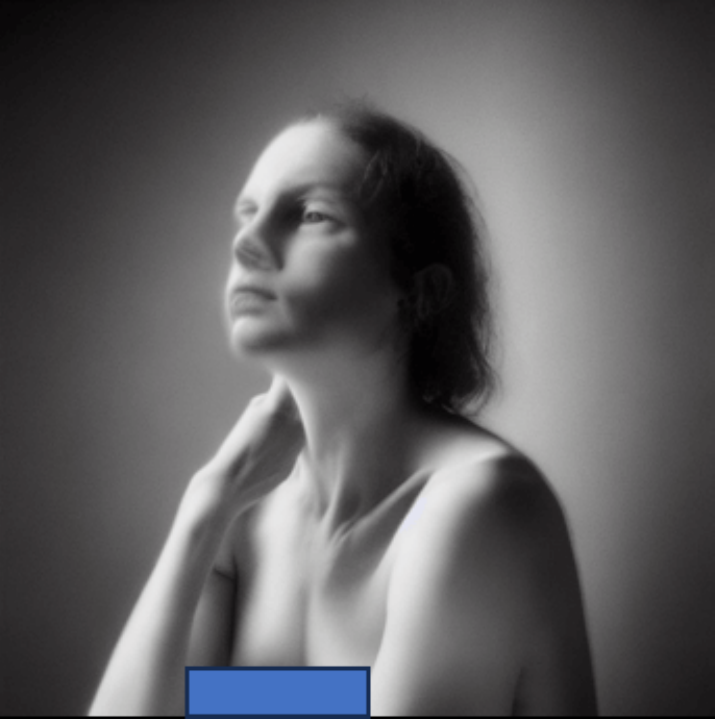}
    \includegraphics[width=0.98\linewidth]{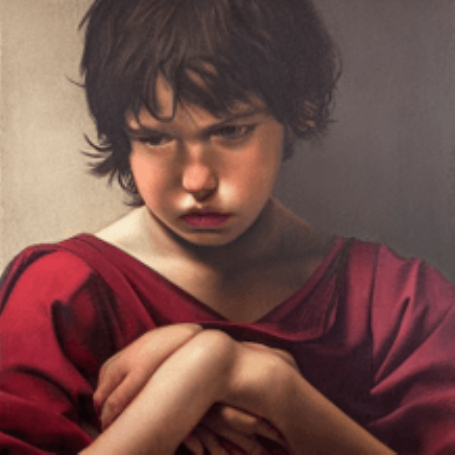}
    \caption{CA}
    \label{fig:app-ca artistic unerased}
  \end{subfigure}
    \begin{subfigure}{0.16\linewidth}
    \centering
    \includegraphics[width=0.98\linewidth]{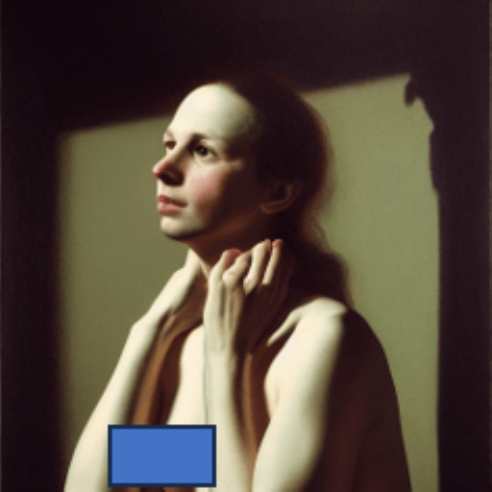}
    \includegraphics[width=0.98\linewidth]{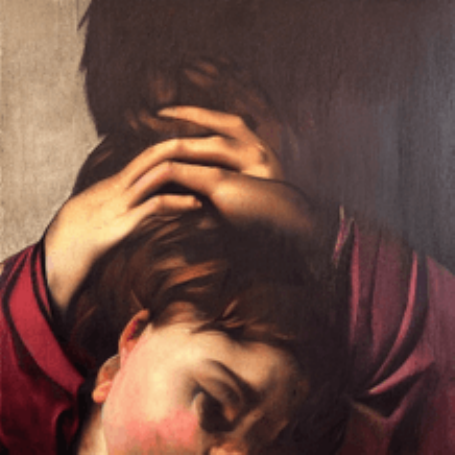}
    \caption{Ours}
    \label{fig:app-ours artistic unerased}
  \end{subfigure}
  \caption{Qualitative comparison on erasing Van Gogh's style, conditioned by artists that are \textbf{not erased}.}
  \label{fig:app-artistic unerased}
\end{figure}

\cref{fig:app-artistic unerased} shows images conditioned by prompts in which the artistic style shouldn't be erased. ESD and CA cause significant and unnecessary distortion to unerased artistic styles. Our method and UCE exert minimal influence.

\begin{figure}[tb]
  \centering
  \captionsetup[subfigure]{labelformat=empty} 
  \begin{subfigure}{0.144\linewidth}
    \centering
    \includegraphics[width=\linewidth]{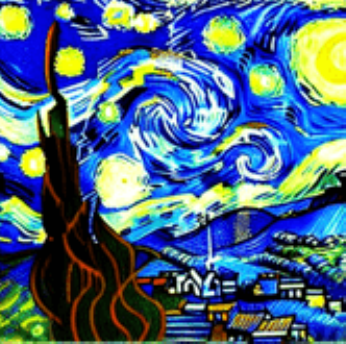}
    \caption{Original SD}
    \vspace{0.5\linewidth}
    \label{fig:app-sd artistic attack}
  \end{subfigure}
  \begin{subfigure}{.16\linewidth}
    \centering
    \begin{minipage}[c]{0.1\linewidth}
      \raggedleft
      \rotatebox{90}{\scriptsize P4D}
    \end{minipage}%
    \begin{minipage}[c]{0.9\linewidth}
      \includegraphics[width=\linewidth]{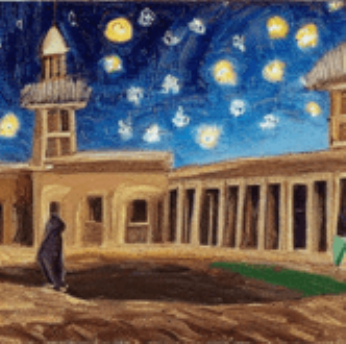}
    \end{minipage}
    \begin{minipage}[c]{0.1\linewidth}
      \raggedleft
      \rotatebox{90}{\scriptsize UnlearnDiff}
    \end{minipage}%
    \begin{minipage}[c]{0.9\linewidth}
      \includegraphics[width=\linewidth]{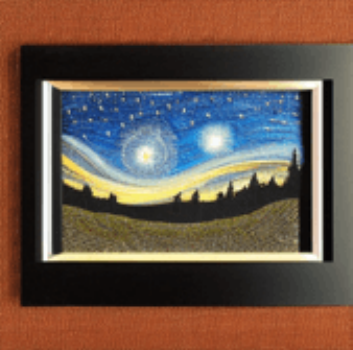}
    \end{minipage}
    \caption{ESD}
    \label{fig:app-esd artistic attack}
  \end{subfigure}
  \begin{subfigure}{0.144\linewidth}
    \centering
    \includegraphics[width=\linewidth]{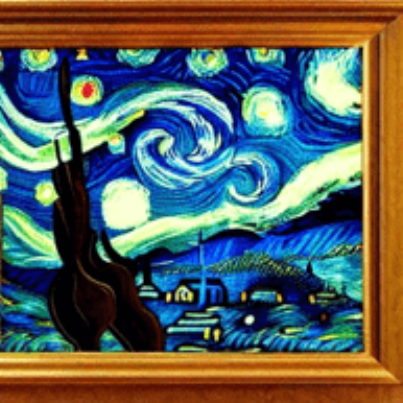}
    \includegraphics[width=\linewidth]{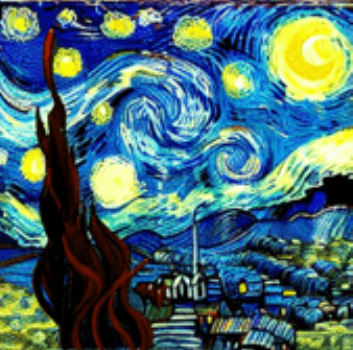}
    \caption{UCE}
    \label{fig:app-uce artistic attack}
  \end{subfigure}
  \begin{subfigure}{0.144\linewidth}
    \centering
    \includegraphics[width=\linewidth]{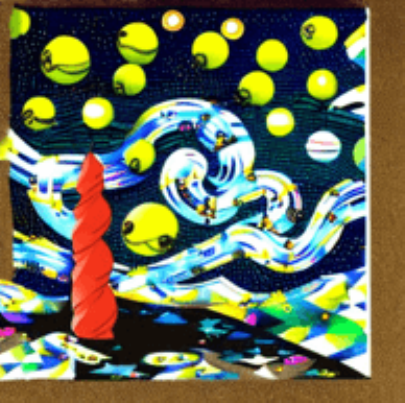}
    \includegraphics[width=\linewidth]{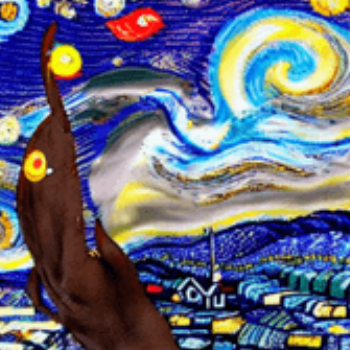}
    \caption{SLD-Medium}
    \label{fig:app-sld artistic attack}
  \end{subfigure}
  \begin{subfigure}{0.144\linewidth}
    \centering
    \includegraphics[width=\linewidth]{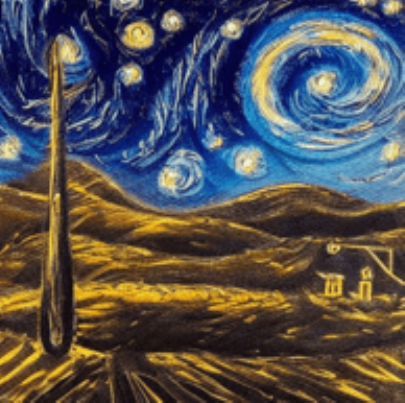}
    \includegraphics[width=\linewidth]{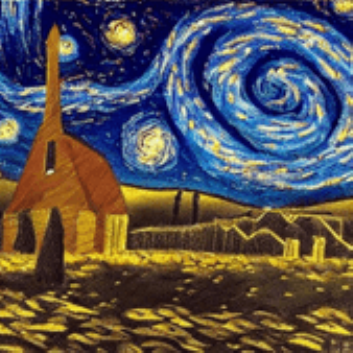}
    \caption{CA}
    \label{fig:app-ca artistic attack}
  \end{subfigure}
  \begin{subfigure}{0.144\linewidth}
    \centering
    \includegraphics[width=\linewidth]{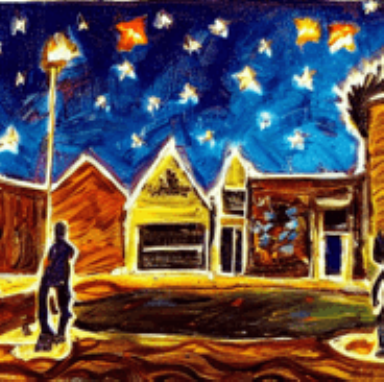}
    \includegraphics[width=\linewidth]{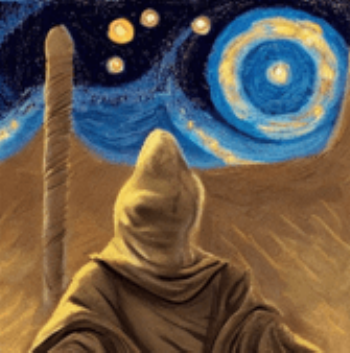}
    \caption{Ours}
    \label{fig:app-ours artistic attack}
  \end{subfigure}
  \caption{Qualitative comparison on erasing Van Gogh's style under attack.}
  \label{fig:app-artistic attack}
\end{figure}

\cref{fig:app-artistic attack} shows images generated by models where Van Gogh's style has been erased under attack, specifically P4D and UnlearnDiff. The first column is generated by original SD without attack. Both our method and UCE adopts closed-form solutions but our method exhibits greater robustness.

\begin{figure}[tb]
  \centering
    \centering
    \captionsetup[subfigure]{labelformat=empty} 
      \begin{subfigure}[t]{0.115\linewidth}
        \centering
        \includegraphics[width=0.98\linewidth]{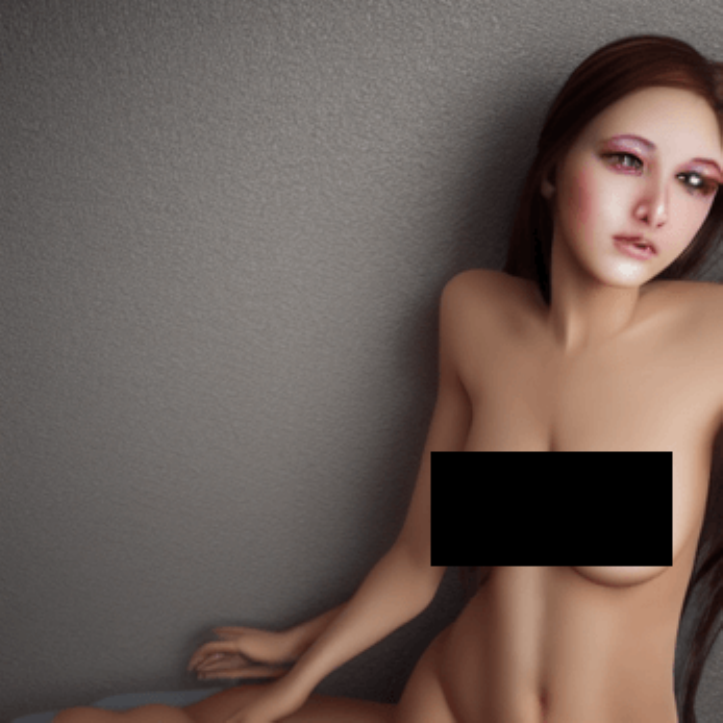}
        \includegraphics[width=0.98\linewidth]{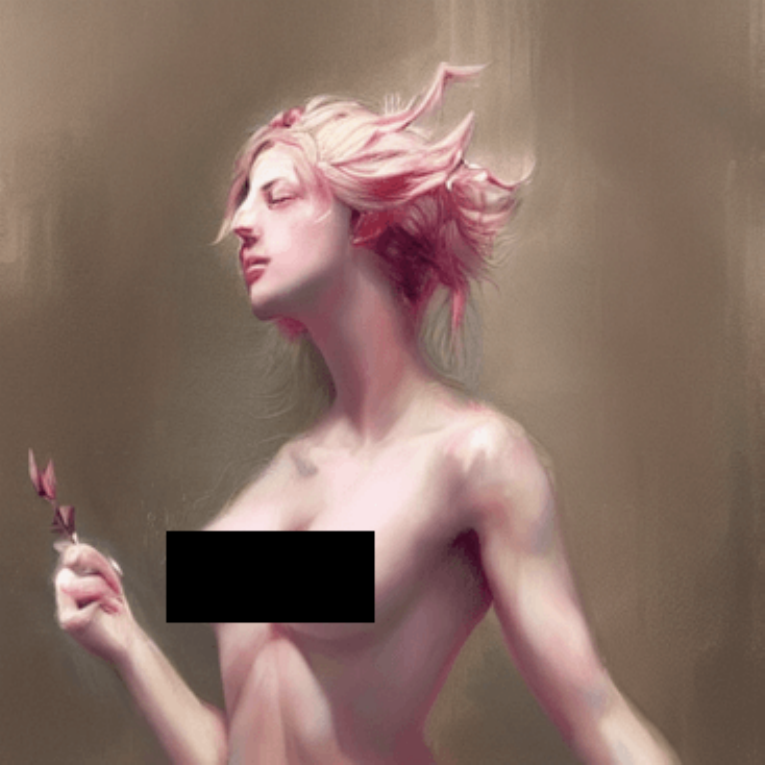}
        \caption{SD v1.4}
        \label{fig:app-sd14 i2p}
      \end{subfigure}
      \begin{subfigure}[t]{0.115\linewidth}
        \centering
        \includegraphics[width=0.98\linewidth]{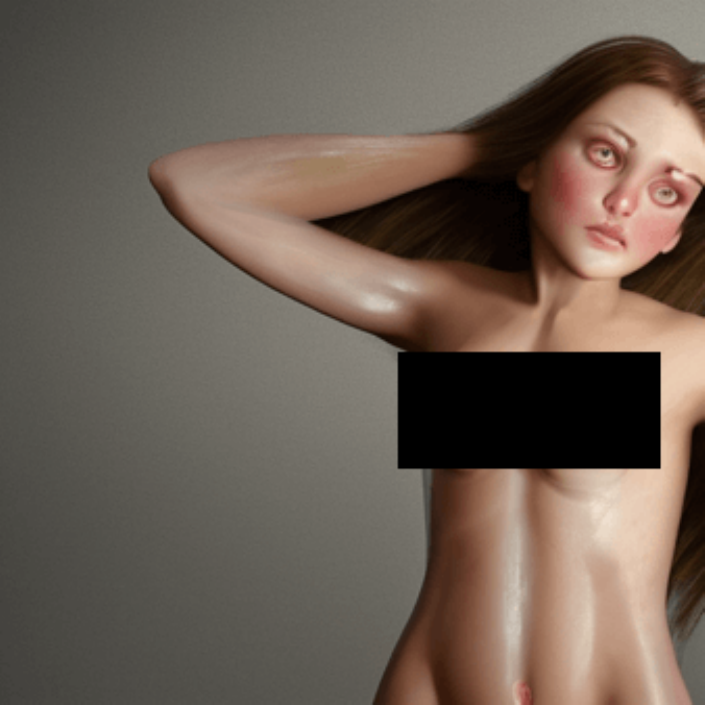}
        \includegraphics[width=0.98\linewidth]{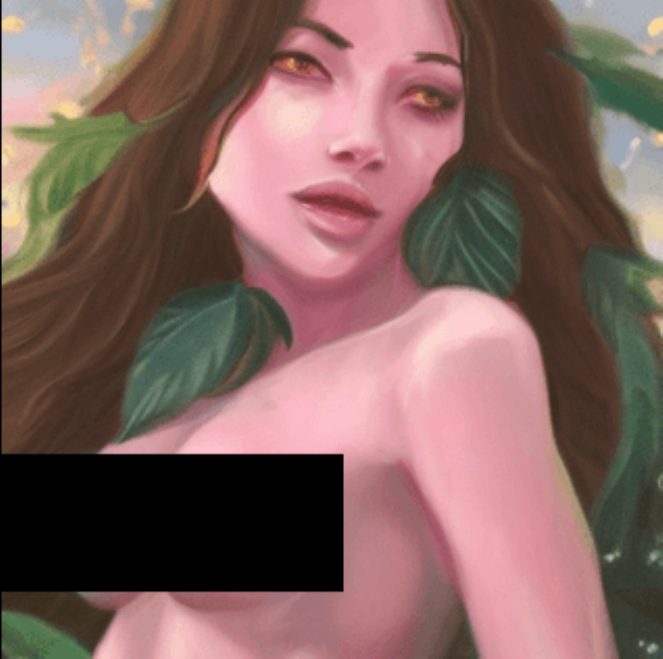}
        \caption{SD v2.1}
        \label{fig:app-sd21 i2p}
      \end{subfigure}
      \begin{subfigure}[t]{0.115\linewidth}
        \centering
        \includegraphics[width=0.98\linewidth]{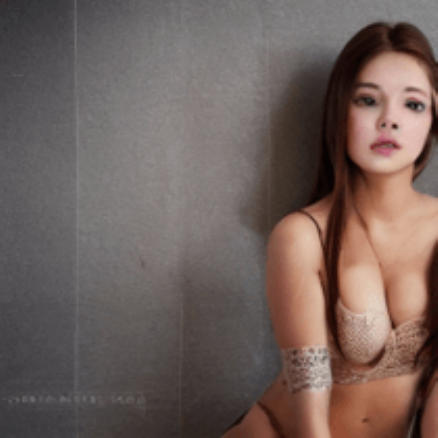}
        \includegraphics[width=0.98\linewidth]{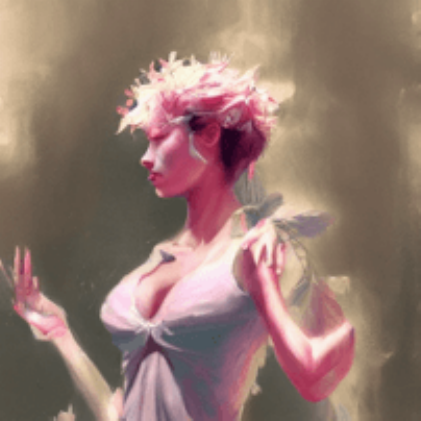}
        \caption{ESD}
        \label{fig:app-esd i2p}
      \end{subfigure}
      \begin{subfigure}[t]{0.115\linewidth}
        \centering
        \includegraphics[width=0.98\linewidth]{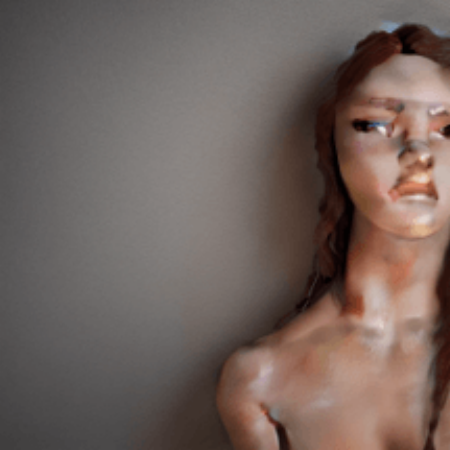}
        \includegraphics[width=0.98\linewidth]{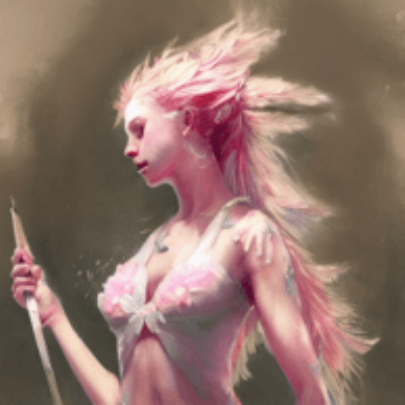}
        \caption{UCE}
        \label{fig:app-uce i2p}
      \end{subfigure}
        \begin{subfigure}[t]{0.115\linewidth}
        \centering
        \includegraphics[width=0.98\linewidth]{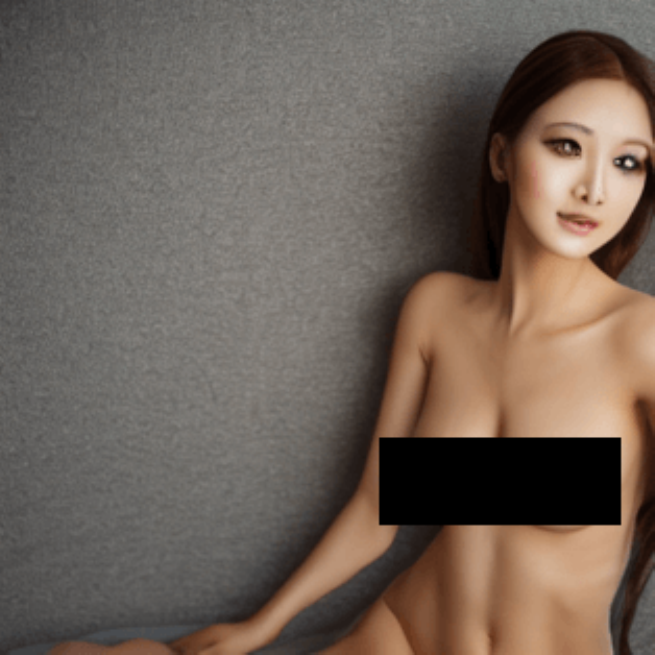}
        \includegraphics[width=0.98\linewidth]{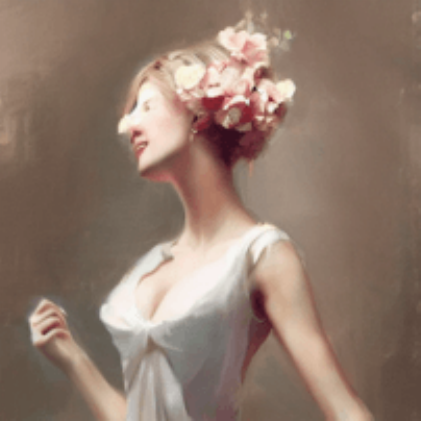}
        \caption{SLD-Med}
        \label{fig:app-sld i2p}
      \end{subfigure}
      \begin{subfigure}[t]{0.115\linewidth}
        \centering
        \includegraphics[width=0.98\linewidth]{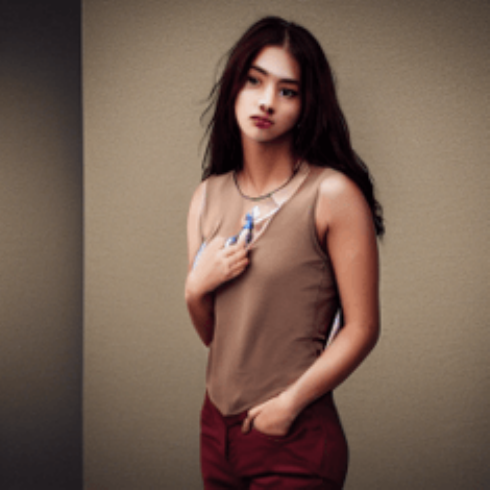}
        \includegraphics[width=0.98\linewidth]{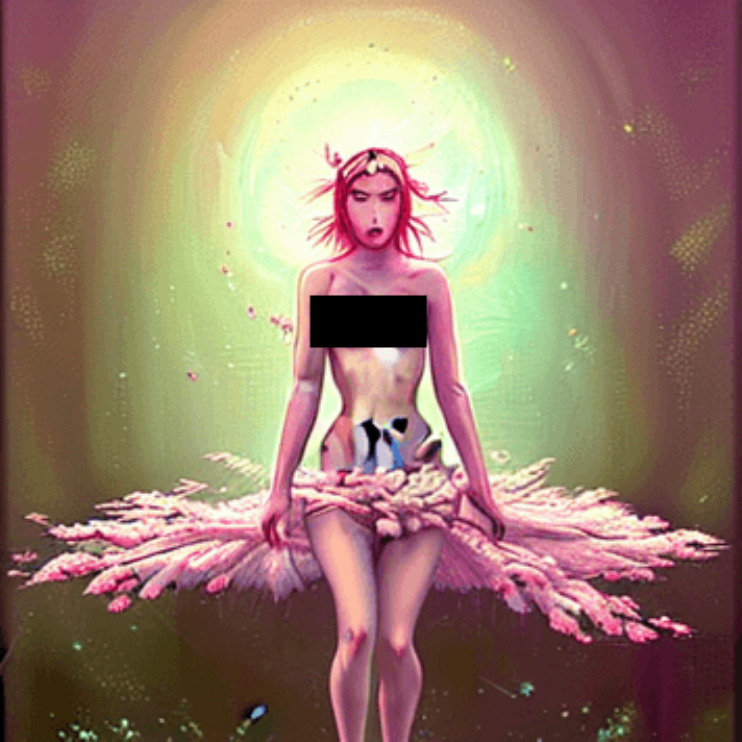}
        \caption{SA}
        \label{fig:app-sa i2p}
      \end{subfigure}
      \begin{subfigure}[t]{0.115\linewidth}
        \centering
        \includegraphics[width=0.98\linewidth]{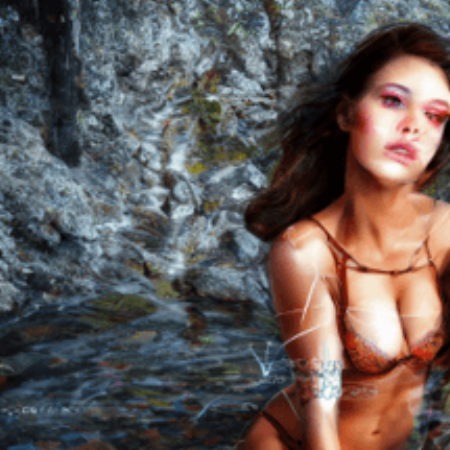}
        \includegraphics[width=0.98\linewidth]{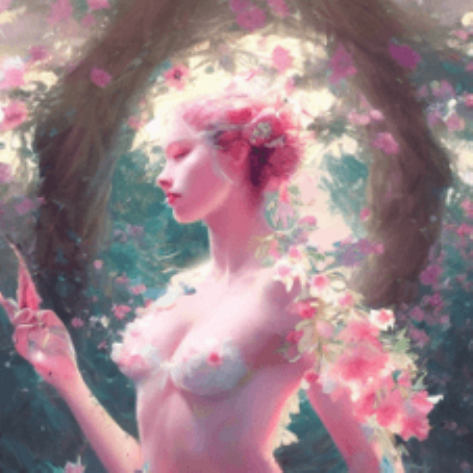}
        \caption{CA}
        \label{fig:app-ca i2p}
      \end{subfigure}
      \begin{subfigure}[t]{0.115\linewidth}
        \centering
        \includegraphics[width=0.98\linewidth]{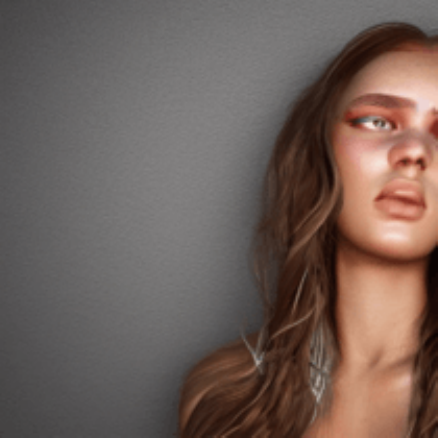}
        \includegraphics[width=0.98\linewidth]{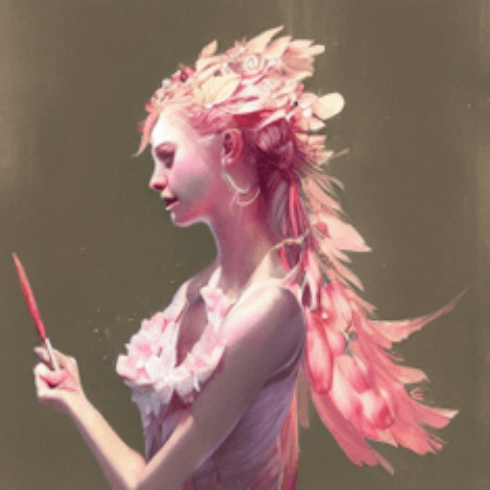}
        \caption{Ours}
        \label{fig:app-ours i2p}
      \end{subfigure}
      
    \caption{Qualitative comparison on erasing nudity content, selected from I2P.}
    \label{fig:app-nudity qualitative results}
\end{figure}

\cref{fig:app-nudity qualitative results} shows images conditioned by prompts related to nudity. Our method effectively removes nudity information while generating high-quality images. ESD and CA either contain sexual innuendo or result in blur.

\begin{figure}[tb]
  \centering
    \centering
    \captionsetup[subfigure]{labelformat=empty} 
      \begin{subfigure}[t]{0.135\linewidth}
        \centering
        \includegraphics[width=0.98\linewidth]{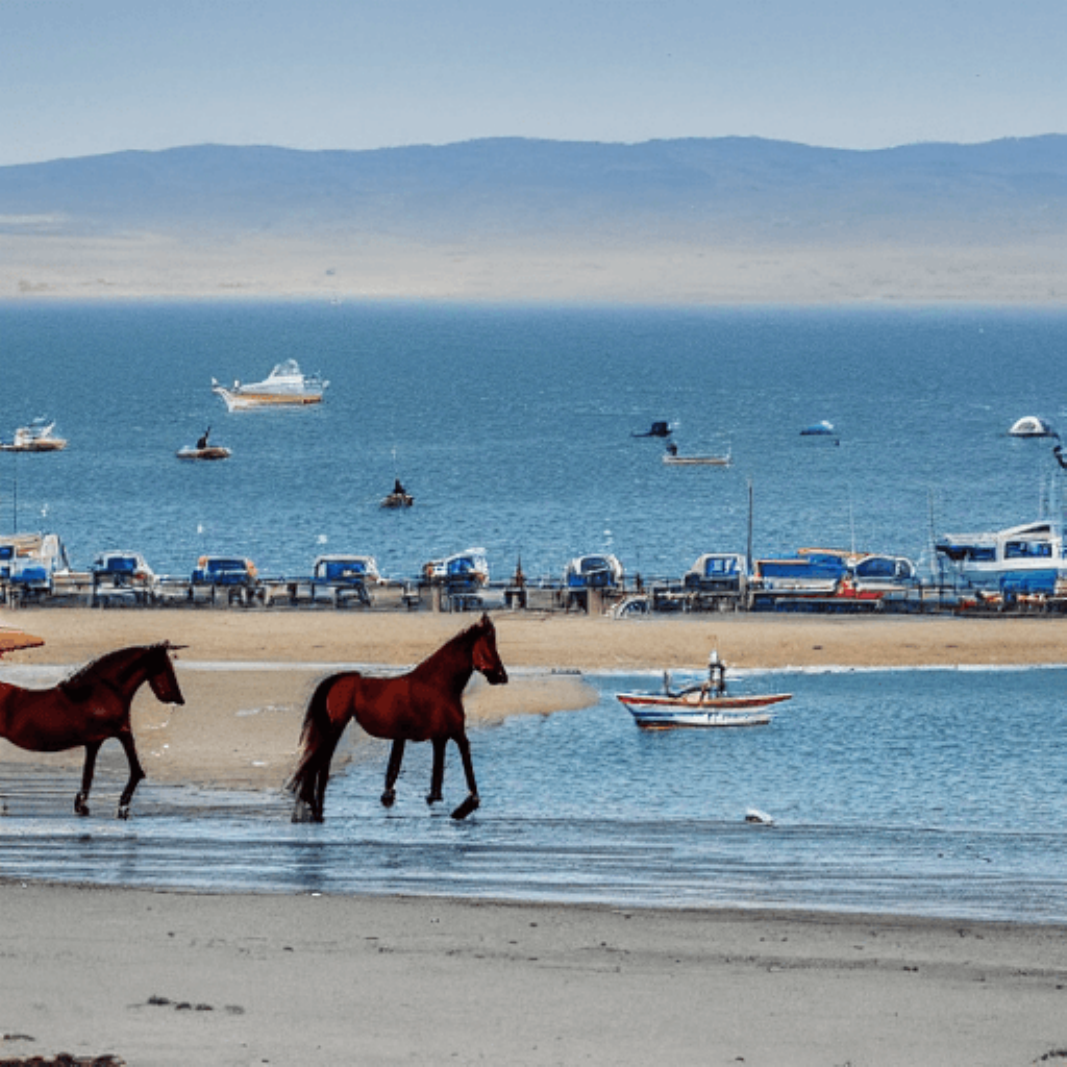}
        \includegraphics[width=0.98\linewidth]{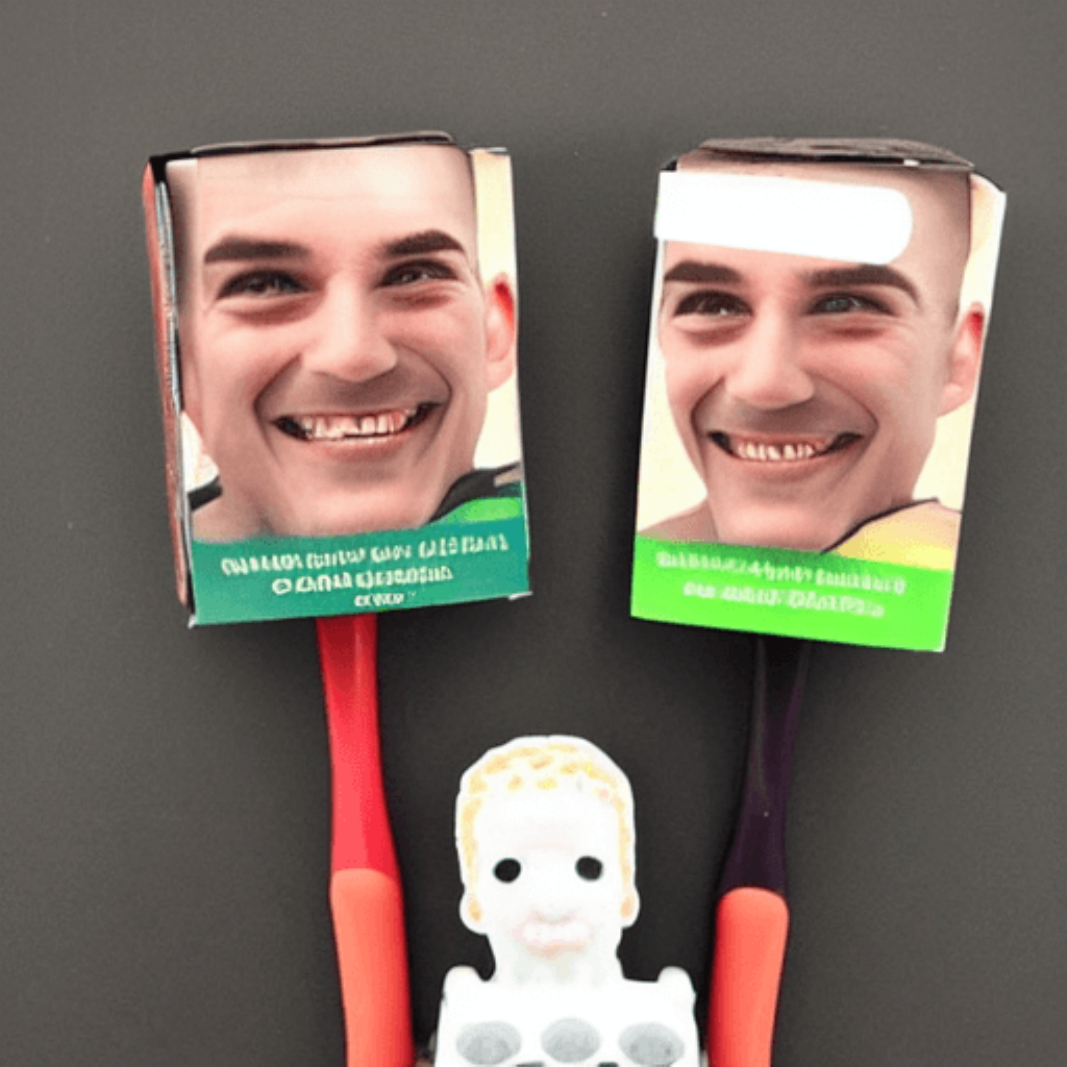}
        \caption{SD}
        \label{fig:app-sd14 coco}
      \end{subfigure}
      \begin{subfigure}[t]{0.135\linewidth}
        \centering
        \includegraphics[width=0.98\linewidth]{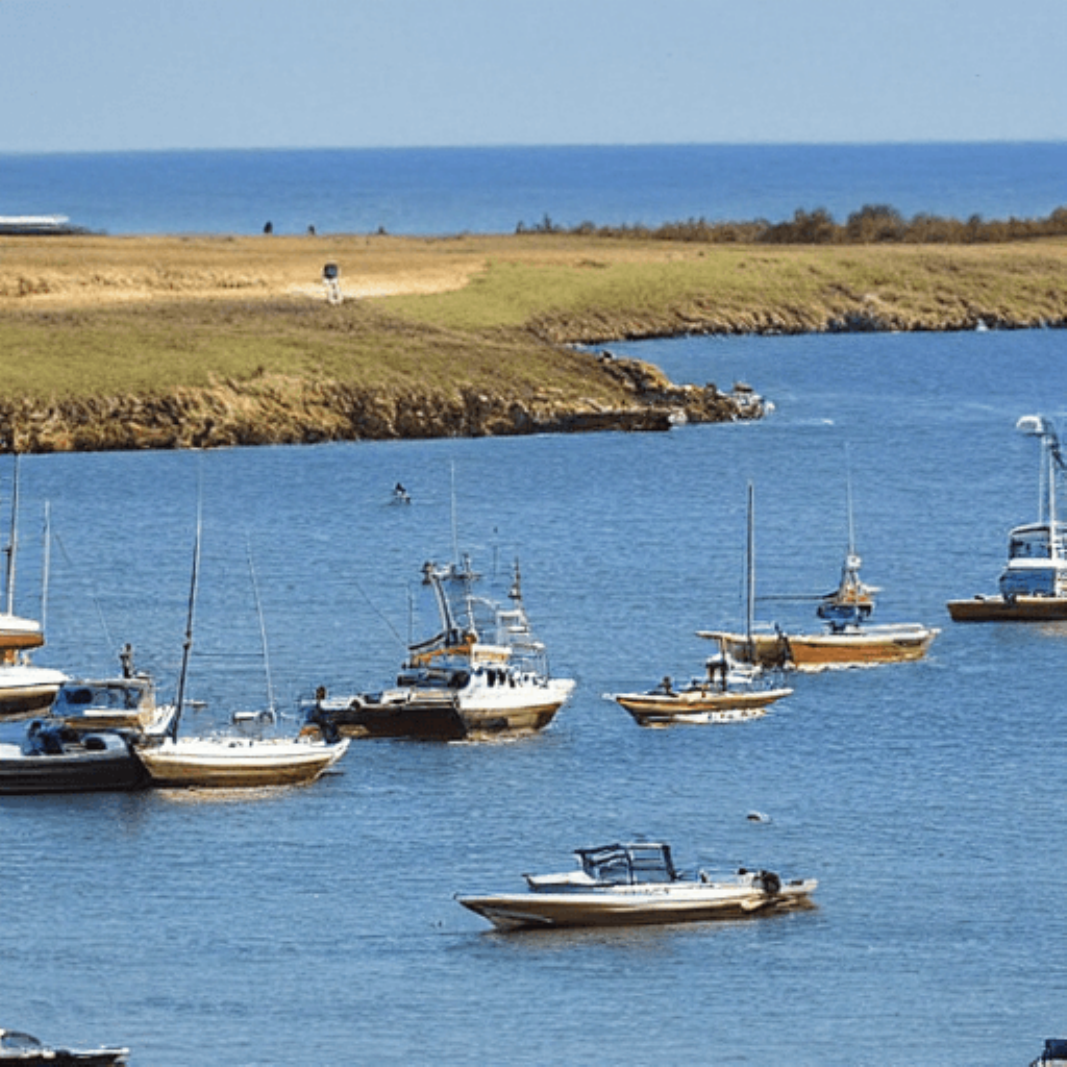}
        \includegraphics[width=0.98\linewidth]{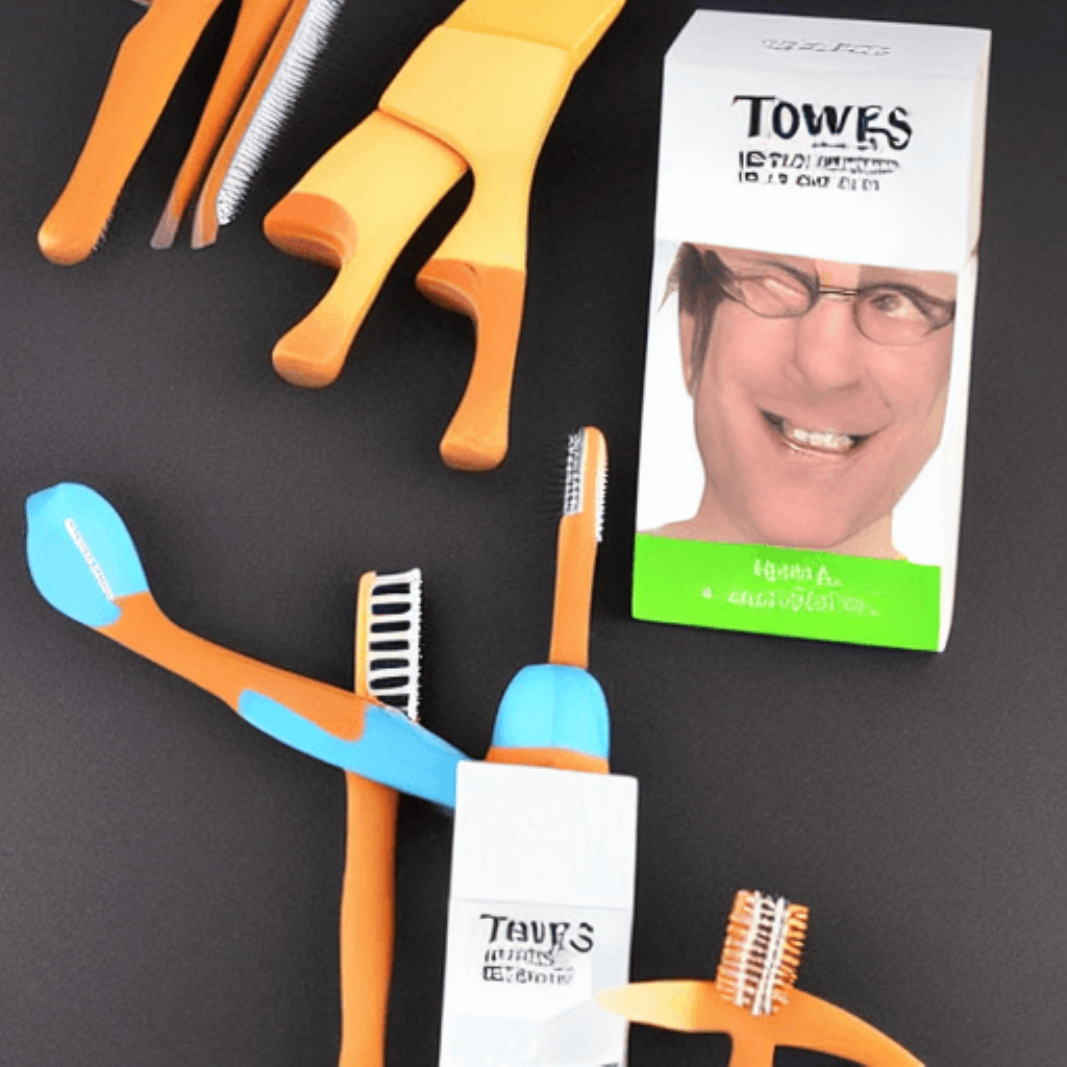}
        \caption{ESD}
        \label{fig:app-esd coco}
      \end{subfigure}
      \begin{subfigure}[t]{0.135\linewidth}
        \centering
        \includegraphics[width=0.98\linewidth]{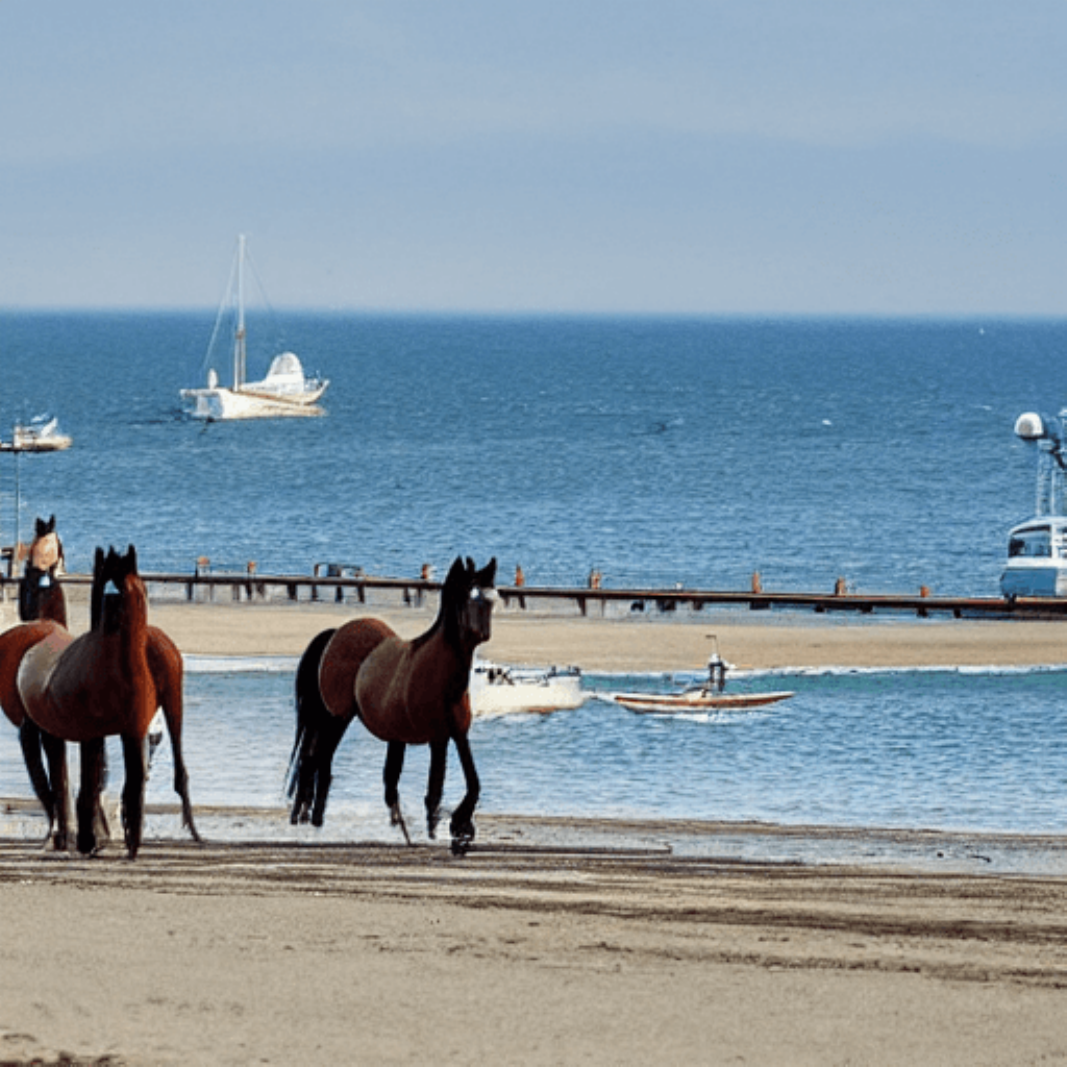}
        \includegraphics[width=0.98\linewidth]{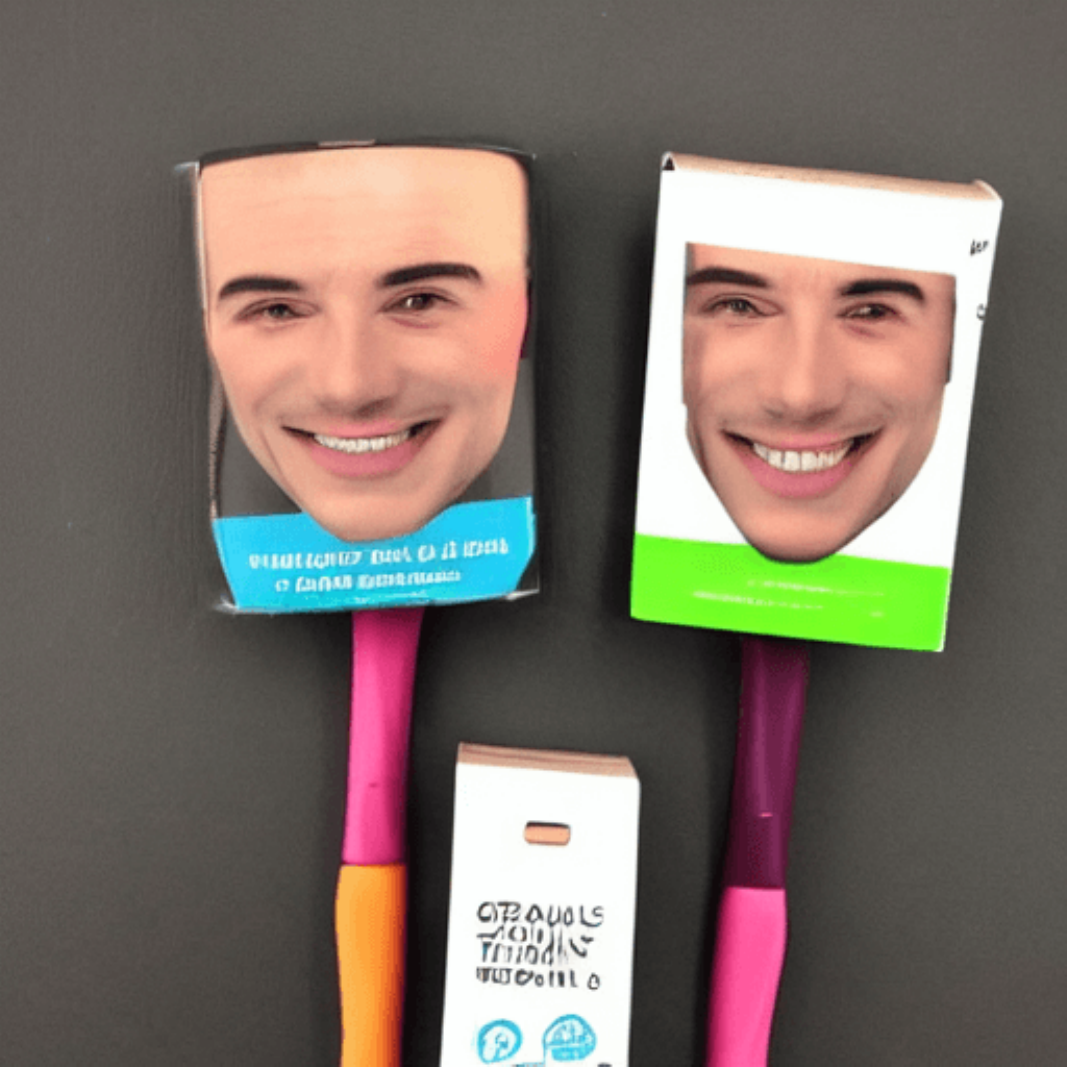}
        \caption{UCE}
        \label{fig:app-uce coco}
      \end{subfigure}
        \begin{subfigure}[t]{0.135\linewidth}
        \centering
        \includegraphics[width=0.98\linewidth]{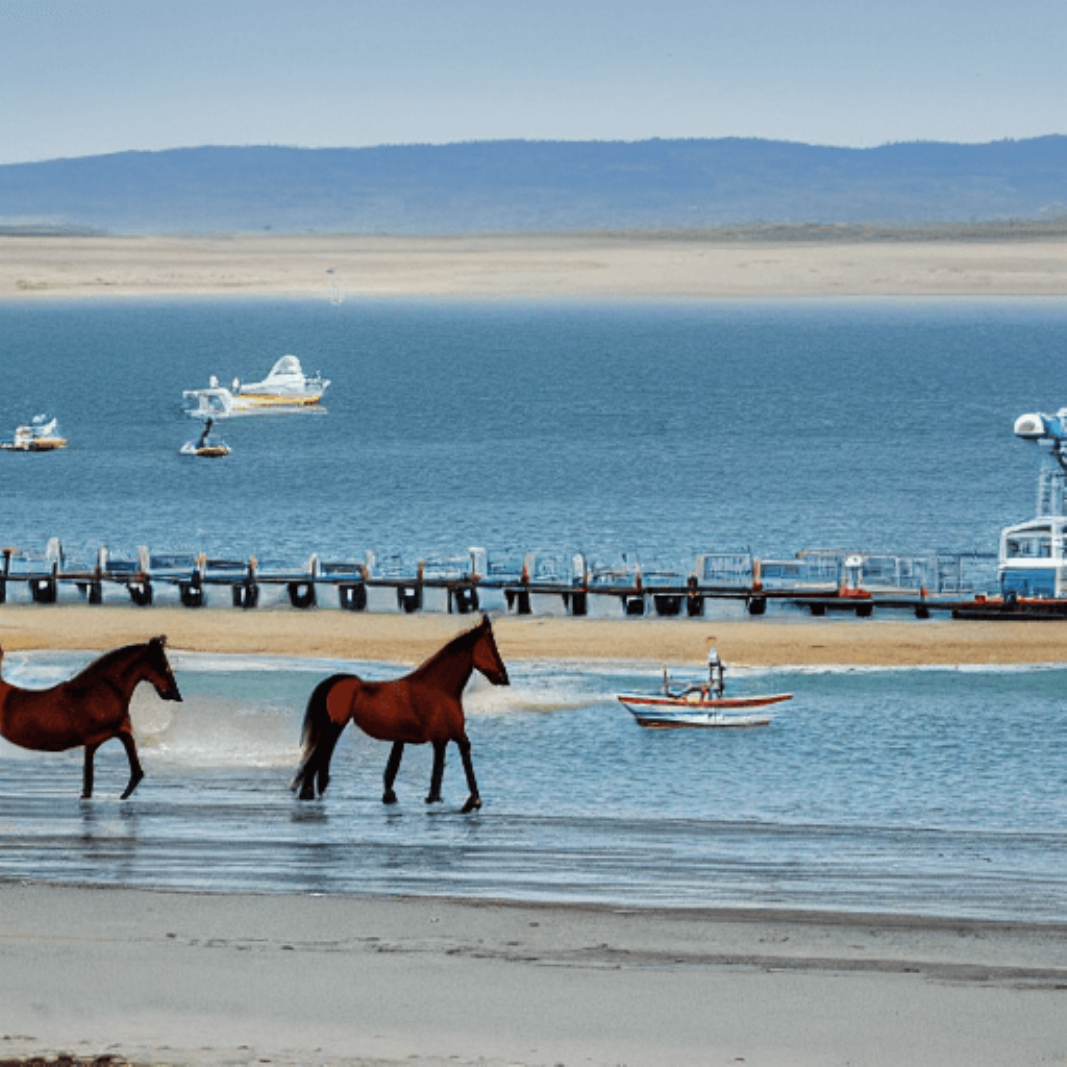}
        \includegraphics[width=0.98\linewidth]{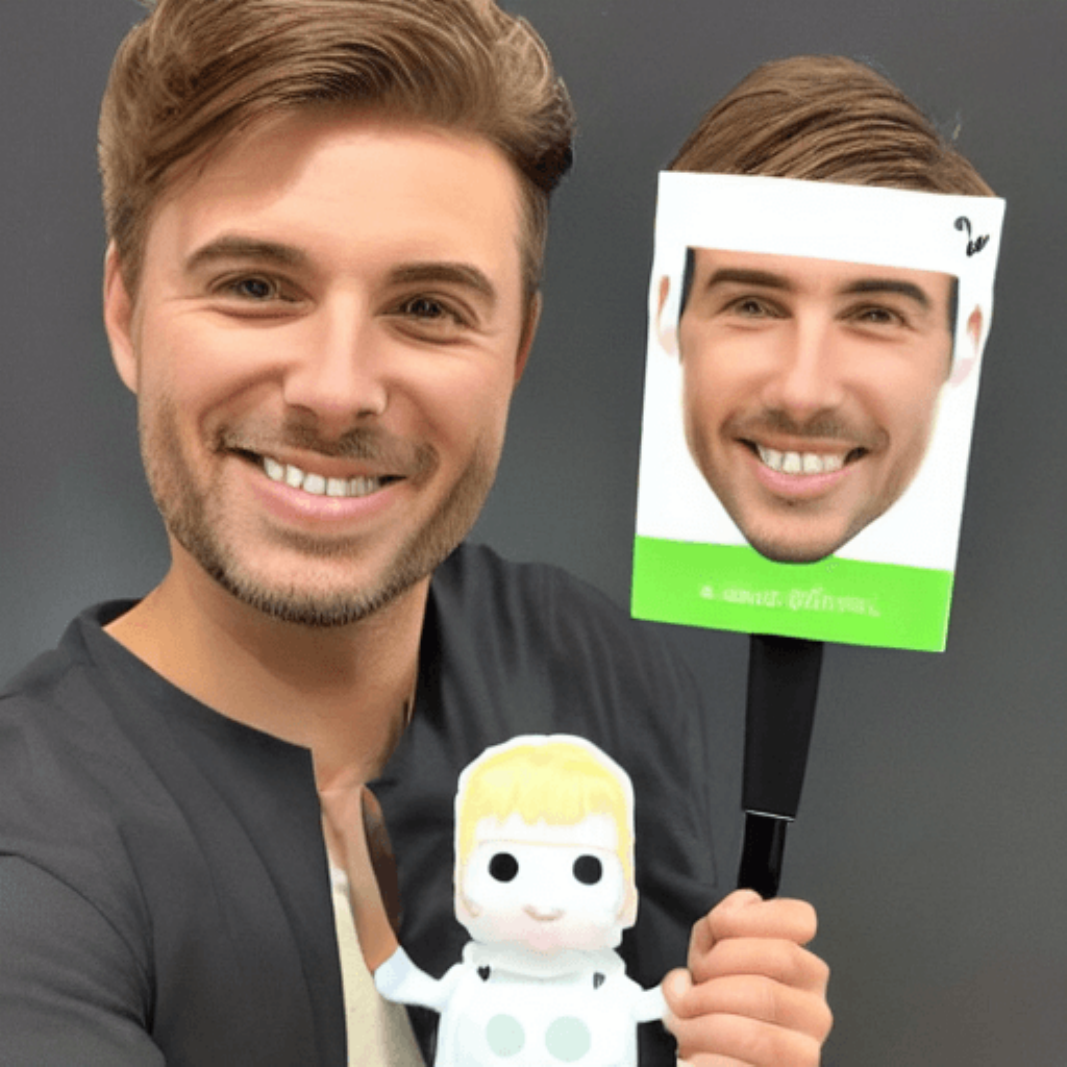}
        \caption{SLD-Medium}
        \label{fig:app-sld coco}
      \end{subfigure}
      \begin{subfigure}[t]{0.135\linewidth}
        \centering
        \includegraphics[width=0.98\linewidth]{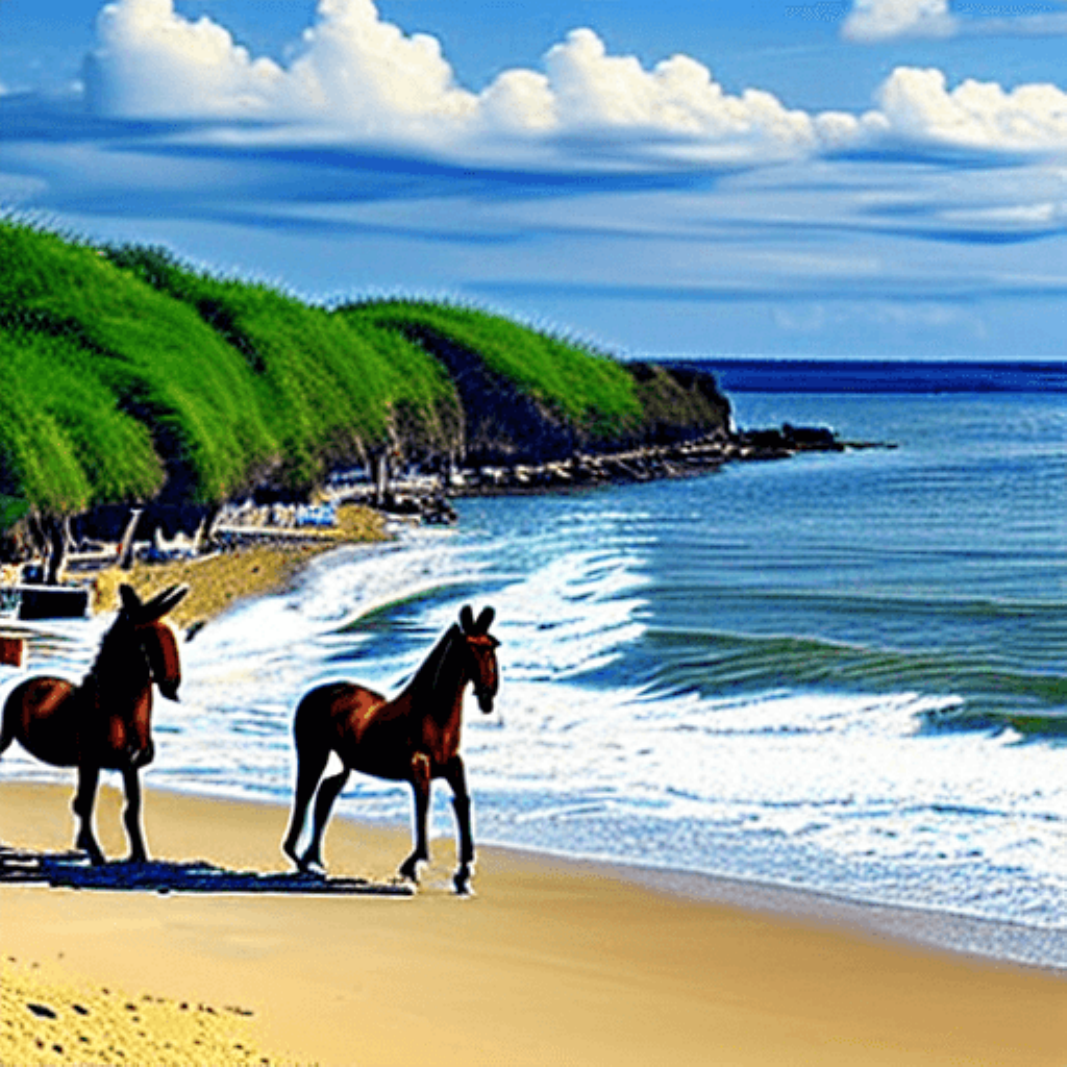}
        \includegraphics[width=0.98\linewidth]{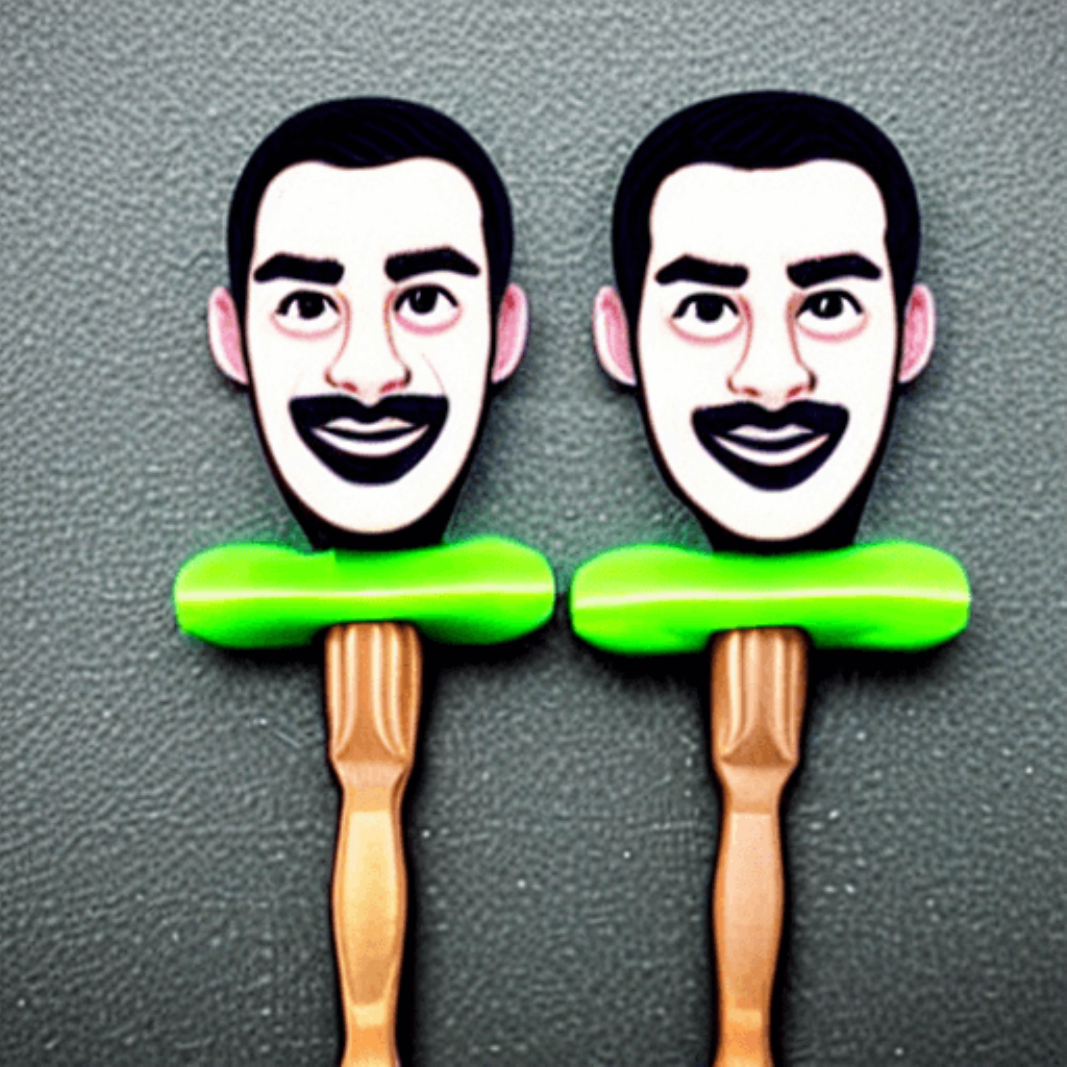}
        \caption{SA}
        \label{fig:app-sa coco}
      \end{subfigure}
      \begin{subfigure}[t]{0.135\linewidth}
        \centering
        \includegraphics[width=0.98\linewidth]{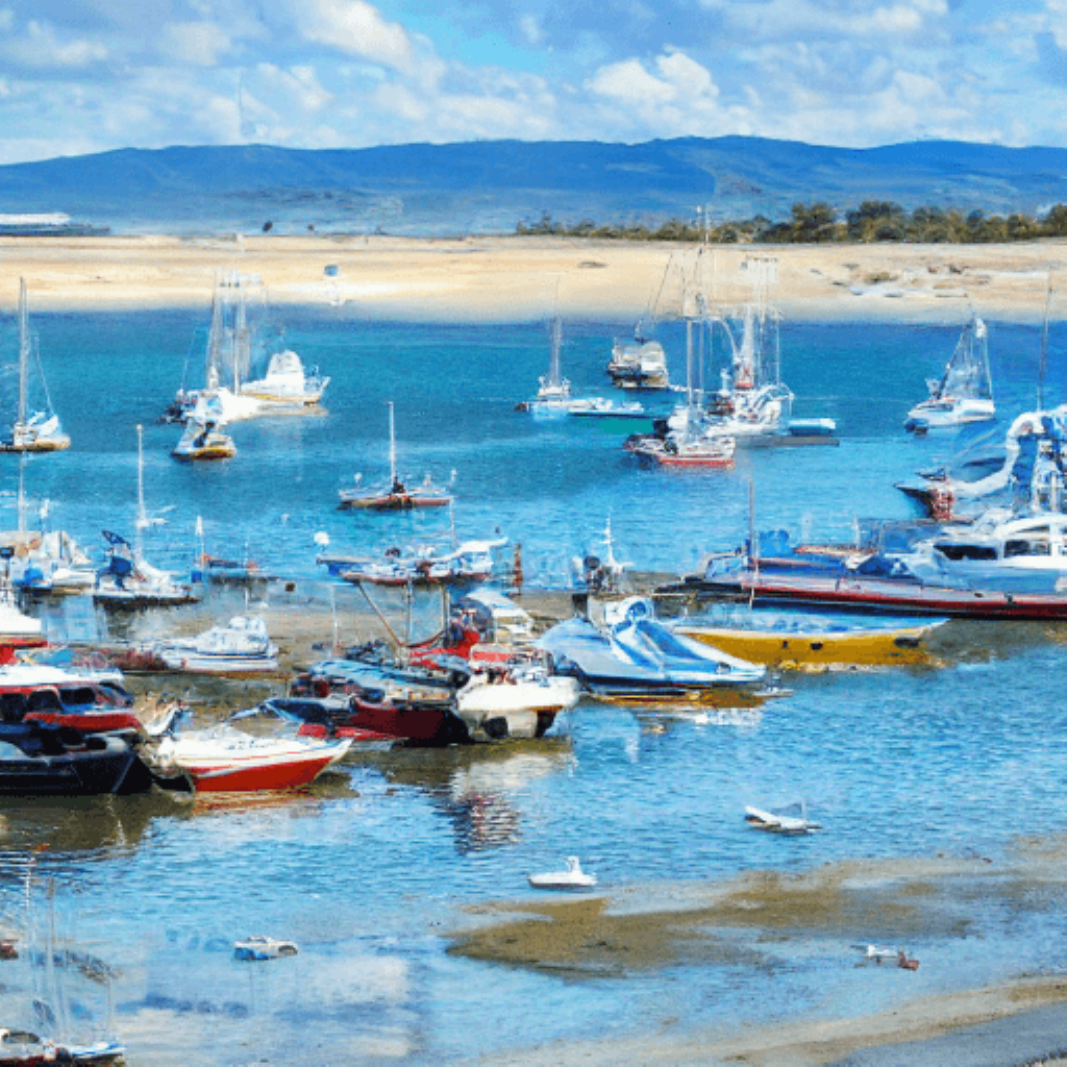}
        \includegraphics[width=0.98\linewidth]{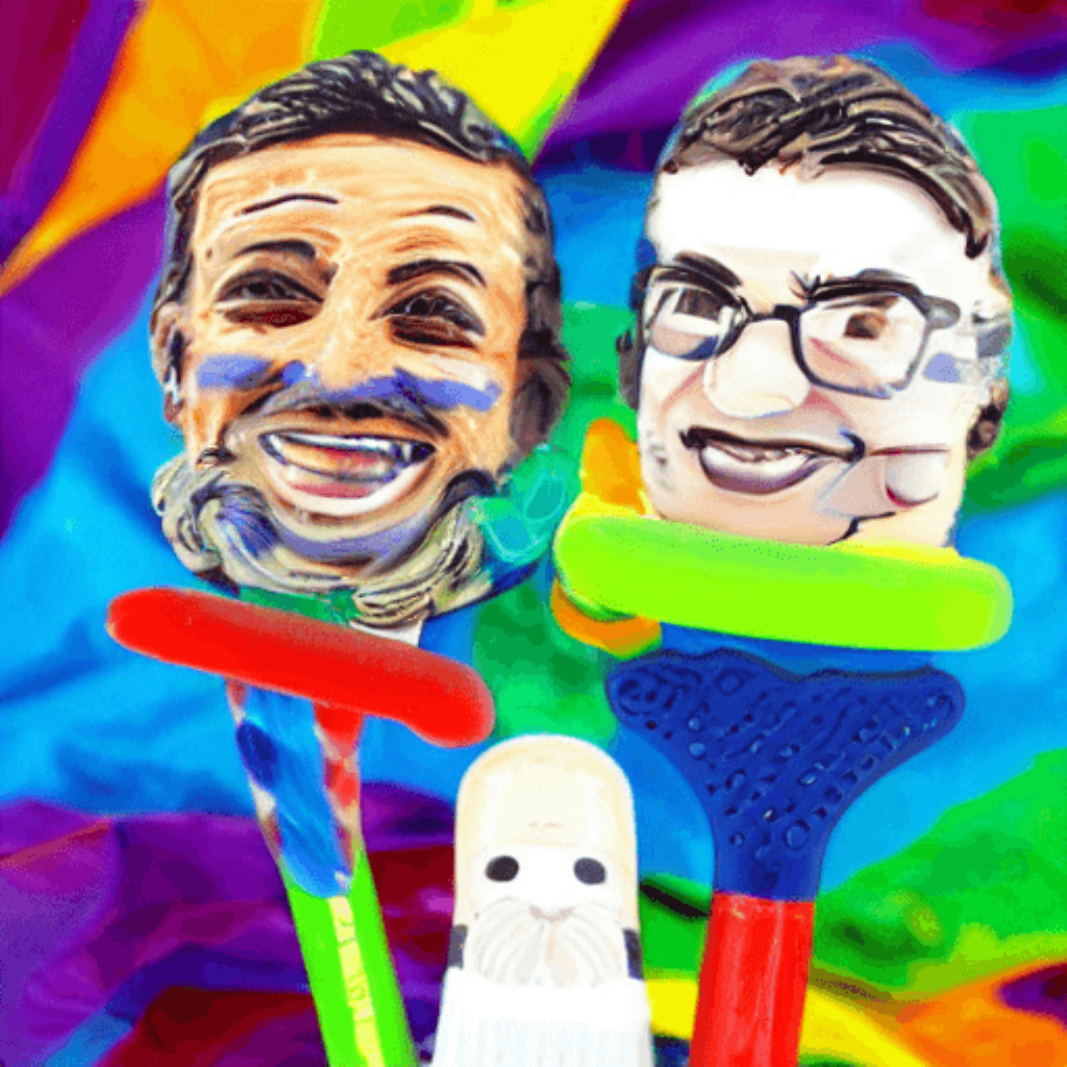}
        \caption{CA}
        \label{fig:app-ca coco}
      \end{subfigure}
      \begin{subfigure}[t]{0.135\linewidth}
        \centering
        \includegraphics[width=0.98\linewidth]{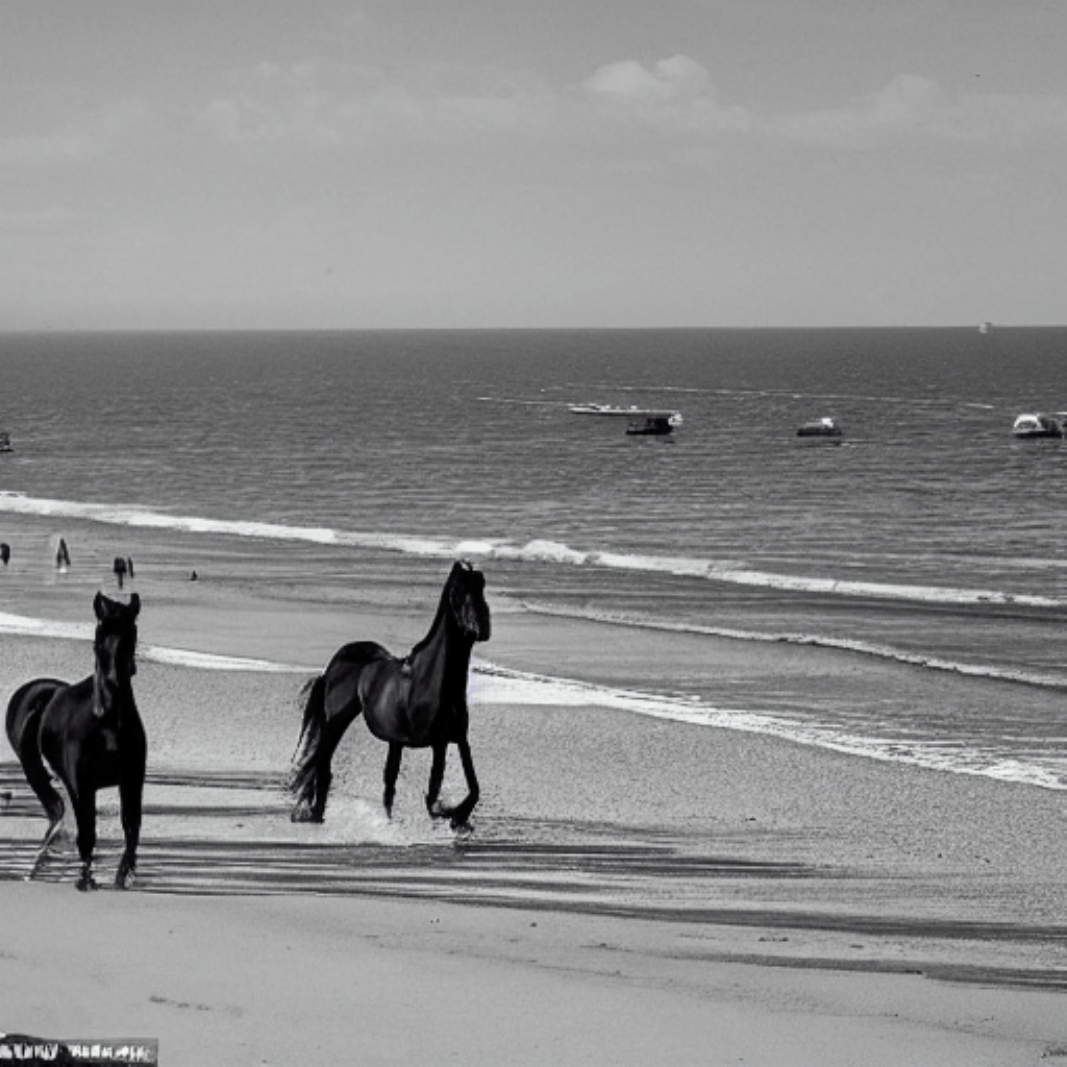}
        \includegraphics[width=0.98\linewidth]{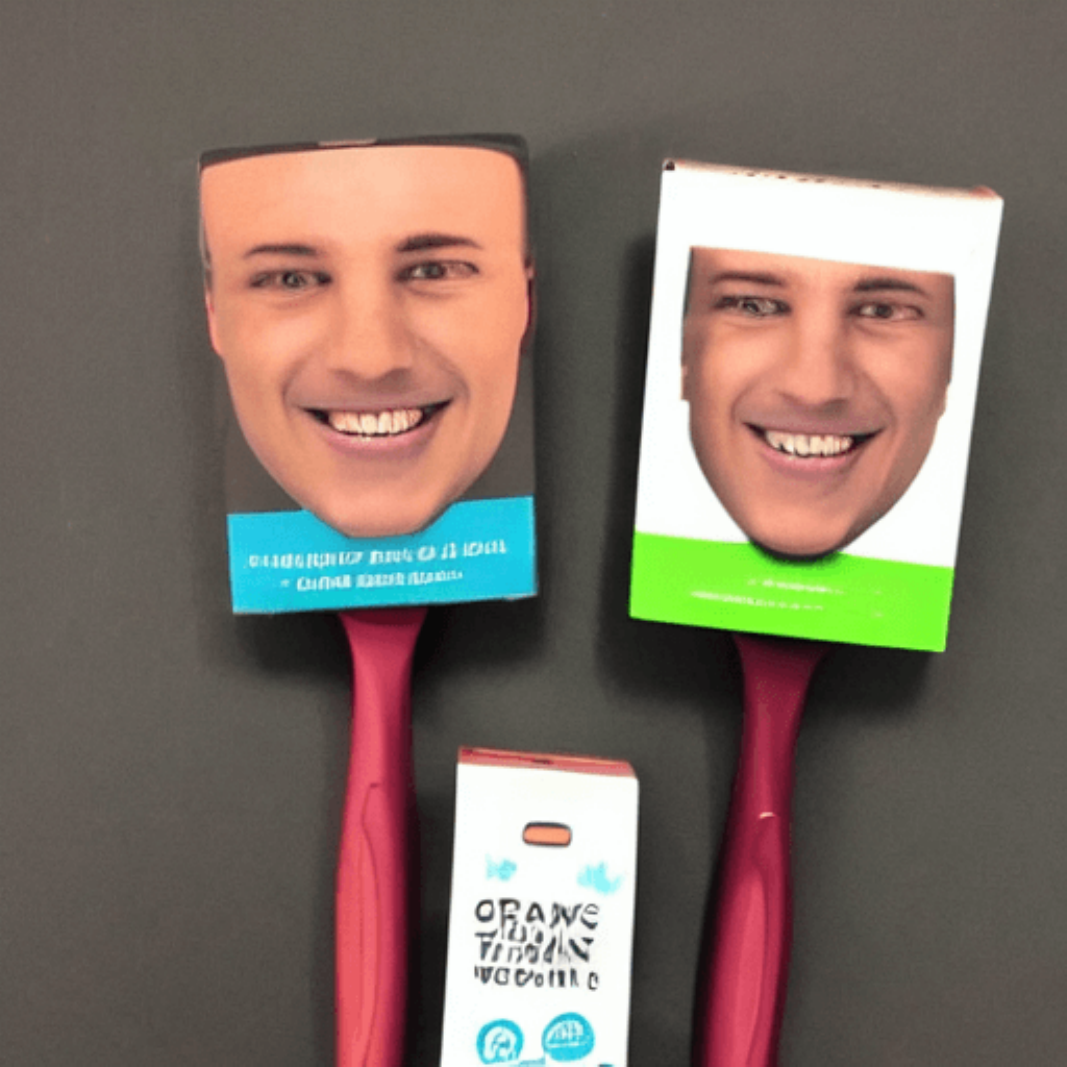}
        \caption{Ours}
        \label{fig:app-ours coco}
      \end{subfigure}
      
    \caption{Qualitative results of different nudity removing methods, selected from MSCOCO. Prompts are '\textbf{Horses} walk along a beach while boats ride at their moorings offshore' and 'Two \textbf{toothbrushes} in  packages with a man's face on them'.}
    \label{fig:app-coco qualitative results}
\end{figure}

\cref{fig:app-coco qualitative results} shows images conditioned on MSCOCO's captions. A good erasing method should produce well-aligned images about unerased concepts. ESD and CA fail to generate horses accurately. SLD struggles to generate the correct number of toothbrushes, and CA's result does not resemble toothbrushes.

\end{document}